\documentclass[10pt,twocolumn,letterpaper]{article}

\usepackage{cvpr}              

\usepackage{multirow}
\usepackage{array}
\usepackage{amsthm}
\theoremstyle{definition}
\newtheorem{definition}{Definition}
\usepackage{algorithm}
\usepackage{algpseudocode}
\usepackage{amsthm}
\usepackage{makecell}
\usepackage{color} 
\usepackage{subcaption}
\usepackage{supertabular,booktabs,graphicx}

\input{trajectoryformat}

\usepackage{tcolorbox}
\tcbuselibrary{skins,breakable,listings}
\newtcblisting{PromptBox}[2][]{%
  listing only,
  colback=gray!2,
  colframe=black!20,
  breakable,
  arc=2mm,
  outer arc=2mm,
  boxsep=1mm,
  left=1mm,right=1mm,top=1mm,bottom=1mm,
  title={#2},
  fonttitle=\bfseries,
  listing options={
    basicstyle=\ttfamily\small,
    columns=fullflexible,
    breaklines=true,
    breakindent=0pt,         
    breakautoindent=false  
  },
  #1
}

\usepackage{pifont}

\usepackage{arydshln}

\definecolor{cvprblue}{rgb}{0.21,0.49,0.74}
\usepackage[pagebackref,breaklinks,colorlinks,allcolors=cvprblue]{hyperref}

\def\paperID{8848} 
\def\confName{CVPR}
\def\confYear{2026}

\title{M$^3$-Bench: Multi-Modal, Multi-Hop, Multi-Threaded Tool-Using MLLM Agent Benchmark}

\author{
Yang Zhou\thanks{Equal contribution.}\quad
Mingyu Zhao\footnotemark[1]\quad 
Zhenting Wang\quad 
Difei Gu\quad
Bangwei Guo\quad \\
Ruosong Ye\quad
Ligong Han\quad
Can Jin\quad 
Dimitris N.\ Metaxas\\[3pt]
Rutgers University\\
\texttt{\{eta.yang,\,zhenting.wang,\,bangwei.guo\}@rutgers.edu}\\
\texttt{\{ligong.han,\,can.jin,\,dnm\}@rutgers.edu}\\
\texttt{\{mz751,\,dg1188,\,ry233\}@scarletmail.rutgers.edu}
}


\begin{document}
\maketitle
\begin{abstract}
    We present M$^3$-Bench, the first benchmark for evaluating multimodal tool use under the Model Context Protocol. 
   The benchmark targets realistic, multi-hop and multi-threaded workflows that require visual grounding and textual reasoning, cross-tool dependencies, and persistence of intermediate resources across steps. 
    We introduce a similarity-driven alignment that serializes each tool call, embeds signatures with a sentence encoder, and performs similarity-bucketed Hungarian matching to obtain auditable one-to-one correspondences. 
   On top of this alignment, we report interpretable metrics that decouple semantic fidelity from workflow consistency. The benchmark spans 27 servers with 232 tools, and provides standardized trajectories curated through an Executor \& Judge pipeline with human verification; 
    an auxiliary four large language models (LLMs) judge ensemble reports end-task Task Completion and information grounding. Evaluations of representative state-of-the-art Multimodal LLMs (MLLMs) reveal persistent gaps in multimodal MCP tool use, particularly in argument fidelity and structure consistency, underscoring the need for methods that jointly reason over images, text, and tool graphs. Our Benchmark's anonymous repository is at \url{https://github.com/EtaYang10th/Open-M3-Bench}
    \end{abstract}

\section{Introduction}
\label{sec:Introduction}

\begin{figure}
\centering
\includegraphics[width=1\linewidth]{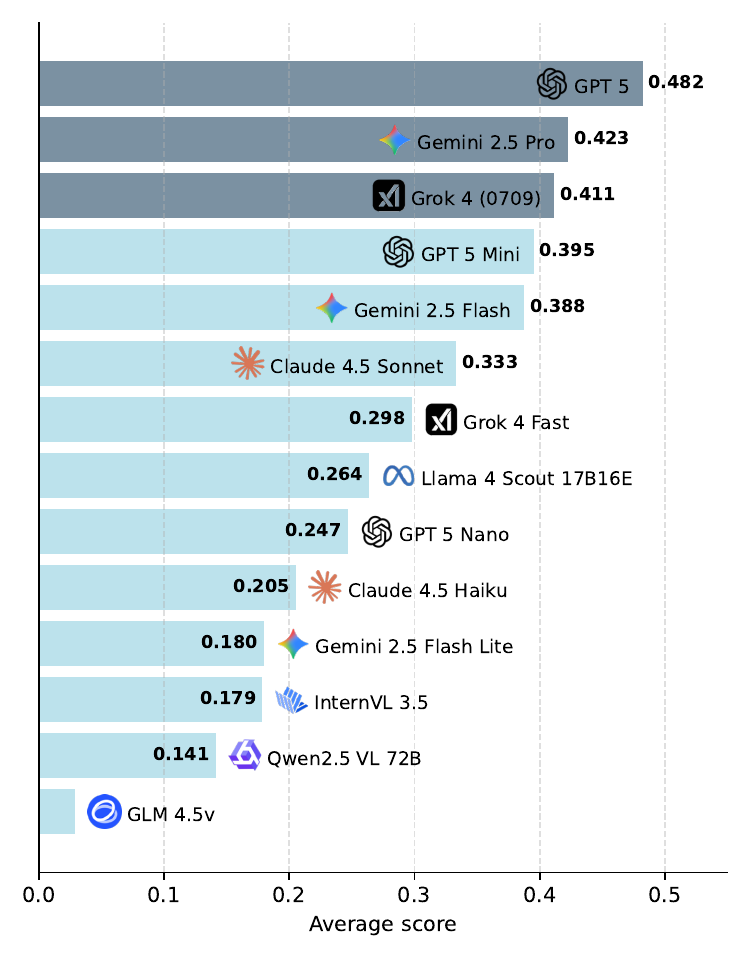}
\caption{Summary leaderboard of state-of-the-art (SOTA) MLLMs on M$^3$-Bench (higher is better).}
\label{fig:metrics_mllm_step_eval}
\vspace{-8pt}
\end{figure}

We introduce M$^3$-Bench, the \emph{Multi-Modal}, \emph{Multiplex}, \emph{Matching-aware} MCP Benchmark, as a principled evaluation suite for multimodal tool use under the Model Context Protocol (MCP). Multimodal Large Language Models (MLLMs) have recently shown that, once they are allowed to perform function calling (tool use), they can query external services and reason over information that is not contained in their parameters \cite{schick2023toolformer,yao2023react,shen2023hugginggpt, openai2025c_o3_o4mini, comanici2025gemini25, anthropic2025claude4, yang2025qwen3, teamkimi2025k2, zeng2025glm45, chen2025minimaxm1}. MCP specifies how models communicate with heterogeneous tools through standardized servers, which makes execution more reproducible across systems \cite{mcp-spec-2025}. However, existing MCP benchmarks are mostly text-only and focus on linear API planning or database queries \cite{qin2023toollm,patil2025bfcl,yao2025taubench,gao2025mcpradar,liu2025mcpeval,luo2025mcpuniverse,fan2025mcptoolbenchpp,scale2025mcpatlas}. A systematic evaluation of \emph{multimodal} MCP workflows, where images and text jointly condition tool calls and results, is still missing \cite{anthropic2024mcp,luo2025mcpuniverse}. 

\begin{figure*}[!htb]
\centering
    \begin{subfigure}{0.22\textwidth}
    \includegraphics[width=1\linewidth]{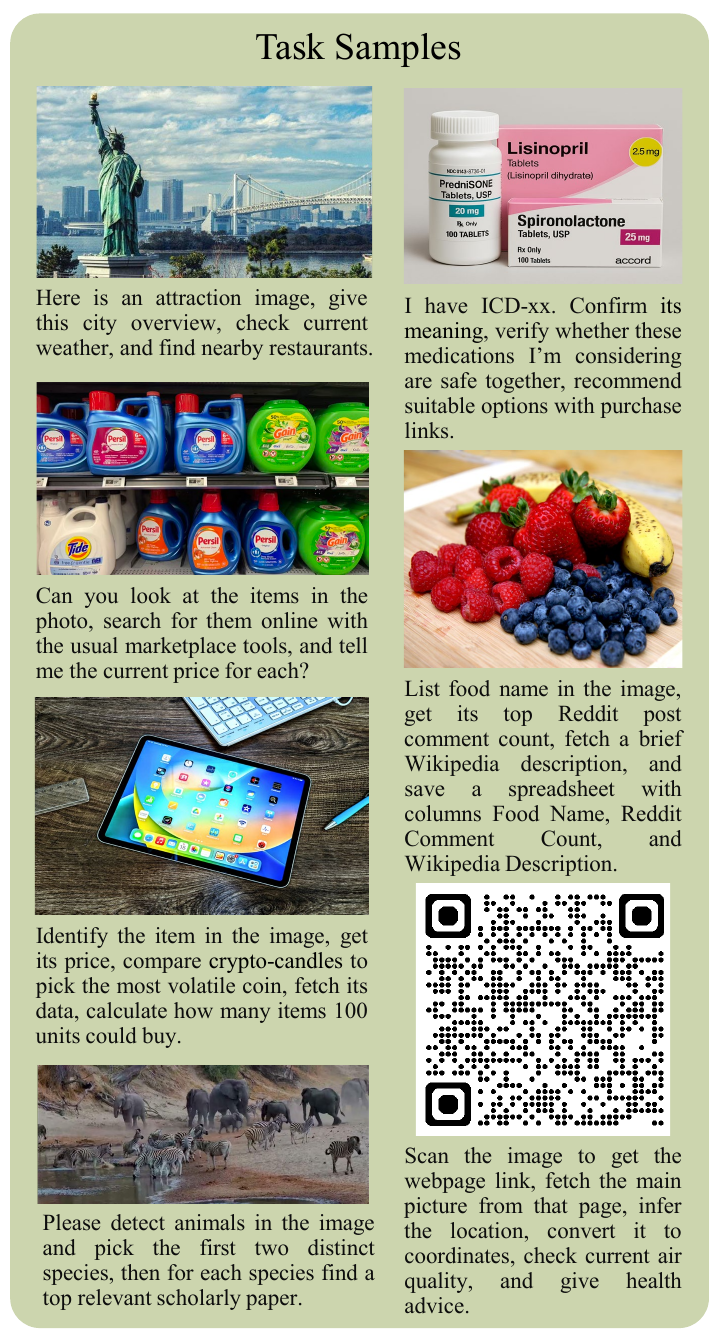}
    \caption{}\label{fig:task_sample}
    \end{subfigure}
\hfill
    \begin{subfigure}{0.77\textwidth}
    \includegraphics[width=1\linewidth]{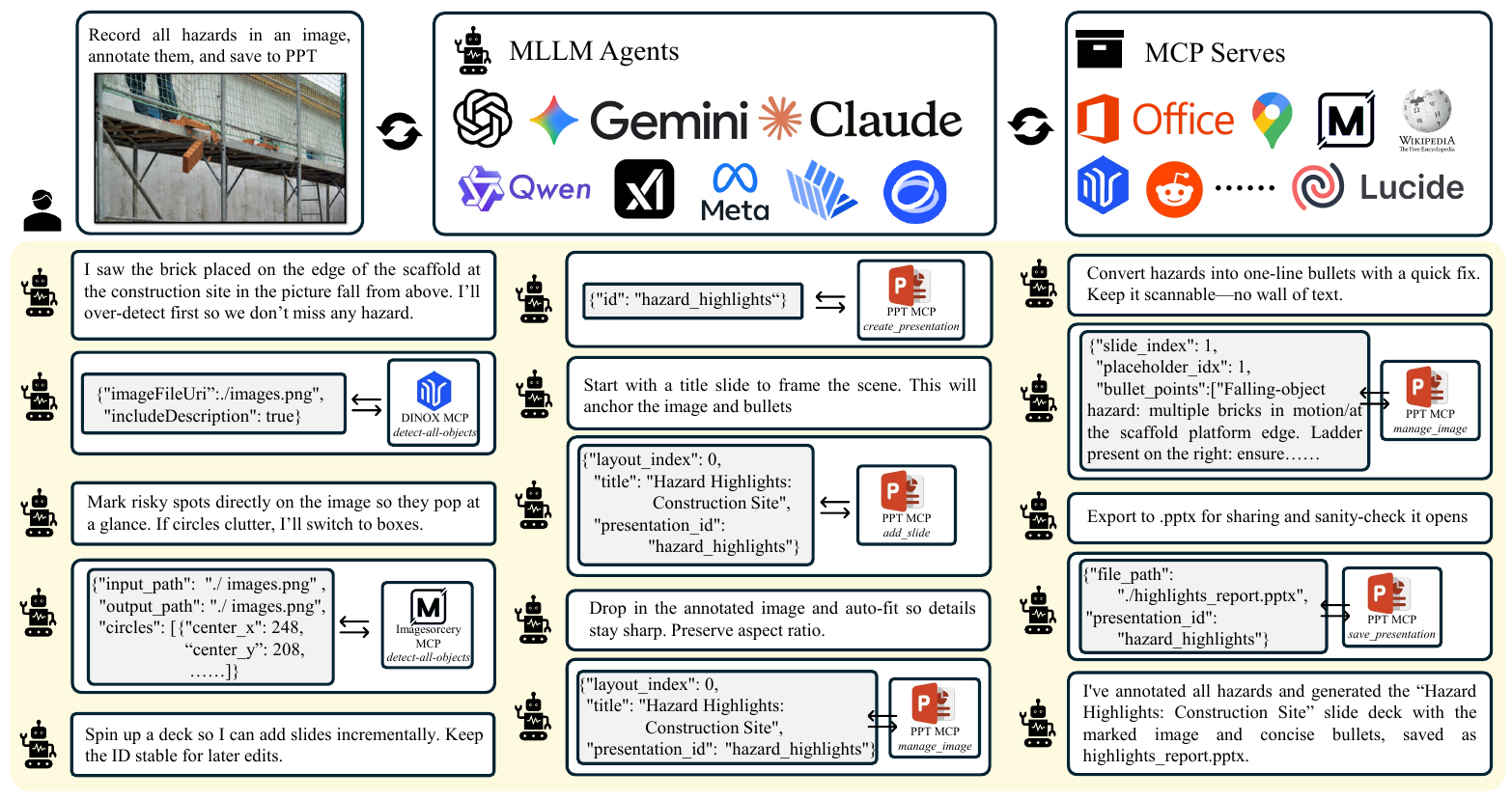}
    \caption{}\label{fig:realistic_challenges}
    \end{subfigure}
\caption{Overview of tasks and a realistic multimodal MCP workflow. (a) Representative task samples in M$^3$-Bench. (b) A realistic multimodal MCP workflow: the agent (i) calls a vision MCP server to detect and mark hazards in the input image, (ii) calls Imagesorcery MCP to process the image, and (iii) calls an office/productivity MCP server to create and populate a slide deck.}
\vspace{-8pt}
\end{figure*}

The core challenge in real-world MCP trajectories is visual grounding: multimodal tool invocation hinges on correctly interpreting the image before any tool can be parameterized. In \autoref{fig:task_sample}, the agent receives a photograph and must first resolve the landmark/city from visual cues, only then can it condition subsequent MCP calls; The agent must fully recognize the fully stocked and well-displayed products on the shelf in a photograph before proceeding to the next MCP tools call.
Second, real-world MCP trajectories are not single-shot calls. They are multi-hop, they contain causally dependent operations, and they frequently execute several tool calls in parallel within one step \cite{mialon2023gaia,zhou2024webarena,li2024seedbench,yu2024mmvet}. The task in \autoref{fig:realistic_challenges} illustrates this setting. The agent receives an image of a hazardous construction scene and a reporting-style instruction. 
Inside one step, actions that do not depend on each other, such as adding an image and updating bullets, can be executed concurrently. Across steps, operations that do depend on earlier results, such as annotating before inserting or creating the deck before saving, must follow the ground-truth order. These properties make simple string matching or linear-sequence scoring inadequate \cite{mialon2023gaia,zhou2024webarena}, and it is precisely what our benchmark aims to stress. We define two important concepts in what follows: \textbf{Multi-Hop} refers to workflows with more than one causally dependent step, where later actions consume artifacts produced earlier \cite{ye2025toolhop, he2024mintqa}; \textbf{Multi-Threaded} refers to order-independent tool calls executed within a single step under shared state, allowing safe parallelism while preserving cross-step causality (Mathematical definition in \autoref{def:multi-hop} \& \autoref{def:multi-threaded}).

\begin{table}[!t]
\centering
\scriptsize
\setlength{\tabcolsep}{2pt}
\renewcommand{\arraystretch}{1.05}
\vspace{1mm}

\newcommand{\cmark}{\textcolor{green!50!black}{\ding{51}}}
\newcommand{\xmark}{\textcolor{red}{\ding{55}}}

\newcommand{\rotcol}[1]{\rotatebox[origin=c]{90}{\parbox{1.4cm}{\centering #1}}}

\begin{tabular}{
    >{\centering\arraybackslash}m{2.4cm}
    >{\centering\arraybackslash}m{0.50cm}
    >{\centering\arraybackslash}m{0.65cm}
    *{6}{>{\centering\arraybackslash}m{0.59cm}}
}
\toprule
\rotcol{Benchmark} & \rotcol{Domains} & \rotcol{Tools} &
\rotcol{MCP Ecosystem} & \rotcol{Information Grounding} &
\rotcol{Fuzzy Task Description} & \rotcol{Multi-Hop \& Threaded} &
\rotcol{Multimodality} & \rotcol{Similarity Metric} \\
\cmidrule(lr){1-1}\cmidrule(lr){2-9}
ToolBench \cite{qin2023toollm} & 49 & 3451 &
\xmark & \xmark & \xmark & \xmark & \xmark & \xmark \\
BFCL v3 \cite{patil2025bfcl} & 8 & 24 &
\xmark & \xmark & \xmark & \xmark & \xmark & \xmark \\
$\tau$-Bench \cite{yao2025taubench} & 2 & 28 &
\xmark & \xmark & \xmark & \xmark & \xmark & \xmark \\
MCP-RADER \cite{gao2025mcpradar} & 9 & 42 &
\cmark & \xmark & \xmark & \xmark & \xmark & \xmark \\
MCPEval \cite{liu2025mcpeval} & 5 & 19 &
\cmark & \xmark & \xmark & \xmark & \xmark & \xmark \\
MCPToolBench++ \cite{fan2025mcptoolbenchpp} & 6 & 1500+ &
\cmark & \cmark & \xmark & \xmark & \xmark & \xmark \\
MCP-Atlas \cite{scale2025mcpatlas} & 40+ & 300+ &
\cmark & \cmark & \xmark & \xmark & \xmark & \xmark \\
MCP-Universe \cite{luo2025mcpuniverse} & 6 & 113 &
\cmark & \cmark & \cmark & \xmark & \xmark & \xmark \\
MCP-Bench \cite{wang2025mcpbench} & 28 & 250 &
\cmark & \cmark & \cmark & \cmark & \xmark & \xmark \\
M$^3$-Bench (Ours) & \textbf{27} & \textbf{232} &
\cmark & \cmark & \cmark & \cmark & \cmark & \cmark \\
\bottomrule
\end{tabular}
\caption{Comparisons to existing tool-using benchmarks (compact single-column).}
\label{tab:tool_benchmarks_compact}
\vspace{-8pt}
\end{table}

\autoref{tab:tool_benchmarks_compact} contrasts M$^3$-Bench with prior tool-use evaluations along 8 axes. 
\emph{Domains} and \emph{Tools} enumerate functional breadth and unique tool count. 
\emph{MCP ecosystem} marks whether the benchmark connect the agent directly to a set of production-grade, live MCP servers, and \emph{Information grounding} indicates that answers must be supported by evidence returned from tools. 
\emph{Fuzzy task description} denotes underspecified, natural instructions without a clear trajectory. Critically, \emph{Multi-Hop \& Threaded} captures cross-step causal dependencies together with safe within-step parallelism; \emph{Multimodality} requires joint image–text conditioning; and \emph{Similarity Metric} denotes large language models (LLM) free, similarity-aware alignment of predicted to reference calls (see \autoref{sec:m3-bench}). As discussed above, Multimodality and Multi-Hop/Threaded causality are central to realistic MCP workflows, and a Similarity-Based scorer is necessary to credit semantically correct calls. Most existing benchmarks are text-only and/or lack explicit multi-threaded causality, and none pair these with a similarity-aware alignment; in contrast, M$^3$-Bench satisfies all three while retaining the key advantages of prior benchmarks.

We align predicted and reference tool calls with a \emph{similarity-bucketed Hungarian alignment}: each call is serialized, embedded with a fixed sentence encoder, and one-to-one matched within tool-name buckets under weak/strong cosine thresholds. This gives deterministic, auditable correspondences without relying on an LLM judge for call-level scoring. 
On top of this alignment, we report a compact, recall-aware suite that separates \autoref{sec:m3-bench}. A small four-model judge ensemble is used for evaluating the overall quality of the trajectory to retain some of the advantages of the LLM judge feature. 
Besides, to standardize references, our experiments also provide a \emph{best trajectory} obtained via an Executor–Judge loop.

\vspace{-8pt}
\paragraph{Contributions.}
\begin{enumerate}
\item We present, to our knowledge, the first benchmark explicitly targeting \textbf{multimodal} MCP workflows.

\item Our repository provides an end-to-end pipeline for generating MCP best trajectories, with an optional lightweight human verification pass to enhance stability.


\item We introduce a structure-aware metric suite that aligns tool calls via bucketed Hungarian matching on high-dimensional feature similarities, yielding robust and auditable correspondences for effectively diagnosing distinct error modes.

\item We evaluate a wide range of mainstream MLLMs and conduct a comprehensive assessment across tasks, budgets, error types, and other factors.

\end{enumerate}

\section{Related Work}
\label{sec:Relate_Work}

\subsection{LLM Tool Use and Protocols}
Early studies showed that large language models (LLMs) can autonomously decide \emph{when/what/how} to call external tools via simple APIs \cite{schick2023toolformer,mialon2023gaia,andrews2025scaling,qin2023toollm, mo2025livemcpbench, jin2025twoheads, jin2025apeer, jin2025rewardprm, dong2026evaluating, dong2026labels}. Reliability improves with API-aware finetuning and retrieval \cite{patil2023gorilla}; instruction/data scaling expands tool repertoires and evaluations (ToolBench) \cite{qin2023toollm}; and hierarchical retrieval with self-reflection addresses large tool spaces and brittle protocols (AnyTool/AnyToolBench) \cite{du2024anytool}. Benchmarks targeting the \emph{decision to use a tool} and tool selection (MetaTool Benchmark) complement these directions \cite{huang2023metatool-benchmark}. To reduce integration friction across heterogeneous tools, the Model Context Protocol (MCP) standardizes interfaces \cite{anthropic2024mcp,mcp-spec-2025}.

\subsection{Benchmarks for Tool Use and MCP}
Beyond general agent evaluations such as AgentBench \cite{liu2023agentbench} and web-agent suites like WebArena and BrowserGym \cite{zhou2024webarena,lesellier2024browsergym}, recent MCP-native efforts emphasize realistic, multi-hop use over live servers: \emph{MCP-Bench} links LLMs to dozens of MCP servers and hundreds of tools \cite{wang2025mcpbench}; \emph{MCP-Universe} provides an end-to-end framework grounded in real-world servers and long-horizon tasks \cite{mcpuniverse-github,mcpuniverse-web}; \emph{MCPToolBench++} scales tool categories and tasks via marketplace-sourced servers \cite{fan2025mcptoolbenchpp}; and \emph{MCP-Atlas} curates a human-authored dataset and leaderboard for multi-tool workflows \cite{scale2025mcpatlas,scale2025mcpatlas-leaderboard}. Relative to these, we target \emph{multimodal} tasks and introduce a similarity-aware alignment with structure-sensitive diagnostics.

\begin{figure}[!tb]
    \centering
    \includegraphics[width=1\linewidth]{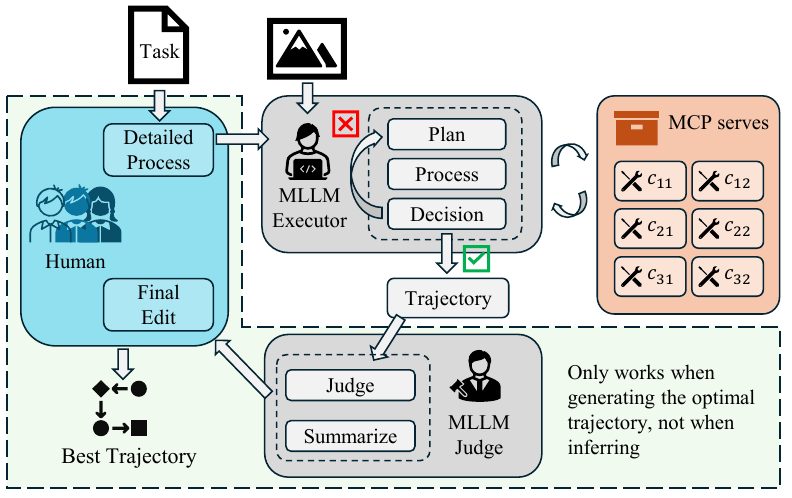}
    \caption{\footnotesize Benchmark trajectory generation and judging pipeline.
The Executor plans tools, executes calls with controlled multi-threaded execution, and summarizes; the Judge reconstructs an optimal trajectory and verifies correctness. where $c_{\ell,n_\ell}$ denotes the $n_\ell$-th tool call in step $\ell$.}
    \label{fig:pipeline_generation}
    \vspace{-8pt}
\end{figure}

\section{M$^3$-Bench}
\label{sec:m3-bench}

\subsection{Benchmark Construction}
\label{subsec:benchmark-trajectory}

We generate benchmark trajectories with an Executor–Judge loop that decouples tool execution from adjudication (\autoref{fig:pipeline_generation}). For each image–task pair, a human instruction fixes intent and permissible tools. Given the prompt, image, and registry, the Executor produces a stepwise trajectory; the Judge audits and consolidates it into a standardized \emph{optimal trajectory} for alignment. All prompts are in Appendix \autoref{prompt:prepare}.


\label{exp:plan}
Via system prompts, the Executor receives the full tool list (with docstrings) and outputs the planned tool set for the current step—\emph{selection only} to stabilize reasoning and keep the chain controllable. (See Appendix \href{prompt:prepare}{Prompt: Plan}.)

\vspace{-8pt}
\paragraph{Process}
\label{exp:process}
The Executor returns strictly valid JSON \texttt{tool\_calls} (each with \texttt{name} and \texttt{arguments}). A multi-hop workflow allows multi-threaded interaction with the MCP Ecosystem within a single step. Structured outputs are appended to the dialogue, and any file-image paths from MCP tools persist in a temporary workspace for later steps. (See Appendix \href{prompt:process}{Prompt: Process}.)

\vspace{-8pt}
\paragraph{Decision and Final Reply}
\label{exp:decision}
After each step, a binary prompt checks whether evidence suffices to conclude; if so, the Executor emits only the final natural-language answer. (See Appendix \href{prompt:stop}{Prompt: Decision}, \href{prompt:final}{Prompt: Final}.)

\vspace{-8pt}
\paragraph{Judge and Summarize}
\label{exp:judge_summarize}
The Judge consumes the initial prompt, image, and full Executor dialogue (including the tool-call chain). Trajectories with causal misordering or factual errors are rejected; otherwise, the Judge reconstructs the \emph{optimal trajectory}, removing failed/redundant calls and merging operations that could run concurrently. Human verifiers audit scope and logical consistency; the audited result becomes the benchmark reference. (See Appendix \href{prompt:judge}{Prompt: Judge}, \href{prompt:summarize}{Prompt: Summarize}.)

\begin{figure}[!tb]
    \centering
    \includegraphics[width=\linewidth]{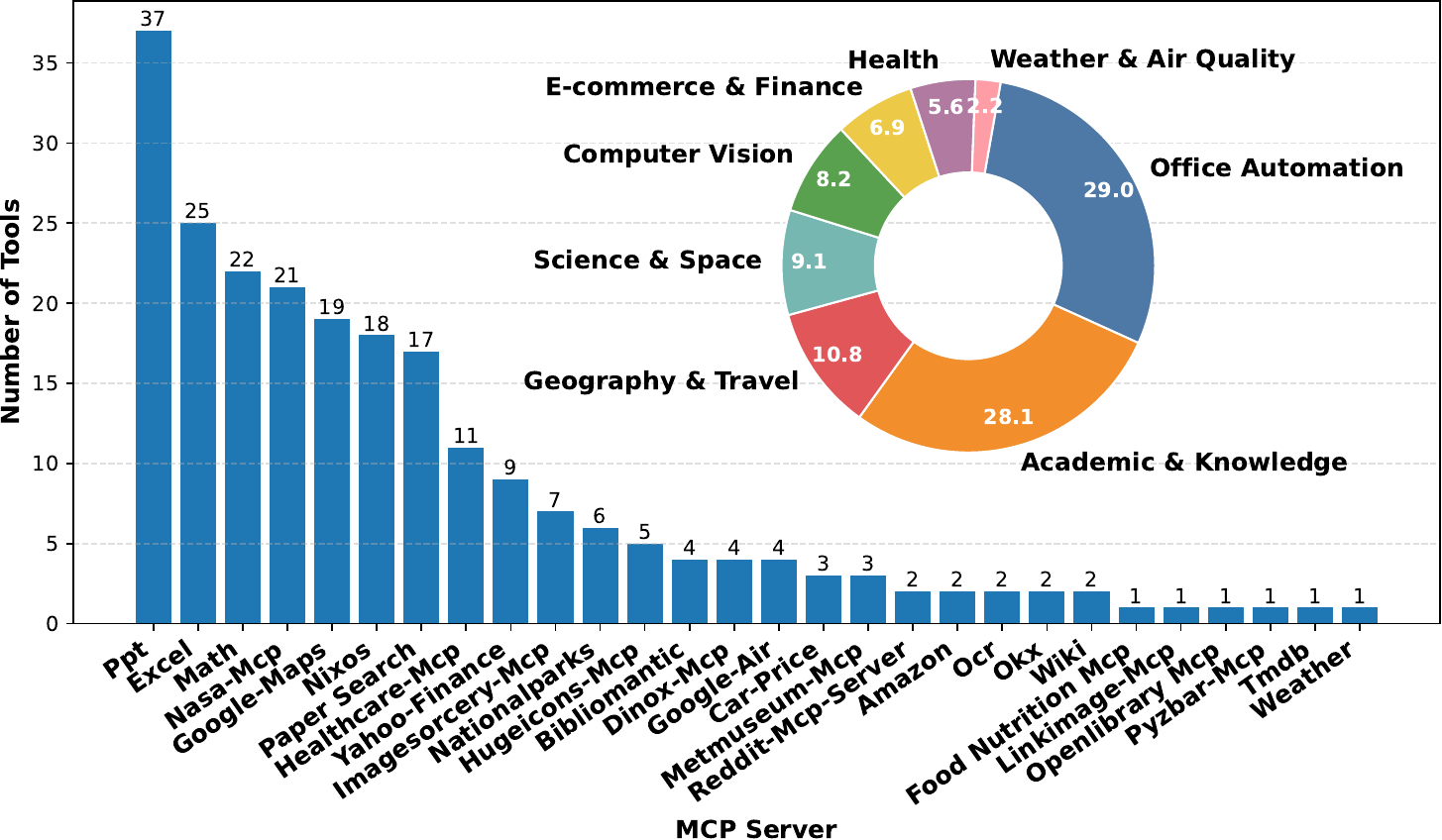}
    \caption{\footnotesize Overview of MCP server ecosystem used in the M$^3$-Bench. Category distribution of MCP servers; and tool distribution across servers. The inventory follows the MCP design and public specification.}
    \label{fig:mcp_tools_per_server}
     \vspace{-8pt}
\end{figure}

\subsection{Task Distribution}
\label{subsec:task-distribution}
We construct a curated suite of 28 multimodal MCP tasks; detailed per-task goals and input–output constraints are provided in Appendix \autoref{subsec:task-descriptions}, and representative full trajectories appear in \autoref{sec:trajectory-case-study}. In total, the evaluation set contains 211 trajectories, whose reference trajectories comprise 644 steps and 1337 MCP tool calls.

\subsection{MCP Ecosystem}
\label{subsec:mcp-ecosystem}
To keep trajectories controllable and reproducible, we curate a compact set of MCP servers and tools with minimal functional overlap.
We favor orthogonal, well-documented, and stable utilities, discarding redundant or unstable options to reduce confounds during evaluation.
Our benchmark spans 232 representative tools across 27 MCP servers (\autoref{fig:mcp_tools_per_server}(a)).
This diverse ecosystem covers Academic \& Knowledge, Science \& Space, Geography \& Travel, E-commerce \& Finance, Office Automation, Weather \& Air Quality, and Health, enabling broad capability coverage in M$^3$-Bench (see \autoref{fig:mcp_tools_per_server}(b)).

\section{Metrics}
\label{metrics}
In \autoref{subsec:mathematical}, we first establish the mathematical definitions and foundations for our metrics. Building on these theoretical bases, we then construct the Detection Metrics \autoref{subsec:detection-metrics}, Trajectory Alignment Metrics \autoref{subsec:trajectory-alignment}, and LLM Judges Metrics \autoref{subsec:llm-judge}.

\subsection{Mathematical foundations}
\label{subsec:mathematical}
A trajectory is a DAG ordered by turns (\emph{steps}). Denote $L$ as the number of steps in a trajectory $\mathcal{T}$. The ground-truth $\mathcal{T}^\star$ and predicted trajectories $\widehat{\mathcal{T}}$ are:
$\mathcal{T}^\star=\{S^\star_{1},\ldots,S^\star_{L^\star}\}$,
$\widehat{\mathcal{T}}=\{\widehat{S}_{1},\ldots,\widehat{S}_{\widehat{L}}\}$, 
\noindent where each step $S$ enumerates its calls $c$,
$S^\star_{\ell}=\{\,c^\star_{\ell,1},\ldots,c^\star_{\ell,n^\star_\ell}\,\}$,
$\widehat{S}_{\hat{\ell}}=\{\,\hat{c}_{\hat{\ell},1},\ldots,\hat{c}_{\hat{\ell},n_{\hat{\ell}}}\,\}. $
\noindent where $\ell$ is the step of the trajectory, so $n^\star_\ell$ and $n_{\hat{\ell}}$ are per-step call counts. A call is $c=(\mathrm{\texttt{tool}}(c),\mathrm{\texttt{args}}(c)),$ 
where $\mathrm{\texttt{tool}}(\cdot)$ is the categorical tool identifier and $\mathrm{\texttt{args}}(\cdot)$ serializes all deterministic inputs.

\begin{definition}[Multi-Hop]\label{def:multi-hop}
Let $\mathcal{T}=\{S_1,\dots,S_L\}$ be a tool-use trajectory with steps $S_\ell=\{c_{\ell,1},\dots,c_{\ell,n_\ell}\}$ and inter-call dependency edges across steps. 
We call $\mathcal{T}$ \emph{multi-hop} if $L\ge2$ and there exists a dependency $(c_{i,a}\!\to\! c_{j,b})$ with $i<j$, where $a,b \in \mathbb{N}$ and $a \in [1, n_i],\ b \in [1, n_j]$.
\end{definition}

\begin{definition}[Multi-Threaded]\label{def:multi-threaded}
A step $S_\ell$ is \emph{multi-threaded} if it contains at least two calls $c_{\ell,a}\neq c_{\ell,b}$ that are order-independent given the shared state at depth $\ell$ (no dependency edge between them). 
\end{definition}

These definitions make explicit that a trajectory is a structured object with cross-step causal edges and within-step parallelism. A general option is to delegate this alignment to an LLM judge, but it introduces (i) \emph{circularity/lineage bias}, where judges trained like the systems under test tend to over-credit them; (ii) susceptibility to \emph{prompt gaming}, as long rationales or CoT responses can inflate scores even when the underlying tool use is wrong; (iii) \emph{weak argument verification}, since judges rarely check whether argument objects are semantically correct or whether order-invariant sets are handled; (iv) \emph{knowledge substitution}, where missing tool evidence is replaced by the model’s own world knowledge; and (v) \emph{opacity}, as scores lack auditable, one-to-one correspondences between predicted and reference calls. Real MCP trajectories exacerbate these issues: steps often contain \emph{unordered} sets of calls, later calls can \emph{self-correct} earlier errors, and partial success is common. Consequently, the alignment primitive must be deterministic, auditable, and robust to surface-form variation.

\paragraph{Similarity-Bucketed Hungarian Alignment}
\label{subsubsec:embed}
To address this problem, we introduce a similarity computation method based on a text encoder.
Each call is flattened to a canonical text string by a serialization function $\phi(\cdot)$ that preserves tool name, required arguments and order-invariant sets.
We embed serialized calls with a sentence encoder $\mathcal{E}$ and compute cosine similarity:
$S_{ij}=\frac{\langle \mathbf{x}^\star_i,\widehat{\mathbf{x}}_j\rangle}{\lVert \mathbf{x}^\star_i\rVert_2\,\lVert \widehat{\mathbf{x}}_j\rVert_2}\in[-1,1]$,
where 
$\mathbf{x}^\star_i=\mathcal{E}\!\big(\phi(c^\star_i)\big)$ and
$\widehat{\mathbf{x}}_j=\mathcal{E}\!\big(\phi(\hat{c}_j)\big)$.
    
We align with a tool-specific, step-agnostic assignment that yields auditable one-to-one correspondences. Concretely, we serialize every call into a canonical text signature and embed it with the sentence encoder SentenceTransformers \cite{reimers2019sentencebert} from \href{subsubsec:embed}{Textual Encoding and Similarity}, giving a similarity matrix $S$. For each tool $k$, we form the bucketed index sets
$\mathcal{I}_k=\{\,i\mid \mathrm{\texttt{tool}}(c^\star_i)=k\,\}$,
$\widehat{\mathcal{J}}_k=\{\,j\mid \mathrm{\texttt{tool}}(\hat{c}_j)=k\,\}$,    
so that only same-tool pairs are considered. We convert similarity to cost with a weak screening threshold $\tau_{\mathrm{weak}}$ to forbid low-similarity edges:
\begin{equation}
\label{eq:cost}
C^{(k)}_{rc}=
\begin{cases}
1-S^{(k)}_{rc}, & S^{(k)}_{rc}\ge \tau_{\mathrm{weak}},\\
\lambda_{\mathrm{pen}}, & S^{(k)}_{rc}<\tau_{\mathrm{weak}},
\end{cases}
\end{equation}
and solve a linear assignment in each bucket via the Hungarian (Kuhn--Munkres) algorithm  $\min_{\Pi^{(k)}\in\mathcal{P}}\ \langle C^{(k)},\Pi^{(k)}\rangle$
retaining only feasible pairs above the weak threshold to form the global match set:
$M=\{(i,j,S_{ij})\ \mid\ S_{ij}\ge\tau_{\mathrm{weak}}\}.$

\paragraph{Properties and rationale.}
Tool bucketing prevents cross-tool credit; step agnosticism allows \emph{early-wrong, later-correct} fixes to match without forcing same-step alignment; and similarity gating recognizes paraphrase and order-invariant arguments without inflating coverage. Together, these choices deliver: (i) deterministic, auditable correspondences in place of LLM judge; (ii) invariance to within-step call permutations (sets are unordered); and (iii) faithful scoring under \emph{partial success}, misaligned or missing calls simply remain unmatched, while correctly recovered calls still receive credit. This alignment thus targets exactly the failure modes we observe in MCP trajectories while satisfying the three requirements in \autoref{subsubsec:embed}.

\begin{figure}[!tb]
    \centering
    \includegraphics[width=1\linewidth]{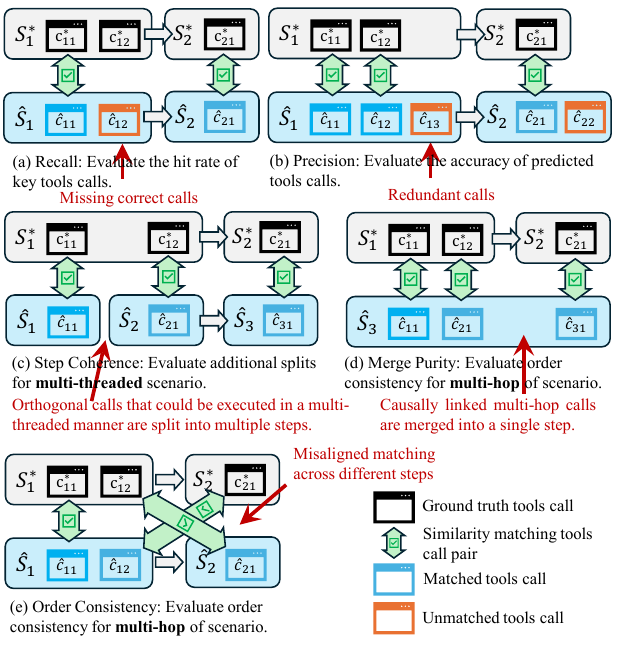}
    \caption{\footnotesize Overview of our structure-sensitive metrics. where $S^\star_{\ell}$ and $\widehat{S}_{\hat{\ell}}$ denote the sets of calls in the ground-truth step $\ell$ and the predicted step $\hat{\ell}$, respectively, and $c^\star_{\ell,n^\star_\ell}$ and $\hat{c}_{\hat{\ell},n_{\hat{\ell}}}$ denote the $n^\star_\ell$-th ground-truth call in step $\ell$ and the $n_{\hat{\ell}}$-th predicted call in step $\hat{\ell}$.}
    \label{fig:metrics_overview}
     \vspace{-8pt}
\end{figure}

\subsection{Detection Metrics}
\label{subsec:detection-metrics}
Based on the above mathematical foundations, we introduce detection-based metrics:
\begin{enumerate}
    \item Systems may call incorrect tools or omit required ones, motivating \textbf{Recall};
    \item MLLM may make unnecessary calls, resulting in additional waste of resources \textbf{Precision}.
\end{enumerate}

\vspace{-8pt}
\paragraph{Recall and Precision.}\ (\autoref{fig:metrics_overview} (a)-(b))
These metrics quantify, respectively, how many ground-truth calls are successfully matched (coverage) and how many predicted calls are valid (false-positive control). Given per-sample match sets ${M_m}$ with counts $N_{\mathrm{gt},m}$ and $N_{\mathrm{pred},m}$, we define
$\mathrm{Recall}=\frac{\sum_m |M_m|}{\sum_m N_{\mathrm{gt},m}}$, 
$\mathrm{Precision}=\frac{\sum_m |M_m|}{\sum_m N_{\mathrm{pred},m}}$,
where $|M_m|$ counts matched pairs for sample $m$, $N_{\mathrm{gt}}=\sum_{\ell}n^\star_\ell$ and $N_{\mathrm{pred}}=\sum_{\hat\ell}n_{\hat\ell}$ be the total numbers of calls in $\mathcal{T}^\star$ and $\widehat{\mathcal{T}}$, respectively.


\subsection{Trajectory Alignment Metrics}
\label{subsec:trajectory-alignment}
Next, we introduce a metric to quantify the degree of Trajectory Alignment.
\begin{enumerate}
    \item MLLM may select the correct tool but supply semantically incorrect arguments, requiring an \textbf{Average Similarity} measure over serialized calls;
    \item MLLM may unnecessarily split operations that could run in a single step in a \textbf{multi-threaded} scenario, captured by \textbf{Step Coherence};
    \item MLLM may collapse causally distinct multi-hop subgoals into one step in a \textbf{multi-hop} scenario, quantified by \textbf{Merge Purity};
    \item MLLM may invert the execution order of causally dependent steps in a \textbf{multi-hop} scenario, detected by \textbf{Order Consistency}.
\end{enumerate}

\paragraph{Argument Similarity.}
We summarize the semantic faithfulness of matched arguments as
\vspace{-8pt}
\begin{equation}
\mathrm{ArgSim}
= \frac{\sum_{(i,j)\in M,\, S_{ij}\ge\tau_{\mathrm{strong}}} S_{ij}}
       {|\{(i,j)\in M : S_{ij}\ge\tau_{\mathrm{strong}}\}|},
\end{equation}
where $\tau_{\mathrm{strong}}$ controls alignment confidence.
\vspace{-8pt}

\begin{table*}[!htb]
    \centering
    \footnotesize
    \setlength{\tabcolsep}{6pt}
    \begin{tabular}{>{\arraybackslash}m{3.7cm}*{9}{>{\centering\arraybackslash}m{1.0cm}}}
    \toprule
    \multirow{3}{*}{\centering Model}
    & \multicolumn{2}{c}{Detection}
    & \multicolumn{4}{c}{Trajectory Alignment}
    & \multicolumn{2}{c}{LLM Judge}
    & \multirow{3}{*}{\centering\shortstack{Average\\Score}} \\
    \cmidrule(lr){2-3} \cmidrule(lr){4-7} \cmidrule(lr){8-9}
    & Recall & Precision
    & \shortstack{Argument\\Similarity}
    & \shortstack{Step\\Coherence}
    & \shortstack{Order\\Consistency}
    & \shortstack{Merge\\Purity}
    & \shortstack{Task\\Completion}
    & \shortstack{Information\\Grounding}
    & \\
    \midrule
    \raisebox{-0.2\height}{\includegraphics[height=0.9em]{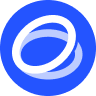}}\hspace{3pt}
    GLM 4.5v \cite{hong2025glm45v}
        & 0.013 & 0.023 & 0.011 & 0.012 & 0.000 & 0.013 & 0.285 & 0.130 & 0.029 \\
    \raisebox{-0.2\height}{\includegraphics[height=0.9em]{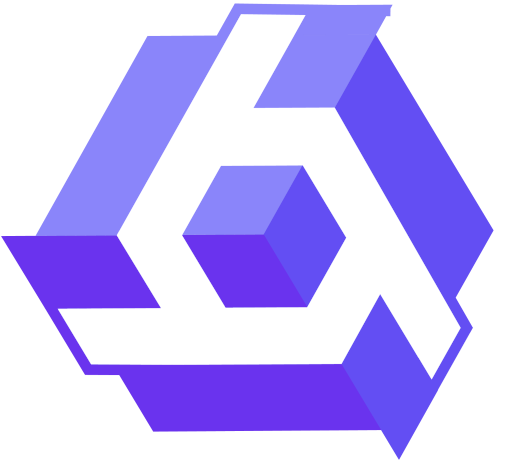}}\hspace{3pt}
    Qwen2.5-VL-72B \cite{bai2025qwen25vl}
        & 0.143 & 0.073 & 0.125 & 0.125 & 0.022 & 0.112 & 0.303 & 0.388 & 0.141 \\
    \raisebox{-0.2\height}{\includegraphics[height=0.9em]{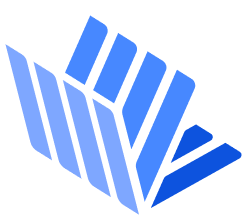}}\hspace{3pt}
    InternVL 3.5 \cite{wang2025internvl35}
        & 0.224 & 0.143 & 0.193 & 0.191 & 0.048 & 0.161 & 0.325 & 0.298 & 0.179 \\
    \raisebox{-0.2\height}{\includegraphics[height=0.9em]{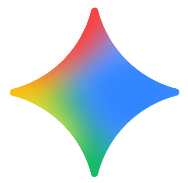}}\hspace{3pt}
    Gemini 2.5 Flash Lite \cite{comanici2025gemini25}
        & 0.200 & 0.173 & 0.168 & 0.179 & 0.030 & 0.152 & 0.299 & 0.241  & 0.180 \\
    \raisebox{-0.2\height}{\includegraphics[height=0.9em]{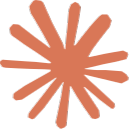}}\hspace{3pt}
    Claude 4.5 Haiku \cite{anthropic2025haiku45}
        & 0.193 & 0.480 & 0.166 & 0.190 & 0.053 & 0.108 & 0.412 & 0.246 & 0.205 \\
    \raisebox{-0.2\height}{\includegraphics[height=0.9em]{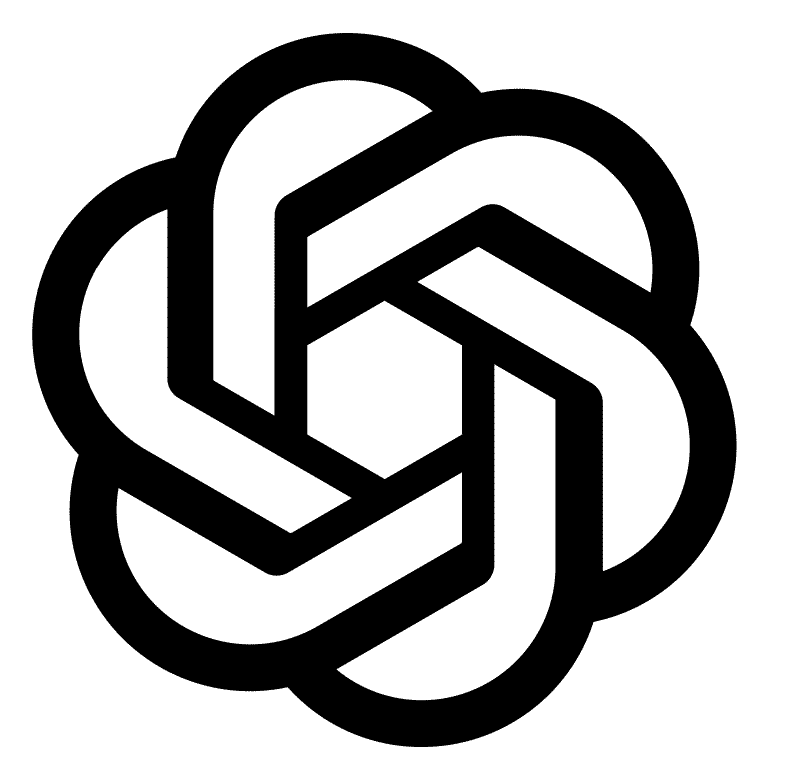}}\hspace{3pt}
    GPT-5 Nano \cite{openai2025gpt5nano}
        & 0.287 & 0.103 & 0.276 & 0.228 & 0.045 & 0.258 & 0.551 & 0.529 & 0.247 \\
    \raisebox{-0.2\height}{\includegraphics[height=0.6em]{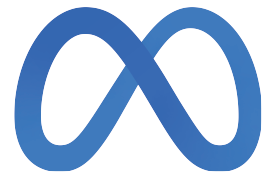}}\hspace{3pt}
    Llama-4-Scout-17B16E \cite{meta2025llama4scout17b16e}
        & 0.367 & 0.176 & 0.325 & 0.286 & 0.068 & 0.290 & 0.367 & 0.335 & 0.264 \\
    \raisebox{-0.2\height}{\includegraphics[height=0.9em]{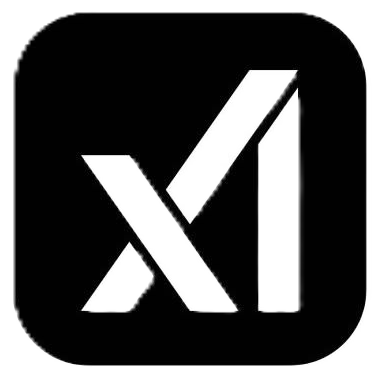}}\hspace{3pt}
    Grok-4 Fast \cite{xai2025grok4fast}
        & 0.343 & 0.359 & 0.309 & 0.284 & 0.122 & 0.289 & 0.501 & 0.378 & 0.298 \\
    \raisebox{-0.2\height}{\includegraphics[height=0.9em]{figure/icon/claude.png}}\hspace{3pt}
    Claude 4.5 Sonnet \cite{anthropic2025sonnet45}
        & 0.395 & 0.361 & 0.354 & 0.389 & 0.127 & 0.257 & 0.535 & 0.451 & 0.333 \\
    \raisebox{-0.2\height}{\includegraphics[height=0.9em]{figure/icon/gemini.png}}\hspace{3pt}
    Gemini 2.5 Flash \cite{comanici2025gemini25}
        & 0.454 & 0.397 & 0.409 & 0.357 & 0.211 & 0.412 & 0.530 & 0.474 & 0.388 \\
    \raisebox{-0.2\height}{\includegraphics[height=0.9em]{figure/icon/gpt.png}}\hspace{3pt}
    GPT-5 Mini \cite{openai2025gpt5mini}
        & 0.481 & 0.238 & 0.462 & 0.392 & 0.199 & 0.360 & 0.652 & 0.633 & 0.395 \\
    \raisebox{-0.2\height}{\includegraphics[height=0.9em]{figure/icon/grok.png}}\hspace{3pt}
    Grok 4 (0709) \cite{xai2025grok4}
        & 0.465 & 0.387 & 0.408 & 0.415 & 0.177 & 0.390 & 0.664 & 0.638 & 0.411 \\
    \raisebox{-0.2\height}{\includegraphics[height=0.9em]{figure/icon/gemini.png}}\hspace{3pt}
    Gemini 2.5 Pro \cite{comanici2025gemini25}
        & 0.501 & 0.449 & 0.433 & 0.464 & 0.249 & 0.424 & 0.520 & 0.438 & 0.423 \\
    \raisebox{-0.2\height}{\includegraphics[height=0.9em]{figure/icon/gpt.png}}\hspace{3pt}
    GPT-5 \cite{openai2025gpt5systemcard}
        & 0.627 & 0.264 & 0.583 & 0.502 & 0.290 & 0.453 & 0.671 & 0.657 & 0.482 \\
    \bottomrule
    \end{tabular}
    \caption{Cross-Model Evaluation of Multimodal MCP Tool-Calling Trajectories: Detection, Trajectory Alignment, and LLM-Judge Metrics, The average score results correspond to \autoref{fig:metrics_mllm_step_eval}.}
    \label{tab:mcp_multimodal_results}
    \vspace{-10pt}
    \end{table*}

\paragraph{Step Coherence.} (\autoref{fig:metrics_overview} (c))
This metric penalizes scattering matches from one ground-truth step across multiple predicted steps, reflecting unnecessary splits of operations that could have been executed within a single \textbf{multi-threaded} step. For any call $c$, let $\mathrm{gt_{st}}(c)$ and $\mathrm{pred_{st}}(c)$ denote its step indices in the ground-truth and predicted trajectories. For each ground-truth step $\ell$, collect the predicted step indices touched by its matches: 
$\mathcal{P}(\ell)=\{\ \mathrm{pred_{st}}(\hat{c}_j)\ \mid\ (i,j)\in M, \mathrm{gt_{st}}(c^\star_i)=\ell\ \}$.Then we define the per-step coherence and a similarity-weighted global average as

\vspace{-8pt}
\begin{equation}
\begin{aligned}
\label{eq:frag}
&\mathrm{sc}(\ell)=
\begin{cases}
1, & |\mathcal{P}(\ell)|\le 1,\\[2pt]
\dfrac{1}{|\mathcal{P}(\ell)|}, & \text{otherwise},
\end{cases}
\\
&\mathrm{StepCoh}=\frac{\sum_{\ell} w_\ell\,\mathrm{sc}(\ell)}{\sum_{\ell} w_\ell},
\end{aligned}
\end{equation}

\noindent where $w_\ell=\sum_{(i,j)\in M}[\mathrm{gt_{st}}(c^\star_i)=\ell]$ counts matches originating from ground-truth step $\ell$.

\paragraph{Merge Purity.} (\autoref{fig:metrics_overview} (d))
This metric is used to evaluate whether, in a \textbf{multi-hop} scenario, different causally related steps are incorrectly merged. For total alignment weight between ground-truth step $a$ and predicted step $b$:

\vspace{-8pt}
\begin{equation}
\begin{aligned}
\label{eq:merge-W}
W_{ab}=\sum_{(i,j)\in M} & [\mathrm{gt_{st}}(c^\star_i)=a]\ [\mathrm{pred_{st}}(\hat{c}_j)=b]\ S_{ij},
\end{aligned}
\end{equation}

\noindent where $[\cdot]$ denotes the Iverson bracket (indicator function), equal to (1) when its condition is true and (0) otherwise. Define column sums $S_b=\sum_a W_{ab}$, total mass $S=\sum_b S_b$, and column prior $P(b)=S_b/S$. The conditional distribution of ground-truth steps given a predicted step is $q_{a\mid b}=W_{ab}/S_b$ for active columns with $S_b>0$. The normalized purity is:

\vspace{-8pt}
\begin{equation}
\begin{aligned}
\label{eq:merge-purity}
&\mathrm{MergePur}  = 1 - \frac{H(G\mid P)}{\log G_{\mathrm{act}}} \\
&H(G\mid P)  =\sum_{b: S_b>0} P(b)\Big(-\sum_{a: W_{ab}>0} q_{a\mid b}\log q_{a\mid b}\Big),
\end{aligned}
\end{equation}

\noindent where $G_{\mathrm{act}}=|\{a:\sum_b W_{ab}>0\}|$ counts active ground-truth steps, $H(\cdot)$ uses natural logarithm, and larger values indicate predicted steps concentrating mass on single ground-truth steps (i.e., less over-merging);
$G$ denotes the random variable of ground-truth step indices and $P$ denotes the random variable of predicted step indices; $G_{\mathrm{act}}=|\{a:\sum_b W_{ab}>0\}|$ is the number of active ground-truth steps. The normalization by $\log G_{\mathrm{act}}$ maps $[0,\log G_{\mathrm{act}}]$ to $[0,1]$.

\vspace{-8pt}
\paragraph{Order Consistency.} (\autoref{fig:metrics_overview} (e))
This metric is used to evaluate whether, in a \textbf{multi-hop} scenario, the order of causally related calls is incorrectly swapped, capturing order consistency across steps. Let each match $(i,j)\in M$ be represented by its step pair $(a,b)=(\mathrm{gt_{st}}(c^\star_i),\mathrm{pred_{st}}(\hat{c}_j))$. Over all unordered pairs of matches that are comparable in both axes, we compute:

\vspace{-8pt}
\begin{equation}
\begin{aligned}
\label{eq:order-consistency}
\mathrm{OrdCons}=1-\frac{1}{Q}\sum_{(u,v)}\big[(a_u-a_v)(b_u-b_v)<0\big],
\end{aligned}
\end{equation}
\vspace{-5pt}

\noindent where the sum runs over pairs with $a_u\neq a_v$ and $b_u\neq b_v$, $Q$ is the number of such pairs, and the indicator records inversions between the ground-truth and predicted step orders; higher values indicate fewer inversions.

To penalize degenerate cases with very few matches (weak systems that match only a few easy MCP tools from inflating structure-aware metrics, leading to a very small denominator in the metric calculation, and thus an inflated outcome), we weight per-sample metrics by their recall. Let the per-sample recall be
$ r_m \,=\, \frac{|M_m|}{N_{\mathrm{gt},m}}\,$. For any per-sample metric $F_m\in\{\mathrm{StepCoh},\allowbreak\ \mathrm{MergePur},\allowbreak\ \mathrm{OrdCons}\}$, we report a recall-covered score:
$F^{\mathrm{cov}} \,=\, \frac{\sum_{m} N_{\mathrm{gt},m}\, r_m\, F_m}{\sum_{m} N_{\mathrm{gt},m}}\,$.

\subsection{LLMs Judges Metrics}
\label{subsec:llm-judge}
To leverage the advantages of large language models (LLMs) in complex logical reasoning, we introduce a metric based on LLMs Judges to assess:
\begin{enumerate}
    \item Whether the MLLM achieves \textbf{Task Completion};
    \item Whether the answer provided by the MLLM possesses MCP \textbf{Information Grounding}.
\end{enumerate}

\vspace{-8pt}
\paragraph{Task Completion.}\label{subsubsec:taskcompletion}
For each instance $m$, a fixed set of four LLMs Judges $\mathcal{J}$ produces scalar scores $\{ s^{(j)}_m \}_{j\in\mathcal{J}}\subset[0,1]$. We compute a trimmed mean that discards the highest and lowest scores and averages the remaining two (Prompt: Appendix \autoref{prompt:task_completion}):

\begin{equation}
\begin{aligned}
\label{eq:taskcompletion}
&\mathrm{TaskComp}_m \,=\, \tfrac{1}{2} \! \sum_{j\in\mathcal{J}\setminus\{j_{\min},\,j_{\max}\}} s^{(j)}_m\,, \\
&j_{\min}=\arg\min_{j\in\mathcal{J}} s^{(j)}_m,\ j_{\max}=\arg\max_{j\in\mathcal{J}} s^{(j)}_m.
\end{aligned}
\end{equation}

We report $\mathrm{TaskComp}$ (Task Completion) as the mean of per-instance scores over the evaluation set. The judge pool includes strong large language models (\eg, GPT, Gemini, DeepSeek, Grok).

\vspace{-8pt}
\paragraph{Information Grounding.}\label{subsubsec:information_grounding}
Judges are additionally shown both the ground-truth and predicted trajectories and emit step-level groundedness scores. For instance $m$ with ground-truth steps indexed by $\ell^\star$ and $\hat\ell$, each judge $j\in\mathcal{J}$ outputs $g^{(j)}_{m,\ell^\star,\hat\ell}\in[0,1]$. The calculation of $\mathrm{InfoGrnd}_{m,\ell^\star,\hat\ell}$ follows \autoref{eq:taskcompletion} (Prompt: Appendix \autoref{prompt:information_grounding}).


    \begin{figure*}[!htb]
        \centering
        \includegraphics[width=1\linewidth]{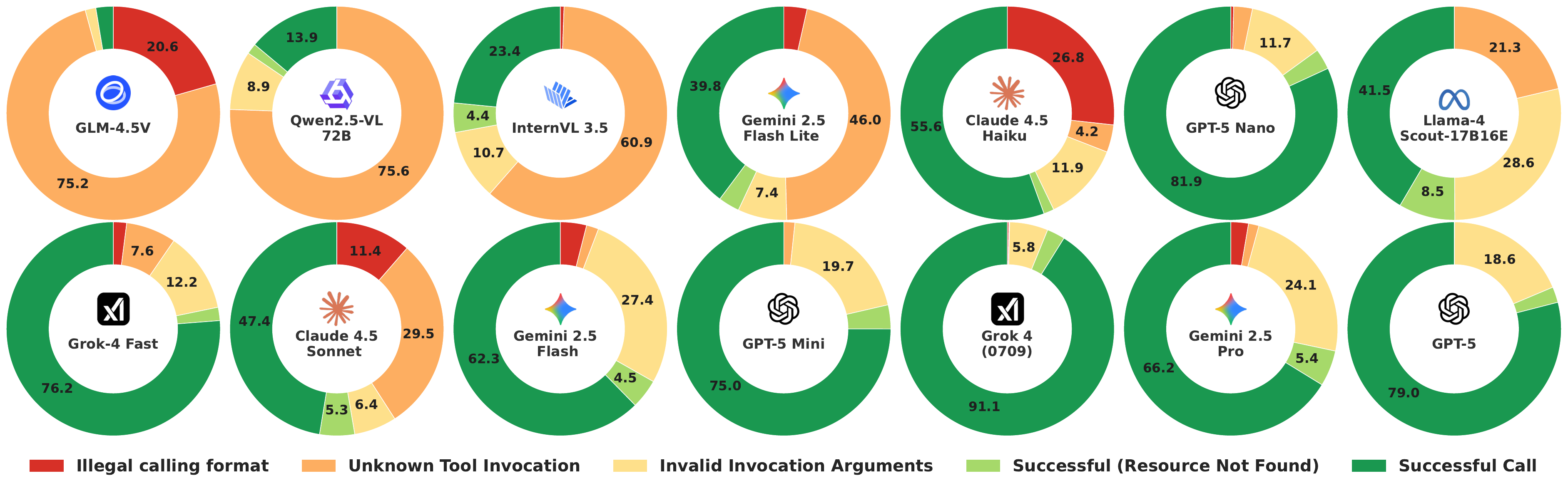}
        \caption{Per-call outcome distributions across models when replaying predicted MCP trajectories through the MCP interpreter.}
        \label{fig:call_eval_pies}
         \vspace{-10pt}
    \end{figure*}


        \begin{table}[!t]
    \centering
    \footnotesize
    \setlength{\tabcolsep}{6pt}
    \begin{tabular}{lcc}
    \toprule
    Model & Input Tokens & Output Tokens \\
    \midrule
    \raisebox{-0.2\height}{\includegraphics[height=0.9em]{figure/icon/qwen.png}}\hspace{3pt}Qwen2.5-VL-72B
    & 0.362 & 0.008 \\
    \raisebox{-0.2\height}{\includegraphics[height=0.9em]{figure/icon/internvl.png}}\hspace{3pt}InternVL 3.5
    & 0.069 & 0.002 \\
    \raisebox{-0.2\height}{\includegraphics[height=0.9em]{figure/icon/claude.png}}\hspace{3pt}Claude 4.5 Haiku
    & 0.118 & 0.007 \\
    \raisebox{-0.2\height}{\includegraphics[height=0.9em]{figure/icon/claude.png}}\hspace{3pt}Claude 4.5 Sonnet
    & 0.408 & 0.029 \\
    \raisebox{-0.2\height}{\includegraphics[height=0.9em]{figure/icon/gpt.png}}\hspace{3pt}GPT-5 Nano
    & 0.002 & 0.001 \\
    \raisebox{-0.2\height}{\includegraphics[height=0.6em]{figure/icon/llama.png}}\hspace{3pt}Llama-4-Scout-17B16E
    & 0.345 & 0.008 \\
    \raisebox{-0.2\height}{\includegraphics[height=0.9em]{figure/icon/grok.png}}\hspace{3pt}Grok-4 Fast
    & 0.210 & 0.028 \\
    \raisebox{-0.2\height}{\includegraphics[height=0.9em]{figure/icon/gemini.png}}\hspace{3pt}Gemini 2.5 Flash Lite
    & 0.477 & 0.010 \\
    \raisebox{-0.2\height}{\includegraphics[height=0.9em]{figure/icon/gemini.png}}\hspace{3pt}Gemini 2.5 Flash
    & 0.621 & 0.032 \\
    \raisebox{-0.2\height}{\includegraphics[height=0.9em]{figure/icon/gpt.png}}\hspace{3pt}GPT-5 Mini
    & 0.125 & 0.014 \\
    \raisebox{-0.2\height}{\includegraphics[height=0.9em]{figure/icon/grok.png}}\hspace{3pt}Grok 4 (0709)
    & 0.132 & 0.027 \\
    \raisebox{-0.2\height}{\includegraphics[height=0.9em]{figure/icon/gemini.png}}\hspace{3pt}Gemini 2.5 Pro
    & 0.606 & 0.031 \\
    \raisebox{-0.2\height}{\includegraphics[height=0.9em]{figure/icon/gpt.png}}\hspace{3pt}GPT-5
    & 0.252 & 0.037 \\
    \bottomrule
    \end{tabular}
    \caption{Average token cost per task for each MLLM in millions.}
    \label{tab:token_counts}
    \vspace{-8pt}
    \end{table}

\section{Experiment}
\label{sec:Experiment}

\subsection{Evaluation Results}

    
    \begin{table}[!t]
    \centering
    \footnotesize
    \setlength{\tabcolsep}{6pt}
    \begin{tabular}{lcc}
    \toprule
    Model & Avg. Steps & Avg. Tool Calls \\
    \midrule
    \raisebox{-0.2\height}{\includegraphics[height=0.9em]{figure/icon/glm.png}}\hspace{3pt}GLM 4.5v
    & 1.91 & 3.84 \\
    \raisebox{-0.2\height}{\includegraphics[height=0.9em]{figure/icon/qwen.png}}\hspace{3pt}Qwen2.5-VL-72B
    & 4.29 & 12.30 \\
    \raisebox{-0.2\height}{\includegraphics[height=0.9em]{figure/icon/internvl.png}}\hspace{3pt}InternVL 3.5
    & 3.65 & 9.57 \\
    \raisebox{-0.2\height}{\includegraphics[height=0.9em]{figure/icon/claude.png}}\hspace{3pt}Claude 4.5 Haiku
    & 0.80 & 2.73 \\
    \raisebox{-0.2\height}{\includegraphics[height=0.9em]{figure/icon/claude.png}}\hspace{3pt}Claude 4.5 Sonnet
    & 2.74 & 6.99 \\
    \raisebox{-0.2\height}{\includegraphics[height=0.9em]{figure/icon/gpt.png}}\hspace{3pt}GPT-5 Nano
    & 5.13 & 17.97 \\
    \raisebox{-0.2\height}{\includegraphics[height=0.6em]{figure/icon/llama.png}}\hspace{3pt}Llama-4-Scout-17B16E
    & 4.72 & 13.26 \\
    \raisebox{-0.2\height}{\includegraphics[height=0.9em]{figure/icon/grok.png}}\hspace{3pt}Grok-4 Fast
    & 2.91 & 5.93 \\
    \raisebox{-0.2\height}{\includegraphics[height=0.9em]{figure/icon/gemini.png}}\hspace{3pt}Gemini 2.5 Flash Lite
    & 2.73 & 7.06 \\
    \raisebox{-0.2\height}{\includegraphics[height=0.9em]{figure/icon/gemini.png}}\hspace{3pt}Gemini 2.5 Flash
    & 4.04 & 7.12 \\
    \raisebox{-0.2\height}{\includegraphics[height=0.9em]{figure/icon/gpt.png}}\hspace{3pt}GPT-5 Mini
    & 3.72 & 12.56 \\
    \raisebox{-0.2\height}{\includegraphics[height=0.9em]{figure/icon/grok.png}}\hspace{3pt}Grok 4 (0709)
    & 3.21 & 7.64 \\
    \raisebox{-0.2\height}{\includegraphics[height=0.9em]{figure/icon/gemini.png}}\hspace{3pt}Gemini 2.5 Pro
    & 3.04 & 6.95 \\
    \raisebox{-0.2\height}{\includegraphics[height=0.9em]{figure/icon/gpt.png}}\hspace{3pt}GPT-5
    & 3.55 & 14.77 \\
    \bottomrule
    \end{tabular}
    \caption{Average interaction steps and tool calls per task across models on our benchmark.}
    \label{tab:avg_rounds_calls}
    \vspace{-8pt}
    \end{table}

As shown in \autoref{tab:mcp_multimodal_results}, we evaluate SOTA MLLMs on M$^3$-Bench. The three flagship families, \texttt{GPT-5} \cite{openai2025gpt5systemcard}, \texttt{Gemini}, and \texttt{Grok}, form the leading cluster under our coverage-weighted alignment and judge signals. \texttt{GPT-5} \cite{openai2025gpt5systemcard} attains the highest Average Score (0.482) with strong Recall (0.627) and high Argument Similarity (0.583), and it also tops all structure-aware metrics: its multi-threaded Step Coherence (0.502) and multi-hop sensitive Order Consistency / Merge Purity (0.290 / 0.453) indicate that it preserves both within-step parallel branches and cross-step merges in long tool graphs. \texttt{Gemini 2.5 Pro} \cite{comanici2025gemini25} (0.423) and \texttt{Grok 4 (0709)} \cite{xai2025grok4} (0.411) exhibit similarly balanced profiles.

In contrast, compact/latency-oriented variants tend to trade coverage and structural consistency for speed: for example, \texttt{Claude 4.5 Haiku} \cite{anthropic2025haiku45} achieves high Precision (0.480) but much lower Recall, and its structure metrics remain modest, pointing to short, locally correct trajectories that under-utilize multi-threaded branches and often break multi-hop dependencies. Finally, \texttt{GLM 4.5v} \cite{hong2025glm45v}, \texttt{Qwen2.5-VL-72B} \cite{bai2025qwen25vl}, and \texttt{InternVL 3.5} \cite{wang2025internvl35} lag notably on both coverage and structure, suggesting that MLLMs SFT objectives favor pure vision–language understanding over instruction following and trajectory planning.

\subsection{Per-Task Analysis}
\label{subsec:per-task-lines}

    \begin{figure}[!htb]
    \centering
    \includegraphics [width=\linewidth]{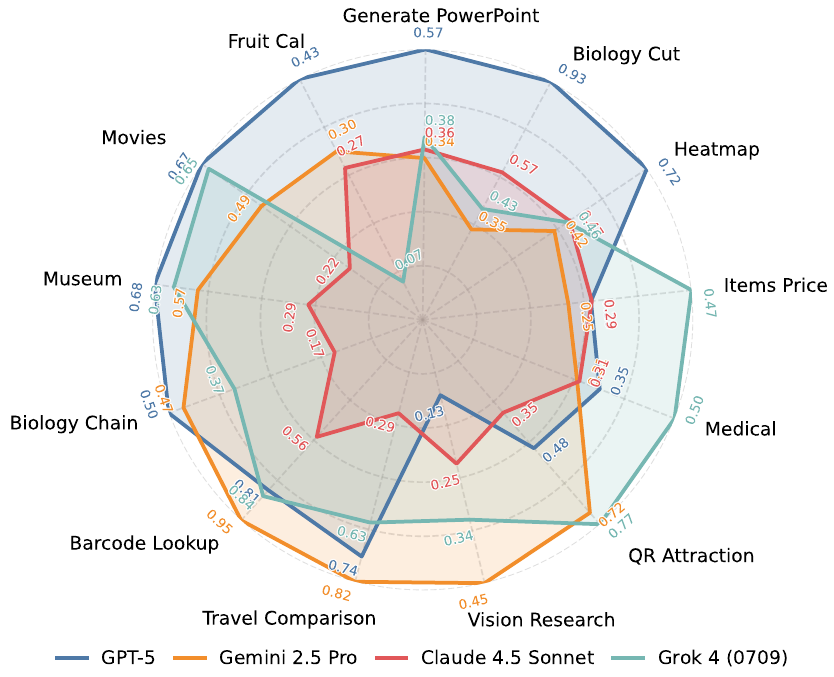}
    \caption{Average score of each selected task (details in \autoref{subsec:task-descriptions}) in the representative model.}
    \label{fig:task-lines}
    \vspace{-12pt}
    \end{figure}
    
    \autoref{fig:task-lines} shows that model capabilities are far from uniform across selected tasks (details in \autoref{subsec:task-descriptions}): curves cross frequently, indicating complementary strengths rather than a single, uniformly dominant system. \texttt{Grok-4} is weak on \emph{Fruit Cal} (numerical calibration/unit handling) yet strong on \emph{QR-code}/\emph{Attraction} (optical character recognition (OCR) and points-of-interest (POI) chaining). \texttt{GPT-5} trails on \emph{Vision Research} (fine-grained figure grounding) but dominates \emph{Biology Cut} (precise localization/cropping). \texttt{Gemini 2.5 Pro} excels on \emph{Travel Comparison} (web aggregation and itinerary synthesis). \texttt{Claude 4.5} lags on vision-dependent tasks (e.g., \emph{Items Price}), suggesting comparatively weaker multimodal grounding; disparities align with differences in OCR robustness. In \autoref{sec:per_task_eval} (See Appendix), we present the performance of different models on each task, based on the evaluation metrics of M$^3$-Bench.

\subsection{Budgets across models.}

\autoref{tab:token_counts} complements these interaction statistics by reporting average per-task input and output token budgets. \autoref{tab:avg_rounds_calls} summarizes the average interaction depth per task. Larger models often pursue longer, multi-hop plans: \texttt{GPT-5} \cite{openai2025gpt5systemcard} and \texttt{GPT-5 Nano} \cite{openai2025gpt5nano} average (3.55 / 14.77) and (5.13 / 17.97) (steps / tool calls), indicating frequent retries and non-trivial orchestration; \texttt{Llama-4-Scout-17B16E} \cite{meta2025llama4scout17b16e} and \texttt{Qwen2.5-VL-72B} \cite{bai2025qwen25vl} show similarly heavy tool use. By contrast, \texttt{Gemini 2.5 Pro} \cite{comanici2025gemini25} (3.04 / 6.95), \texttt{Gemini 2.5 Flash} \cite{comanici2025gemini25} (4.04 / 7.12), and \texttt{Grok-4 Fast} \cite{xai2025grok4fast} (2.91 / 5.93) reach goals with fewer, more precise calls, consistent with their higher detection precision. Very short dialogues (\eg, \texttt{Claude 4.5 Haiku} \cite{anthropic2025haiku45}, \texttt{GLM 4.5v} \cite{hong2025glm45v}) reflect under-coverage rather than efficiency.

\subsection{Analysis of Error Distributions}
\label{subsec:call-quality}

We assess per-call reliability by replaying each model's trajectory through the MCP interpreter and categorizing every \texttt{tool\_call} into one of five mutually exclusive outcomes shown in \autoref{fig:call_eval_pies}: (i) \emph{Illegal calling format}: the MCP request cannot be parsed into a valid schema (\eg, missing \texttt{name} or \texttt{arguments}); (ii) \emph{Unknown Tool Invocation}: the request targets a non-existent or non-exposed tool; (iii) \emph{Invalid Invocation Arguments}: the tool is found but arguments are syntactically/semantically invalid (typically surface as HTTP~400 or ``[Tool error] Invalid arguments''); (iv) \emph{Successful (Resource Not Found)}: the call is structurally correct yet the backend cannot route/locate the requested resource (\eg, HTTP~404); and (v) \emph{Successful call}: the tool executes and returns a normal result. These categories allow us to decouple schema conformance, tool selection, argument fidelity, and downstream retrieval robustness.

Overall, higher-performing systems allocate a larger share to \emph{Successful call} in \autoref{fig:call_eval_pies}, with the GPT-5 Grok and Gemini families exhibiting the strongest success rates in our setting. For \texttt{Qwen2.5-VL-72B} \cite{bai2025qwen25vl} and \texttt{GLM-4.5V} \cite{hong2025glm45v}, the shortfall appears driven less by multi-hop reasoning capacity and more by tool-use hygiene: hallucination-triggered invocations of non-existent tools and weaker adherence to instruction/schema constraints (the trajectory error analysis in Appendix \href{traj:Model: glm-4.5v}{Error: GLM-4.5v}, \href{traj:Model: Qwen2.5-VL-72B-Instruct}{Error: Qwen2.5-VL}, \href{traj:Model: internvl-3.5}{Error: Internvl-3.5}). This pattern suggests that improvements in schema guidance and instruction following would likely yield larger gains than refining high-level reasoning alone.
Although the Claude family is generally regarded as possessing strong capabilities in coding, mathematics, and reasoning, its cross-modal prior knowledge is not particularly prominent \autoref{traj:Model: claude-sonnet-4-5}.

\vspace{-5pt}


\section{Conclusion}
\label{sec:Conclusion}

We introduced M$^3$-Bench, a benchmark tailored to multimodal tool use under the Model Context Protocol (MCP), with trajectories that support multi-hop and multi-threaded execution. Our evaluation pipeline aligns predicted and reference calls by serializing each call, embedding it with a text encoder, and performing tool-name–bucketed Hungarian matching to obtain auditable one-to-one correspondences. On this alignment, we report a structure-aware suite of interpretable metrics.
Across representative MLLMs, results reveal persistent gaps in argument fidelity and structural consistency, highlighting the need for agents that jointly reason over images, text, and tool graphs.

{
    \small
    \bibliographystyle{ieeenat_fullname}
    \bibliography{main}
}

\clearpage
\setcounter{page}{1}
\maketitlesupplementary

\subsection{Task Descriptions}
\label{subsec:task-descriptions}

\paragraph{(0) Items Price}
Estimate prices for all visible retail items.
\begin{enumerate}
    \item Identify product names from the image (OCR/visual grounding as needed).
    \item Use Amazon MCP (search\_products) to retrieve current prices.
    \item Aggregate results into a concise table.
\end{enumerate}

\paragraph{(1) Attraction}
Resolve the attraction, surface context, and plan around-visit options.
\begin{enumerate}
    \item Use Wikipedia MCP to obtain the city/background of the attraction.
    \item Call weather/get\_weather for current conditions.
    \item Use google-maps/places\_nearby\_search to find nearby restaurants.
\end{enumerate}

\paragraph{(2) Movies}
Identify the movie and retrieve showtimes contextually.
\begin{enumerate}
    \item Use OCR to read the movie title in the image.
    \item Query TMDB to disambiguate the title.
    \item Search cinemas near New York showing the resolved film via Google Maps MCP.
\end{enumerate}

\paragraph{(3) Vision Research}
Bridge detected fauna to literature snippets.
\begin{enumerate}
    \item Detect animals and identify two distinct categories from detections.
    \item For each, call paper\_search/search\_arxiv with a targeted query.
    \item Download and read the paper; extract the shortest Abstract sentence containing the animal name.
\end{enumerate}

\paragraph{(4) Camping}
Plan a 3-day national-park camping itinerary with verified amenities and logistics.
\begin{enumerate}
    \item Use nationalparks/getCampgrounds and nationalparks/getAlerts to select an open campground with $\geq$ 3 amenities and no active alerts.
    \item Call nationalparks/getVisitorCenters and pick the closest visitor center; record hours.
    \item Use google-maps/compute\_route, distance\_matrix, and elevation to compute routes and elevation; compile the itinerary.
\end{enumerate}

\paragraph{(5) Fish}
\label{json:fish}
Plan a New Jersey fishing trip for the detected species.
\begin{enumerate}
    \item Use wiki/summary to summarize the detected fish species.
    \item Use wiki/search to search for the top spot for fishing in NJ recently.
    \item Call weather/get\_weather for local conditions at the spot.
    \item Use google-maps/compute\_route to plan a route from a default start.
    \item Use google-maps/places\_nearby\_search to search for restaurants and bait shops.
\end{enumerate}

\paragraph{(6) Drug}
Compare medicines for a given condition safely and conveniently.
\begin{enumerate}
    \item Look up drug facts via fda\_drug\_lookup MCP.
    \item Search similar products via Amazon (search\_products).
    \item Create a comparison list using Excel MCP.
\end{enumerate}

\paragraph{(7) Hazard}
Document construction-safety hazards with visual annotations and slides.
\begin{enumerate}
    \item Run a detector to find hardhats, vests, ladders, and unguarded edges.
    \item Use imagesorcery-mcp/draw\_circles to annotate detections.
    \item Create a presentation with ppt/create\_presentation, ppt/add\_slide and ppt/manage\_image.
\end{enumerate}

\paragraph{(8) Food Reddit Wiki Excel}
Quantify food popularity and descriptions, then export to Excel.
\begin{enumerate}
    \item For each food, query Reddit MCP for the hottest post and extract comment count.
    \item Use Wiki MCP to get a brief description.
    \item Create a workbook and write two columns (Food, Wikipedia Description) via Excel MCP; save the file.
\end{enumerate}

\paragraph{(9) MedExcel}
Summarize PubMed signal into key terms in Excel.
\begin{enumerate}
    \item Use pubmed\_search to retrieve the first three titles for the drug/disease.
    \item Double-check whether each title is semantically relevant to the queried drug/disease.
    \item Write the drug/disease name and the relevant paper count to an Excel file under ./media and save.
\end{enumerate}

\paragraph{(10) Barcode Lookup}
\label{json:barcodelookup}
Identify the book via barcode and fetch bibliographic data.
\begin{enumerate}
    \item Decode the barcode using pyzbar.
    \item Query openlibrary\_mcp to return the book record.
\end{enumerate}

\paragraph{(11) Biology Cut}
\label{json:biologycut}
Produce one image per detected animal instance.
\begin{enumerate}
    \item Detect all animals in the image.
    \item Crop each detection to an individual image.
\end{enumerate}

\paragraph{(12) Fruit Cal}
Assess daily calories and potential long-term risks from fruit-only meals.
\begin{enumerate}
    \item Use Nutrition MCP to estimate total daily calories from the shown fruits.
    \item Use Healthcare MCP to assess long-term health concerns of such a diet.
\end{enumerate}

\paragraph{(13) Medical}
Recommend safe medicines with purchase links conditioned on ICD-10.
\begin{enumerate}
    \item Use lookupICDCode MCP to validate the ICD-10 code.
    \item Use fda\_drug\_lookup and clinical\_trials\_search to check safety/compatibility.
    \item Suggest appropriate options via Amazon MCP with links.
\end{enumerate}

\paragraph{(14) QR Attraction}
\label{json:qrattraction}
Infer location from a QR-sourced page and report air quality and guidance.
\begin{enumerate}
    \item Use a QR scanner tool to extract the URL; fetch the primary image.
    \item Infer the location and geocode it via Google Maps MCP.
    \item Query google-air/current\_conditions for AQI and health recommendations.
\end{enumerate}

\paragraph{(15) Routine}
Plan a multi-stop route with a supermarket stop and choose the fastest option.
\begin{enumerate}
    \item Geocode the start, the image landmark, and the destination via Google Maps MCP.
    \item Search for nearby supermarkets as waypoints.
    \item Compare $\geq$ 3 candidate routes (compute\_route or directions) and select the fastest.
\end{enumerate}

\paragraph{(16) Biology Chain}
\label{json:biologychain}
Evaluate whether detected organisms can form a plausible food chain.
\begin{enumerate}
    \item Detect all organisms; blur any human instances.
    \item Use Wikipedia to retrieve ecological roles (diet, trophic level).
    \item Check whether roles form a valid chain.
\end{enumerate}

\paragraph{(17) Generate PowerPoint}
\label{json:generatepowerpoint}
Recreate a title slide with style and optional Q\&A enhancement.
\begin{enumerate}
    \item OCR the slide and segment into blocks; estimate font color/size/style.
    \item Call create\_presentation to get presentation\_id; use create\_slide\_from\_template to build the slide.
    \item If a question is present, fetch and summarize Reddit opinions for a Q\&A slide.
    \item save\_presentation to output PPTX.
\end{enumerate}

\paragraph{(18) Travel Comparison}
Choose the cleaner destination between two landmarks and suggest nearby venues.
\begin{enumerate}
    \item Recognize both landmarks and geocode each via Google Maps MCP.
    \item Use google-air/current\_conditions to compare AQI and pollutant levels.
    \item Recommend the cleaner destination; list nearby cinemas and hotels.
\end{enumerate}

\paragraph{(19) Heatmap}
Generate an air-pollution heatmap visualization for the depicted city.
\begin{enumerate}
    \item Geocode the city via Google Maps MCP.
    \item Render a static base map (static\_map) with zoom=10.
    \item Overlay PM2.5 heatmap tiles via air-quality/heatmap\_tile (indexType=US\_AQI).
\end{enumerate}

\paragraph{(20) Museum}
Retrieve a Met object and attach a concise Wikipedia summary.
\begin{enumerate}
    \item Call metmuseum-mcp/list-departments and select the best-aligned departmentId.
    \item Call metmuseum-mcp/search-museum-objects (hasImages=true, departmentId) and select an objectId.
    \item Call metmuseum-mcp/get-museum-object to get title, classification, and image URL.
    \item Use wiki/search then wiki/summary to obtain a short cultural summary.
\end{enumerate}

\paragraph{(21) Research Station}
Relate solar flare events to nearby research stations for the inferred location.
\begin{enumerate}
    \item Use nasa-mcp/get\_notifications to filter M5+ flares within the specified window.
    \item For each event date, call get\_solar\_flare and get\_coronal\_mass\_ejection; store flux/peak\_time and CME IDs/types.
    \item Infer photo location; use google-maps-mcp/places\_text\_search and place\_details to list top research stations and compile a table.
\end{enumerate}

\paragraph{(22) Hugeicons}
Map salient objects to Wikipedia titles and validated icons.
\begin{enumerate}
    \item Use dinox-mcp/detect-all-objects and keep detections with area $\textgreater$ 10{,}000.
    \item For each object, call wiki/search\_wikipedia to get an article title.
    \item Call hugeicons/search\_icons and hugeicons/list\_icons; cross-validate up to two icon names per object.
    \item Output a JSON table with Object, Wikipedia Title, and Icon Names.
\end{enumerate}

\paragraph{(23) Crypto Purchase Estimation}
Estimate how many main products 100 units of the most volatile currency can buy.
\begin{enumerate}
    \item Use dinox-mcp/detect-all-objects and select the object with the largest bounding box as the main product.
    \item Query amazon/search\_products for price/title; compute 24h volatilities from okx-mcp/get\_candlesticks across candidate instruments and pick the most volatile.
    \item Get current price and 24h volume via okx-mcp/get\_price.
    \item Compute purchasing power and estimated items purchasable; return structured JSON.
\end{enumerate}

\paragraph{(24) Stock Growth Estimation}
Estimate the short-term growth potential of the companies and annotate the image with the final result.
\begin{enumerate}
    \item Use yahoo-finance/get\_stock\_info to retrieve real-time prices for a group of candidate company tickers and keep the two highest-priced companies for further analysis.
    \item For selected companies, call yahoo-finance/get\_historical\_stock\_prices over 2024-10-01 to 2024-10-03, compute the average of their daily Close prices, and keep the company with higher average for further analysis.
    \item Call yahoo-finance/get\_recommendations for the final selected company and compute the upgrade ratio. Then compute the short-term growth percentage using the formula: 
    Growth Percentage = \(\frac{\text{Real-time Price} - \text{Average Close Price}}{\text{Average Close Price}} \times 100\) \(+\) (Upgrade Ratio \(\times 10\)).
    \item Use imagesorcery-mcp/draw\_texts to write the computed growth estimate onto the input image, and return the resulting ticker, its price metrics, the upgrade ratio, the estimated growth, and the annotated image.
\end{enumerate}

\paragraph{(25) FIPE Car Variant Lookup}
Identify the vehicle, determine its 20xx FIPE variant, and summarize its fuel type.
\begin{enumerate}
    \item Identify the vehicle and extract both the brand name and model keywords. Query car-price/get\_car\_brands for the complete list of supported brands. Call okx/get\_exchange\_rate to obtain the current BRL→USD and BRL→BTC conversion rates.
    \item If the brand exists in the supported list, then use its brand name and model keywords, call car-price/search\_brand\_model\_price to retrieve all matching FIPE entries, filter these results to keep only the 20xx variant. 
    \item Use wiki/summary to look up the fuel type of the 20xx variant, and provide a short description based on the search result.
\end{enumerate}

\paragraph{(26) Class Demand Estimation}
Estimate class demand, compare it with the detected object count, and determine whether the class budget is sufficient.
\begin{enumerate}
    \item Use Math-MCP tools math/mean and math/median to compute median and mode from the collected survey statistic \{6, 8, ...\}. Use dinox-mcp/detect-all-objects to detect all objects and identify the most frequently appearing object category. Compare the survey median, the survey mode, and the detected top object count, then take the largest among them as the estimated demand.
    \item Use amazon/search\_products to query the detected top object category, obtain the first returned product, and compute the total cost as the ceiling of the estimated demand multiplied by the item price. 
    \item If the total cost is within the class budget of xxx dollars budget, call Reddit-MCP-Server/search\_hot\_posts to fetch the first recent hot post and wiki/summary to obtain a short description for the detected object category.
\end{enumerate}

\paragraph{(27) NixOS Diagnostic}
Perform a staged NixOS diagnostic workflow using OCR, NixOS tools, and Wiki tools.
\begin{enumerate}
    \item Use ocr/perform\_ocr exactly once to extract the package names appearing in the code on the image. These package names will be used in the following steps.
    \item Take the first five package names in their original order and call nixos/nixos\_search once for each of them to check whether they exist in NixOS. If fewer than three of these packages exist, stop the workflow and report.
    \item Otherwise, collect the first three package names that do exist. Call nixos/nixos\_info on services.openssh.enable. Then call nixos/home\_manager\_search with the query “zsh” and keep the first three returned options; choose the first one and call nixos/home\_manager\_info. Finally, call wiki/summary once for each of the three existing packages to provide brief background information.
\end{enumerate}

\subsection{LLM Prompts Used in Executor and Judge}

\paragraph{Prepare (Tool Planning)}
\label{prompt:prepare}
Following the planning protocol described in Experiment \autoref{exp:plan}, the Executor first selects (but does not execute) the tool set for the current step. The exact prompt template used for this stage is given below.
    \begin{PromptBox}{Prepare (Tool Planning)}
    You are an assistant with MCP tool invocation capability. You must call one or more MCP tools, possibly multi-hop, to complete the task. Consider necessary steps as much as possible, consider completed steps in conjunction with conversation history, and skip unnecessary steps. Extra steps may reduce your final score
    This is step {i} of {self.max_step}.
    Now decide which tools to use for this step.
    Output EXACTLY in the following format (no extra text):

      "server_name/tool_name: <description> 
        Args: <arguments (type): description> 
        Returns: <returns(type): description>"

    <plan for how to use>

    Here are the MCP tools for you. All tools list (name: description):
    [the full tool catalog is appended programmatically]
    \end{PromptBox}

\paragraph{Process (Controlled Execution)}
\label{prompt:process}
Given the selected tools, execution proceeds under controlled, JSON-only invocation with bounded parallelism as specified in Experiment \autoref{exp:process}. The prompt below instantiates that protocol.
    \begin{PromptBox}{Process (Controlled Execution)}
    Now execute the selected tools for this step. Return STRICT JSON ONLY with a single top-level key 'tool_calls' containing 1 to {self.max_concurrent} items. Consider necessary steps as much as possible, consider completed steps in conjunction with conversation history, and skip unnecessary steps. Extra tool calls may reduce your final score
    Each item must be an object: {"name": "server_name/tool_name", "arguments": { ... }}.
    Do NOT include markdown fences or any extra text or special delimiters and tokens.
    If a tool requires images/files, include the appropriate arguments per the tool's schema. You may use the image pool paths listed below if applicable.
    Selected tools (name: description):
    [The selected tools list is appended programmatically]
    [if available, image pool paths are appended]
    \end{PromptBox}

\paragraph{Stop/Continu}
\label{prompt:stop}
A binary stopping criterion gates the transition to summarization, as defined in Experiment \autoref{exp:decision}. The following prompt enforces a strict yes/no decision.
    \begin{PromptBox}{Stop/Continu}
    If the task can be considered complete based on current tool results and question, here is the original question: {last_user},
    answer 'yes'. Otherwise 'no'.
    Answer strictly with 'yes' or 'no', no extra words.
    \end{PromptBox}

\paragraph{Final Reply}
\label{prompt:final}
When the stopping criterion is met, the agent emits only the final natural-language answer (Experiment \autoref{exp:decision}). The corresponding prompt template is provided below.
    \begin{PromptBox}{Final Reply}
    You are finalizing the conversation. Produce ONLY the final answer in natural language.
    Do NOT include any <tool_call> tags or mention tools explicitly. Be concise and accurate,
    relying on the prior tool results contained in the conversation.
    \end{PromptBox}

\paragraph{Judge}
\label{prompt:judge}
The Judge independently audits causal order and factual correctness, per Experiment \autoref{exp:judge_summarize}. The prompt below operationalizes this decision.
    \begin{PromptBox}{Judge}
    You are a strict task finisher. Read the provided interaction history and decide if the task is completed. 
    Very important! If there is an incorrect causal order in multi-step MCP tools calls, or if the information returned by the MCP tools call is clearly wrong, but the call chain proceeds smoothly due to LLM hallucinations, it should be judged as an error.
    Output EXACTLY one of: success or fail. No other words.
    \end{PromptBox}

\paragraph{Summarize}
\label{prompt:summarize}
Upon success, the Judge reconstructs an optimal trajectory by merging/removing calls and grouping parallel operations (Experiment \autoref{exp:judge_summarize}). We provide the standard summarization prompt below.
    \begin{PromptBox}{Summarize}
    You will summarize the optimal trajectory of useful MCP tool calls to solve the task.
    Rules:
    - Remove failed or unnecessary tool calls.
    - Group calls by step; within a step, multiple calls are in parallel (array order arbitrary).
    - Operations that can invoke the MCP tools call in parallel were mistakenly executed across multi-hop --- they should be merged into a single step.
    - Preserve only name and arguments for each call.
    - Output STRICT JSON only. No markdown, no comments, no extra text.
    - If you cannot determine a valid trajectory, output exactly the string fail (no JSON).
    \end{PromptBox}

\paragraph{Task Completion}
\label{prompt:task_completion}
For end-to-end task completion, we use a four-judge trimmed-mean score defined in Experiment \autoref{subsubsec:taskcompletion}. The prompt below elicits the required scalar.
    \begin{PromptBox}{Task Completion}
    You are a strict and professional academic reviewer tasked with evaluating task completion when a model uses MCP tools.
    Judge solely based on the visible prediction content; do not use or allude to any ground truth.
    Scoring rule: output a single scalar in the form \\boxed{S}, and nothing else. S is a float in [0,10].
    Rules:
    - 1) Planning (0-3): Clearly presents the MCP call process and plan; complete and organized.
    - 2) Process (0-3): MCP calls obtain desirable tool feedback and intermediate results; tool use is effective.
    - Final result (0-4): Degree of task completion (partial credit allowed); final output addresses the task.
    - Please provide an overall score from 0 to 10 combining the three parts, and output only \\boxed{S}.
    \end{PromptBox}

\paragraph{Information Grounding}
\label{prompt:information_grounding}
Step-level groundedness is evaluated against the reference steps as defined in Experiment \autoref{subsubsec:information_grounding}. The prompt below collects the coverage-only score.
    \begin{PromptBox}{Information Grounding}
    You are a strict and professional academic reviewer responsible for evaluating information grounding.
    Task: compare the reference steps (ground truth steps) with the predicted steps and determine whether the prediction includes all key steps from the reference.
    Rules:
    - Coverage-only criterion: equivalence/paraphrase/minor order changes count as covered;
    - Extra steps are not penalized;
    - Missing or clearly deviating key steps are penalized linearly by the proportion of missing steps relative to the total reference steps;
    Output format: output only a scalar in the form \\boxed{G}, where G is a float in [0,1].
    \end{PromptBox}

\subsection{Bucketed Hungarian sample pairs based on similarity}
\begin{figure}[!htbp]
    \centering
      \begin{subfigure}[b]{0.5\textwidth}
        \centering
        \includegraphics[width=\textwidth]{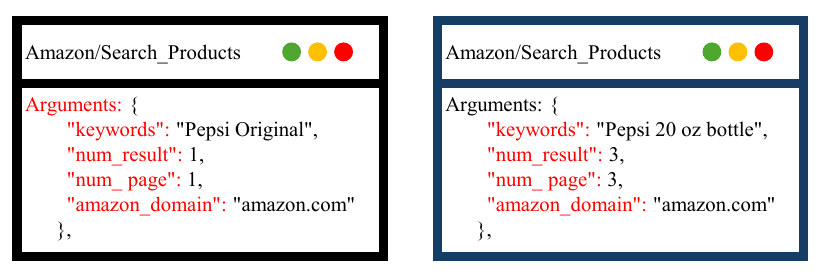}
        \caption{High similarity}
        \label{fig:example1}
      \end{subfigure}
      \begin{subfigure}[b]{0.5\textwidth}
        \centering
        \includegraphics[width=\textwidth]{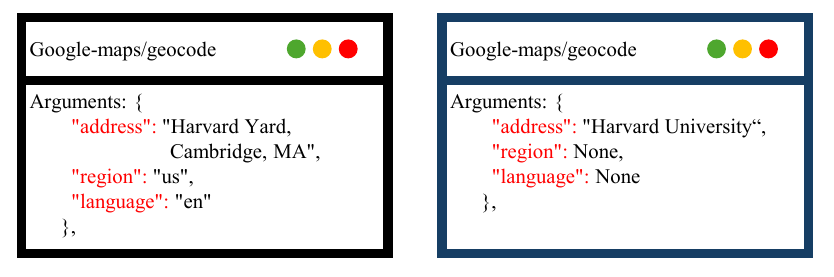}
        \caption{Medium similarity}
        \label{fig:example2}
      \end{subfigure}
      \begin{subfigure}[b]{0.5\textwidth}
        \centering
        \includegraphics[width=\textwidth]{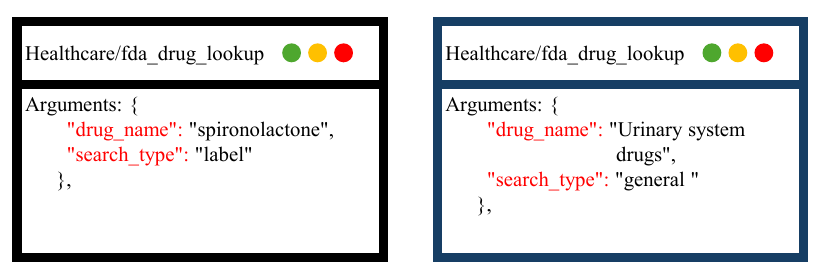}
        \caption{Low similarity}
        \label{fig:example3}
      \end{subfigure}

    \caption{Illustrative call-level similarity under our sentence-encoder + cosine metric with bucketed Hungarian matching: three representative pairs exhibiting high, medium, and low similarity.
    This visualization contextualizes \emph{Argument Similarity} and the weak/strong thresholding used for alignment (see \autoref{subsubsec:embed}; cf. related agent evaluation protocols.}
    \label{fig:similarity_call}
    \end{figure}

To make the alignment intuition concrete, \autoref{fig:similarity_call} shows three representative call pairs after canonicalization and sentence-encoder cosine scoring. Within each tool bucket, only pairs clearing a weak screen $\tau_{\text{weak}}$  are eligible for Hungarian assignment; those above a stronger bar $\tau_{\text{strong}}$ are treated as high-confidence matches. The high/medium/low examples illustrate how semantically close arguments receive credit despite surface variation, while dissimilar pairs are withheld—clarifying how our \emph{Argument Similarity} feeds the bucketed matching used in \autoref{subsubsec:embed} and, downstream, the structure-aware metrics.

\subsection{Human Study}
\label{subsec:human-study}
We conducted a small-scale user study with 30 Ph.D. and researchers working with Model Context Protocol (MCP) tools to validate two design choices: (i) the use of human prompts and human judges (see \autoref{subsec:llm-judge}) for obtaining an optimal trajectory; and (ii) the similarity thresholding used together with the bucketed Hungarian (Kuhn--Munkres) matching (see \autoref{subsubsec:embed}).

\paragraph{Preference for Optimal Trajectory.}
Each participant received 20 randomly sampled pairs of trajectories (the original MLLM-produced trajectory versus the human-optimized \emph{Best Trajectory}), with order randomized and content blinded. Participants chose the better one per pair based on faithfulness and procedural soundness. Overall, the Best Trajectory was preferred in 98.16\% of cases; under our pairwise setup this corresponds to 589/600 selections (98.17\%).

\paragraph{Calibration of similarity thresholds.}
From the evaluation pool, we selected 40 ground-truth/predicted call pairs and asked participants to categorize them as \emph{strong}, \emph{medium}, or \emph{weak} correlation (majority vote). We then swept the similarity thresholds and found that $(\tau_{\mathrm{strong}},\tau_{\mathrm{weak}})=(0.8,0.6)$ produced the closest agreement with human labels.

\section{Trajectory Case Study}
\label{sec:trajectory-case-study}

\noindent\textbf{Format.} We present each trajectory using the two-column key–value layout defined in \texttt{trajectoryformat.tex}. Labels in the left column are typeset in monospace. \texttt{IMG} denotes the input image and \texttt{Q} the task query. \texttt{Step k} indicates the \(k\)-th multi-hop step in the execution. Within a step, dashed separators denote parallel (multi-threaded) tool calls. Each call reports \texttt{Tool} (invoked tool), \texttt{ARG} (arguments), and \texttt{OUT} (returned output). The final answer is summarized under \texttt{Final Reply}.

\noindent Object detection with per-instance cropping; parallel crop operations produce one image per detected animal. See \href{json:biologycut}{(11) Biology Cut}.

\begin{twocolkv}{Example: Object detection and cropping}

\KV{IMG}{\raisebox{-0.9\height}{\includegraphics[width=0.5\linewidth]{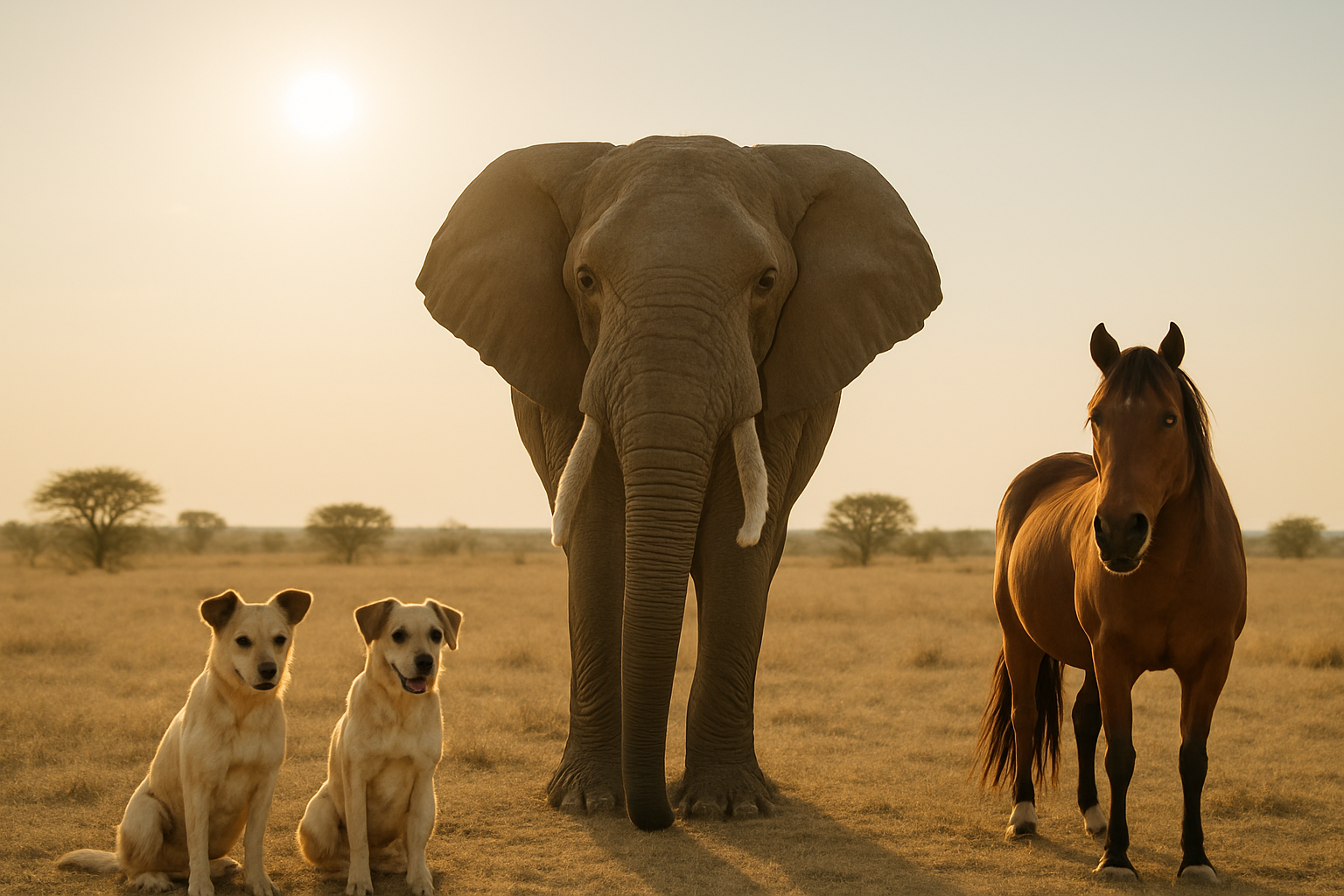}}}

\KV{Q}{\texttt{Could you spot every creature in the picture and give me individual cutouts for each?}}

\KVsection{Step 1}
\KV{Tool}{\texttt{dinox-mcp/detect-all-objects}}
\KV{ARG}{\texttt{imageFileUri=./media/png; includeDescription=false}}
\KV{OUT}{\texttt{Objects detected: horse (1), dog (2), elephant (1). Bboxes use \{xmin, ymin, xmax, ymax\}; coordinates describe exact positions and spatial relations.}}

\KVsection{Step 2}
\KV{Tool}{\texttt{imagesorcery-mcp/crop}}
\KV{ARG}{\texttt{input\_path=./media/png; x1=473; y1=153; x2=1079; y2=952; output\_path=./media/elephant.png}}
\KV{OUT}{\texttt{Image crop completed; saved as \texttt{./media/elephant.png}.}}

\KVsplit
\KV{Tool}{\texttt{imagesorcery-mcp/crop}}
\KV{ARG}{\texttt{input\_path=./media/png; x1=1105; y1=364; x2=1428; y2=1015; output\_path=./media/horse.png}}
\KV{OUT}{\texttt{Image crop completed; saved as \texttt{./media/horse.png}.}}

\KVsplit
\KV{Tool}{\texttt{imagesorcery-mcp/crop}}
\KV{ARG}{\texttt{input\_path=./media/png; x1=61; y1=671; x2=359; y2=1024; output\_path=./media/dog1.png}}
\KV{OUT}{\texttt{Image crop completed; saved as \texttt{./media/dog1.png}.}}

\KVsplit
\KV{Tool}{\texttt{imagesorcery-mcp/crop}}
\KV{ARG}{\texttt{input\_path=./media/png; x1=338; y1=682; x2=531; y2=1024; output\_path=./media/dog2.png}}
\KV{OUT}{\texttt{Image crop completed; saved as \texttt{./media/dog2.png}.}}

\KVsection{Final Reply}
\begin{KVfinal}
\texttt{Found 4 creatures: 1 elephant, 1 horse, 2 dogs. Each has a precise bbox (\{xmin, ymin, xmax, ymax\}, origin at top-left) and a cropped cutout under ./media/*.png.}
\end{KVfinal}
\end{twocolkv}
\vspace{0.8\baselineskip}

\paragraph{Fish}\label{traj:fish}
\noindent Species identification with geospatial planning: Wiki species summary, suitable local spot, weather, routing, and nearby POIs for an executable trip plan. See \href{json:fish}{(5) Fish}.

\begin{twocolkv}{Example: Saltwater fish ID and NJ trip planning}
\KV{IMG}{\raisebox{-0.9\height}{\includegraphics[width=0.5\linewidth]{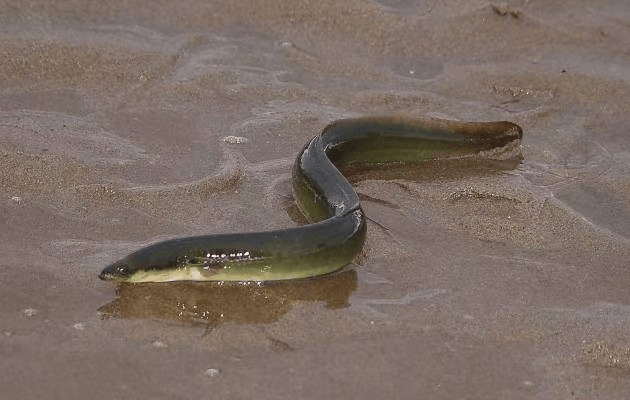}}}

\KV{Q}{\texttt{From the NJ fish photo, can you ID the species with a brief overview, pick a recent good spot to try, check the current weather there, map a route from the usual starting point, and list nearby restaurants and a bait shop within about 5 km?}}

\KVsection{Step 1}
\KV{Tool}{\texttt{wiki/summary}}
\KV{ARG}{\texttt{title="American eel"}}
\KV{OUT}{\texttt{Species identified as American eel (\textit{Anguilla rostrata}); catadromous, slender body, continuous dorsal/anal fin, nocturnal feeder common along Atlantic coast and NJ estuaries.}}

\KVsplit
\KV{Tool}{\texttt{wiki/search}}
\KV{ARG}{\texttt{query="Sandy Hook fishing New Jersey"; n=5}}
\KV{OUT}{\texttt{Sandy Hook (Gateway NRA, near Highlands, NJ) highlighted as productive surf/shore area for eels and other inshore species.}}

\KVsection{Step 2}
\KV{Tool}{\texttt{weather/get\_weather}}
\KV{ARG}{\texttt{location="Highlands, NJ"; units="us"}}
\KV{OUT}{\texttt{Current conditions around Highlands, NJ show mild temperatures (~52.2\textdegree F) and light winds (~4.1 mph), suitable for an evening shore session.}}

\KVsection{Step 3}
\KV{Tool}{\texttt{google-maps/compute\_route}}
\KV{ARG}{\texttt{origin=\{latLng:\{40.7357,-74.1724\}\}; destination=\{latLng:\{40.4333,-73.9885\}\}; travelMode = DRIVE; fields = routes.distanceMeters, \dots}}
\KV{OUT}{\texttt{Route from Newark, NJ to Sandy Hook, NJ computed; distance $\sim$44 miles, duration $\sim$1 hour via Garden State Parkway and NJ-36 S.}}

\KVsection{Step 4}
\KV{Tool}{\texttt{google-maps/places\_nearby\_search}}
\KV{ARG}{\texttt{location=\{40.4333,-73.9885\}; radiusMeters=5000; includedTypes=["restaurant","fishing\_store"]; languageCode="en"; regionCode="US"}}
\KV{OUT}{\texttt{Nearby options include several waterfront restaurants (e.g., Bahrs Landing, The Proving Ground, Off The Hook, One Willow) and bait/tackle shops within 5 km.}}

\KVsection{Final Reply}
\begin{KVfinal}
\texttt{Species: American eel (\textit{Anguilla rostrata}); catadromous, active at night in tidal creeks/marshes. Spot: Sandy Hook, NJ (Gateway NRA near Highlands). Weather: calm, cool, fishable. Route: Newark→Sandy Hook = 44 mi / ~1 h. Amenities: multiple waterfront restaurants and local bait \& tackle within ~5 km. Plan: fish evening high/outgoing tides along bayside creek mouths and rock structures; use bloodworms or cut bait; register with NJ Saltwater Recreational Registry and follow current regulations.}
\end{KVfinal}
\end{twocolkv}
\vspace{0.8\baselineskip}

\paragraph{Generate Powerpoint}\label{traj:generatepowerpoint}
\noindent Text-based slide reconstruction and restyling; recreated via PPT tools with correct presentation\_id; optional Q\&A from Reddit if a question is present. See \href{json:generatepowerpoint}{(17) Generate PowerPoint}.

\begin{twocolkv}{Example: Recreating Slide through Screenshot}

\KV{IMG}{\raisebox{-0.9\height}{\includegraphics[width=0.5\linewidth]{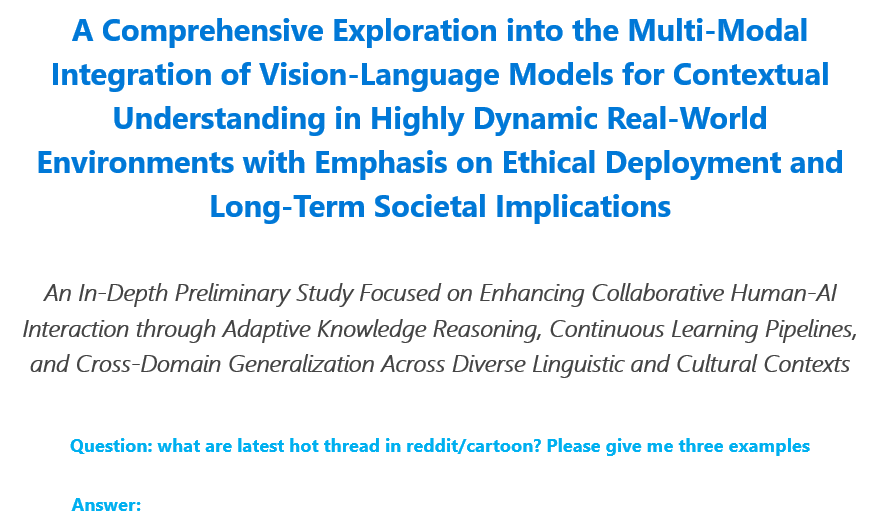}}}

\KV{Q}{\texttt{From the screenshot of the first slide, pull out all visible text grouped into blocks (title, subtitle, footer if any) with rough font size/color/style guesses, recreate it as a new presentation with a matching title page, and if any line is a question, add a short Answer made from a few relevant community takes, then save it.}}

\KVsection{Step 1}
\KV{Tool}{\texttt{ocr/perform\_ocr}}
\KV{ARG}{\texttt{input\_data=./media/png; language="eng"; config="--oem 3 --psm 6"}}
\KV{OUT}{\texttt{Extracted multi-line title and subtitle text; detected embedded question about Reddit cartoons.}}

\KVsplit
\KV{Tool}{\texttt{Reddit-MCP-Server/search\_hot\_posts}}
\KV{ARG}{\texttt{subreddit="cartoons"; limit=3}}
\KV{OUT}{\texttt{Fetched top hot threads: “fav purple character?”, “Ruh Roh Raggy”, “Choo choo I guess...”.}}

\KVsection{Step 2}
\KV{Tool}{\texttt{ppt/create\_presentation}}
\KV{ARG}{\texttt{id="pres"}}
\KV{OUT}{\texttt{Initialized new PowerPoint deck with modern blue theme.}}

\KVsection{Step 3}
\KV{Tool}{\texttt{ppt/create\_slide\_from\_template}}
\KV{ARG}{\texttt{template\_id="title\_slide"; presentation\_id="pres"; content\_mapping=\{title, subtitle, question, answer\}}}
\KV{OUT}{\texttt{Title + subtitle mapped; question block appended for Q\&A enhancement.}}

\KVsplit
\KV{Tool}{\texttt{ppt/manage\_text}}
\KV{ARG}{\texttt{operation="add"; slide\_index=0; text="Question: what are latest hot thread in reddit/cartoon?"; font\_size=20; bold=True; color=(0,120,215)}}
\KV{OUT}{\texttt{Inserted question line below subtitle with bright blue accent.}}

\KVsplit
\KV{Tool}{\texttt{ppt/manage\_text}}
\KV{ARG}{\texttt{operation="add"; slide\_index=0; text="- fav purple character?..."; font\_size=20; bold=True; color=(0,102,204)}}
\KV{OUT}{\texttt{Added Reddit answer list under question; auto-fit layout maintained.}}

\KVsection{Step 4}
\KV{Tool}{\texttt{ppt/save\_presentation}}
\KV{ARG}{\texttt{file\_path=./media/Recreated\_Slide.pptx; presentation\_id="pres"}}
\KV{OUT}{\texttt{File saved successfully; final slide index 0 complete.}}

\KVsection{Final Reply}
\begin{KVfinal}
\texttt{Recreated title slide for a presentation on multi-modal vision–language models.}
\texttt{Title (large, bold, blue \#0078D7); Subtitle (medium, italic, gray \#444444).}
\texttt{Detected question: “what are latest hot thread in reddit/cartoon?”.}
\texttt{Reddit answers inserted as Q\&A section:}
\texttt{1) fav purple character?... 2) Ruh Roh Raggy... 3) Choo choo I guess...}
\texttt{Presentation ID: pres. Final file: Recreated\_Slide.pptx.}
\texttt{Slide successfully rebuilt with question and answers in matching blue theme.}
\end{KVfinal}
\end{twocolkv}
\vspace{0.8\baselineskip}

\paragraph{Barcodelookup}\label{traj:barcodelookup}
\noindent Barcode decoding followed by OpenLibrary lookup for bibliographic metadata. See \href{json:barcodelookup}{(10) Barcode Lookup}.

\begin{twocolkv}{Example: Book Identification via Barcode Lookup}

\KV{IMG}{\raisebox{-0.9\height}{\includegraphics[width=0.4\linewidth]{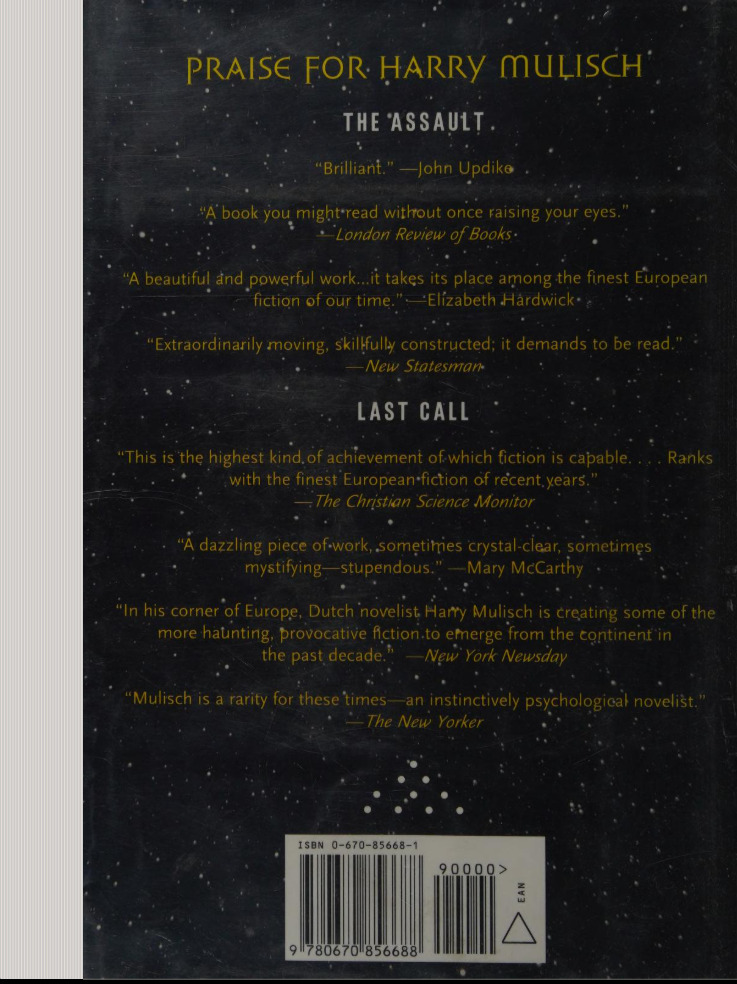}}}

\KV{Q}{\texttt{From the back-cover barcode, can you identify the book and provide its title, author, publisher, year, and page count?}}

\KVsection{Step 1}
\KV{Tool}{\texttt{pyzbar-mcp/scan\_barcode}}
\KV{ARG}{\texttt{image\_paths=[./media/png]}}
\KV{OUT}{\texttt{Barcode scan completed successfully. Found ISBN(s): 9780670856688. No further barcode scanning is needed — proceeding directly to book lookup.}}

\KVsection{Step 2}
\KV{Tool}{\texttt{openlibrary\_mcp/get\_book\_info}}
\KV{ARG}{\texttt{isbn="9780670856688"}}
\KV{OUT}{\texttt{Retrieved metadata from OpenLibrary: \textit{The Discovery of Heaven} by Harry Mulisch; published by Viking in 1996, 730 pages. Query type confirmed as ISBN lookup.}}

\KVsection{Final Reply}
\begin{KVfinal}
\texttt{Title: \textit{The Discovery of Heaven}.}
\texttt{Author: Harry Mulisch.}
\texttt{ISBN-13: 9780670856688 (ISBN-10: 0670856681).}
\texttt{Publisher: Viking; Year: 1996; Pages: 730.}
\texttt{Book successfully recognized from the scanned barcode using pyzbar-mcp and validated via OpenLibrary metadata.}
\end{KVfinal}
\end{twocolkv}
\vspace{0.8\baselineskip}

\paragraph{QR Attraction}\label{traj:qrattraction}
\noindent QR decoding to retrieve image, location inference with geocoding, and air-quality query for current AQI and guidance. See \href{json:qrattraction}{(14) QR Attraction}.

\begin{twocolkv}{Example: QR-linked Attraction and Air Quality Retrieval}
\KV{IMG}{\raisebox{-0.9\height}{\includegraphics[width=0.5\linewidth]{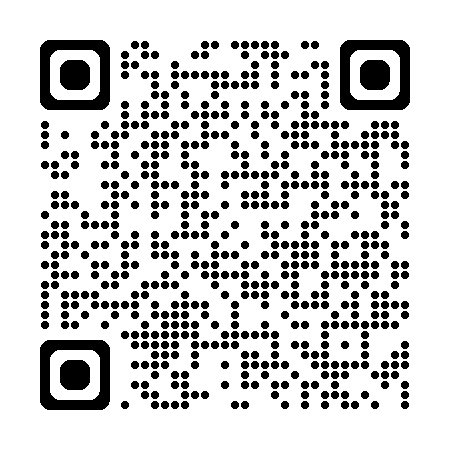}}}

\KV{Q}{\texttt{Pull the webpage link hidden in the image, grab the main photo from that page, infer the location shown, turn that location into coordinates, check current air quality there, and report AQI, pollutant levels, and health advice.}}

\KVsection{Step 1}
\KV{Tool}{\texttt{pyzbar-mcp/scan\_barcode}}
\KV{ARG}{\texttt{image\_paths=./media/png}}
\KV{OUT}{\texttt{Barcode scan completed successfully. Decoded one QRCode linking to \url{https://unsplash.com/photos/people-near-building-\_K2uemedTGo}. No further scanning required — proceeding to image retrieval.}}

\KVsection{Step 2}
\KV{Tool}{\texttt{linkimage-mcp/fetch\_unsplash\_image}}
\KV{ARG}{\texttt{url="https://unsplash.com/photos
/people-near-building-\_K2uemedTGo"}}
\KV{OUT}{\texttt{Downloaded Unsplash photo by Nick Fewings successfully. Image saved locally and accessible at \texttt{./images/\_K2uemedTGo.jpg}.}}

\KVsection{Step 3}
\KV{Tool}{\texttt{google-maps/geocode}}
\KV{ARG}{\texttt{address="Natural History Museum, London"; language="en"}}
\KV{OUT}{\texttt{Geocoding completed — location resolved to Natural History Museum, London. Coordinates: 51.496715, -0.1763672 (Cromwell Rd, South Kensington, London SW7 5BD, UK).}}

\KVsection{Step 4}
\KV{Tool}{\texttt{google-air/current\_conditions}}
\KV{ARG}{\texttt{lat=51.496715; lng=-0.1763672}}
\KV{OUT}{\texttt{Fetched current air-quality data for London (Natural History Museum area). AQI: 82 — Excellent air quality. Dominant pollutant: NO\textsubscript{2}.}}

\KVsection{Final Reply}
\begin{KVfinal}
\texttt{QR decoded to Unsplash link.}
\texttt{Main image downloaded successfully (Nick Fewings).}
\texttt{Detected location: Natural History Museum, London (51.496715, -0.1763672).}
\texttt{Current Air Quality Index (UAQI): 82 — Excellent. Dominant pollutant: NO\textsubscript{2}.}
\texttt{Health guidance: All individuals may continue outdoor activity safely; those with asthma or NO\textsubscript{2} sensitivity should still avoid prolonged exposure to traffic-heavy roads. Ventilation and light exercise remain safe.}
\end{KVfinal}
\end{twocolkv}
\vspace{0.8\baselineskip}

\paragraph{Biology Chain}\label{traj:biologychain}
\noindent Organism detection with privacy blur, Wikipedia-based ecological roles, and validation of a plausible food chain. See \href{json:biologychain}{(16) Biology Chain}.

\begin{twocolkv}{Example: BiologyChain — Organism Listing, Privacy Blur, and Food-Web Check}

\KV{IMG}{\raisebox{-0.9\height}{\includegraphics[width=0.3\linewidth]{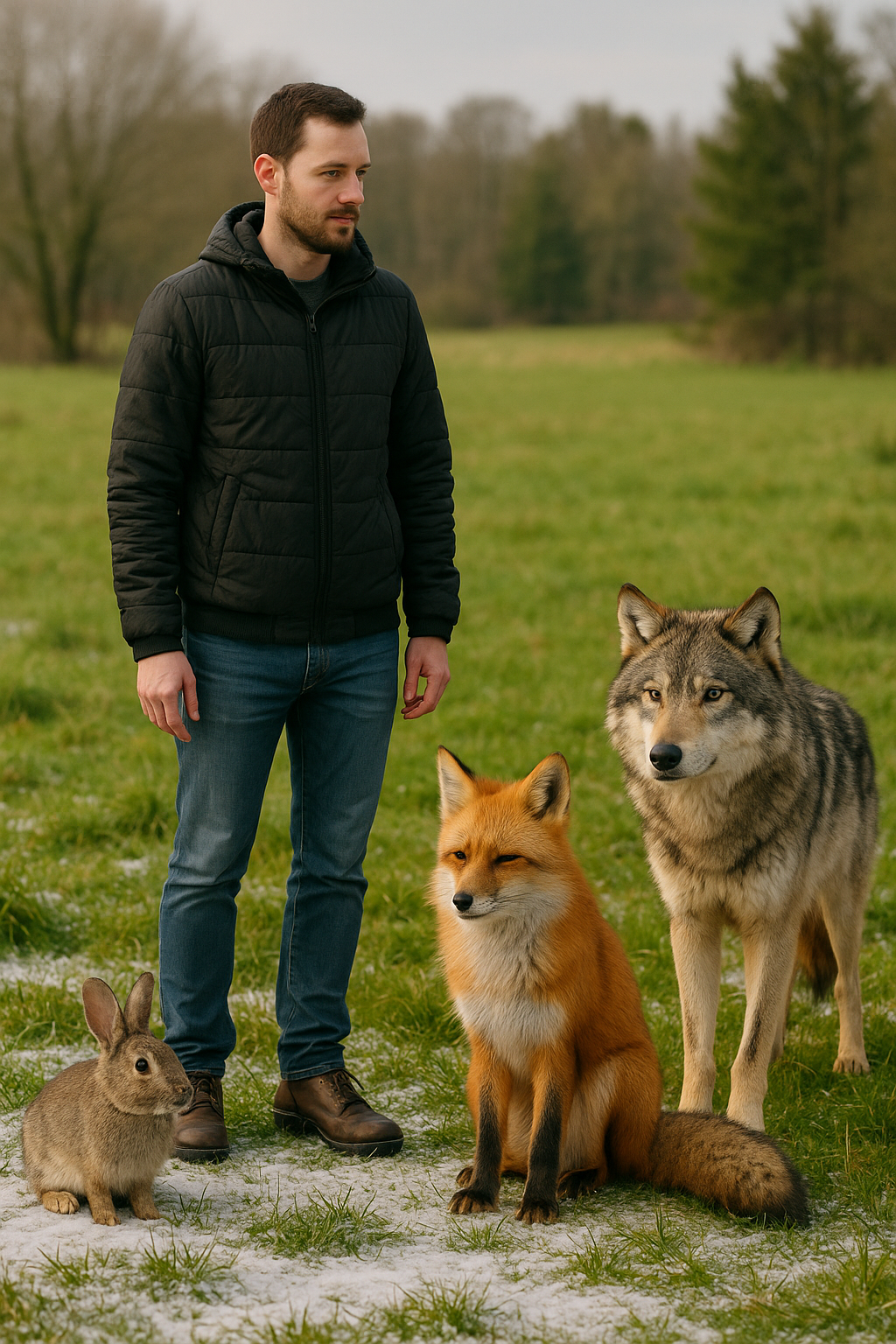}}}

\KV{Q}{\texttt{Can you check this picture, note every organism, blur any people for privacy, figure out what each one eats and where it sits in the web, and tell me if they could make a valid food chain?}}

\KVsection{Step 1}
\KV{Tool}{\texttt{dinox-mcp/detect-all-objects}}
\KV{ARG}{\texttt{imageFileUri=./media/png; includeDescription=false}}
\KV{OUT}{\texttt{Objects detected: wolf (1), person (1), fox (1), rabbit (1). Bboxes use \{xmin, ymin, xmax, ymax\}; coordinates describe exact positions and spatial relations.}}

\KVsection{Step 2}
\KV{Tool}{\texttt{imagesorcery-mcp/blur}}
\KV{ARG}{\texttt{input\_path=./media/png; areas=[\{x1=120,y1=93,x2=528,y2=1330, ...\}]; invert\_areas=false; output\_path=./media/blurred.png}}
\KV{OUT}{\texttt{Privacy blur applied to the person region. Output: {\ttfamily ./media/blurred.png}.}}

\KVsplit

\KV{Tool}{\texttt{wiki/summary}}
\KV{ARG}{\texttt{title="Rabbit"}}
\KV{OUT}{\texttt{Rabbits are small herbivorous mammals (family Leporidae) feeding on grasses and herbs; primary consumers and common prey for foxes and wolves.}}

\KVsplit
\KV{Tool}{\texttt{wiki/summary}}
\KV{ARG}{\texttt{title="Red fox"}}
\KV{OUT}{\texttt{Red fox (\textit{Vulpes vulpes}) — an omnivorous mesopredator eating rabbits, rodents, birds, insects, and fruits; trophic level $\approx$ 3.}}

\KVsplit
\KV{Tool}{\texttt{wiki/summary}}
\KV{ARG}{\texttt{titles="Gray wolf"}}
\KV{OUT}{\texttt{Wolf (\textit{Canis lupus}) — apex carnivore hunting large and small mammals (including hares and occasionally foxes); trophic level $\approx$ 4.}}

\KVsplit
\KV{Tool}{\texttt{wiki/summary}}
\KV{ARG}{\texttt{title="Human"}}
\KV{OUT}{\texttt{Humans (\textit{Homo sapiens}) — omnivorous primates with advanced cognition; top-level consumers but not part of the natural food chain here.}}

\KVsplit
\KV{Tool}{\texttt{wiki/summary}}
\KV{ARG}{\texttt{title="Grass"}}
\KV{OUT}{\texttt{Grasses (Poaceae) — photosynthetic primary producers forming the base of terrestrial food webs; trophic level 1.}}

\KVsplit
\KV{Tool}{\texttt{wiki/summary (Tree)}}
\KV{ARG}{\texttt{title="Tree"}}
\KV{OUT}{\texttt{Trees — perennial woody producers performing photosynthesis; provide structure and habitat within ecosystems.}}

\KVsection{Final Reply}
\begin{KVfinal}
\texttt{Privacy-safe image: ./media/blurred.png}

\texttt{Organisms noted}\\
\texttt{-- Grass/green plants — producers (photosynthesis).}\\
\texttt{-- Trees — producers.}\\
\texttt{-- Rabbit — herbivore; primary consumer.}\\
\texttt{-- Fox — omnivore/mesopredator; secondary consumer.}\\
\texttt{-- Wolf — apex carnivore; tertiary consumer.}\\
\texttt{-- Person — human omnivore (not part of wild chain).}

\texttt{Valid food chain?}\\
\texttt{Yes. Examples:}\\
\texttt{1) Grass $\rightarrow$ Rabbit $\rightarrow$ Wolf (common).}\\
\texttt{2) Grass $\rightarrow$ Rabbit $\rightarrow$ Fox $\rightarrow$ Wolf (ecologically possible; wolves often kill foxes as competitors and may occasionally consume them).}

\texttt{Plants act as producers; rabbit as primary consumer; fox as secondary; wolf as tertiary/apex. The organisms present can form a valid trophic sequence.}
\end{KVfinal}
\end{twocolkv}

\section{Error Trajectory Distribution Analysis}
\label{subsec:error trajectory distribution analysis}

We illustrate representative failure trajectories to analyze the characteristic error patterns of different multimodal large language models (MLLMs). Each case is presented following the same format.

\paragraph{GLM-4.5v}\label{traj:Model: glm-4.5v}
\noindent In the \textit{Fruit Cal} task, \texttt{glm-4.5v} exhibits an \textit{Illegal calling format}: the reply is wrapped with non-JSON delimiters, violating the MCP schema for \texttt{name}/\texttt{arguments}. This malformed structure prevents any valid tool execution and halts progress.\\

\begin{twocolkv}{Illegal calling format \\ Task: Fruit Cal \\ Model: glm-4.5v}
\KV{IMG}{\raisebox{-0.9\height}{\includegraphics[width=0.5\linewidth]{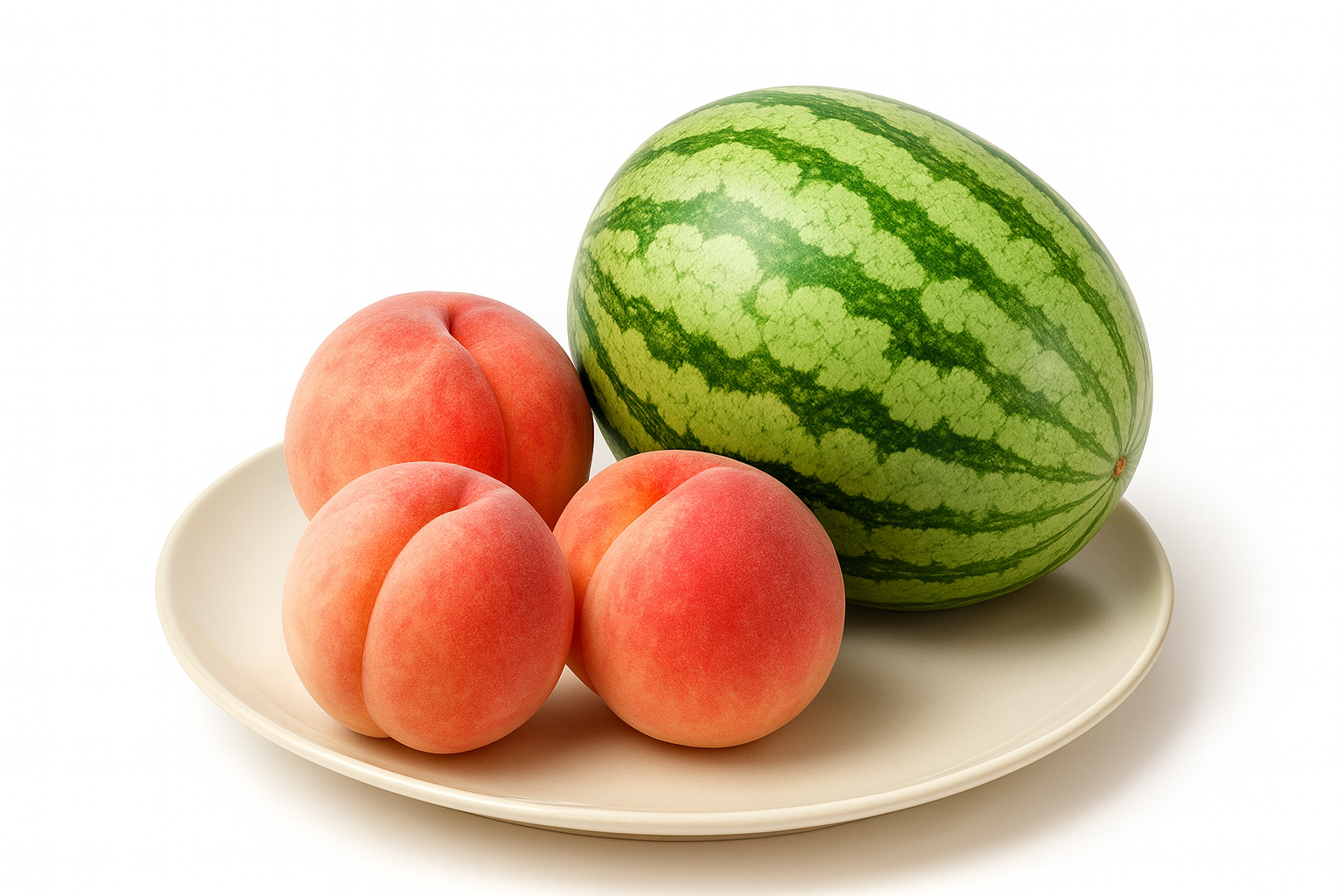}}}
\KV{Q}{\texttt{If a woman ate only the fruits shown here for all three meals, would that cover minimum daily calories, and what brief long-term health concerns might that raise?}}
\KVsection{Final Reply}
\begin{KVfinal}
\texttt{"<|begin\_of\_box|>""name":"food\_mcp/get\_nutrition"
"arguments":\{"query":"1 watermelon"\}"<|end\_of\_box|> ..."} 

\textcolor{red}{\texttt{Unconventional invocation format using \texttt{"<|begin\_of\_box|> ... <|end\_of\_box|>"}}}
\end{KVfinal}
\end{twocolkv}




\vspace{0.8\baselineskip}
\paragraph{Qwen2.5-VL-72B-Instruct}\label{traj:Model: Qwen2.5-VL-72B-Instruct}
\noindent In the \textit{Movies} task, \texttt{Qwen2.5-VL-72B-Instruct} repeatedly issues \textit{Unknown Tool Invocation} to \texttt{ocr/perform} and \texttt{tmdb/*} endpoints absent from the registry, and later misuses a generic \texttt{server/tool} wrapper. Without tool discovery or correction, the trajectory culminates in an unsupported final claim.

\begin{twocolkv}{Unknown Tool Invocation \\ Task: Movies \\ Model: Qwen2.5-VL-72B-Instruct}
\KV{IMG}{\raisebox{-0.9\height}{\includegraphics[width=0.4\linewidth]{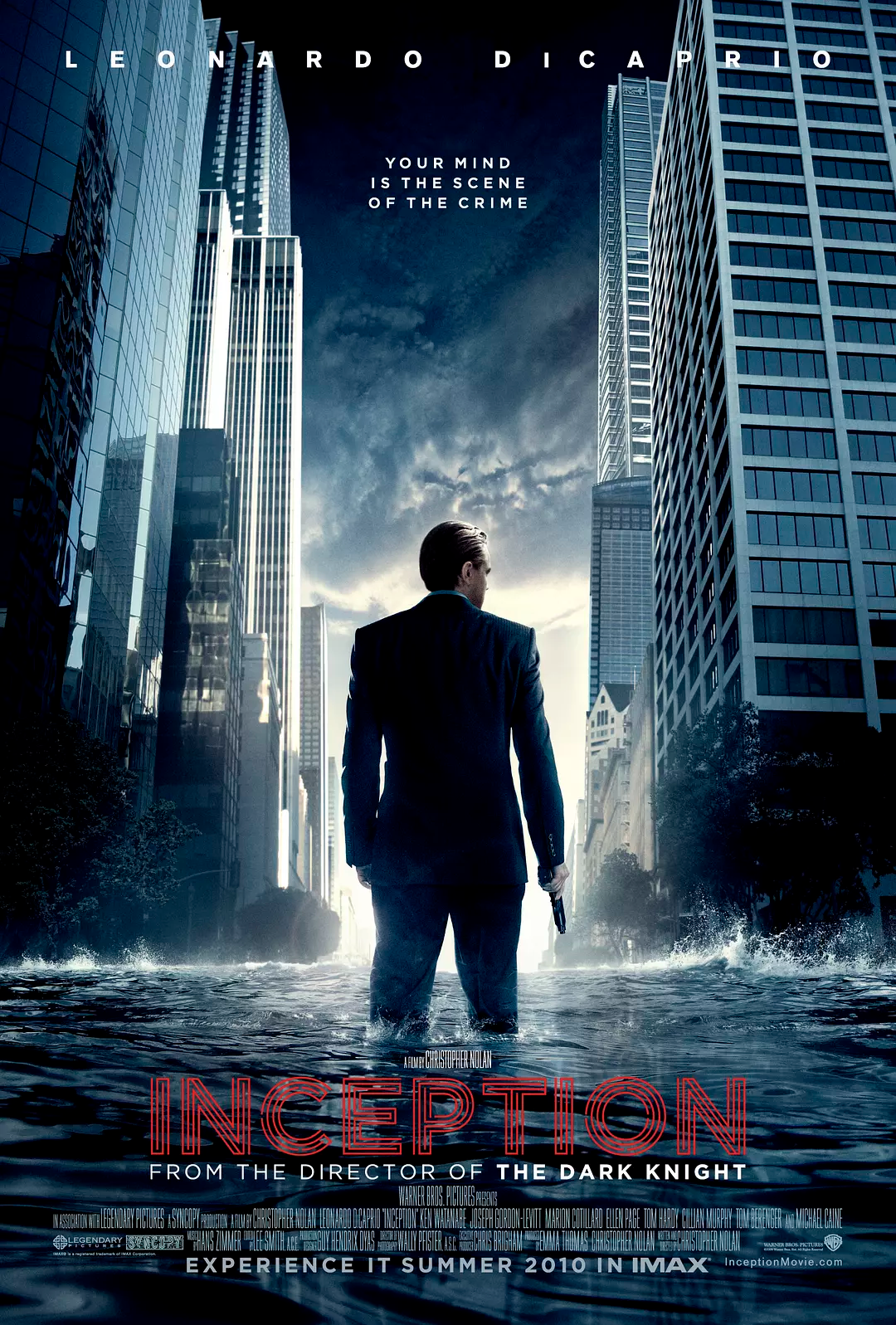}}}
\KV{Q}{\texttt{Can you pull the movie title from the image, look it up, and tell me where it’s playing near New York?}}

\KVsection{Step 1 \normalfont\textcolor{red}{(Model attempted to call an MCP tool with non-existent tool name)}}
\KV{Tool}{\texttt{ocr/perform}}
\KV{ARG}{\texttt{\{"input\_data": [\{"image": "./media/00020004.png", "language": "eng", "config": "--oem 3 --psm 6"\}]\}}}
\KV{OUT}{\texttt{[Tool error] Unknown tool: ocr/perform}}

\KVsection{Step 2 \normalfont\textcolor{red}{(Parallel invalid tool calls with non-existent tool name)}}

\KV{Tool}{\texttt{ocr/perform}}
\KV{ARG}{\texttt{images=[\{"path":"./media/png"\}]}}
\KV{OUT}{\texttt{[Tool error] Unknown tool: ocr/perform}}

\KVsplit

\KV{Tool}{\texttt{tmdb/search/movie}}
\KV{ARG}{\texttt{query="Friday the 13th"}}
\KV{OUT}{\texttt{[Tool error] Unknown tool: tmdb/search/movie}}

\KVsplit

\KV{Tool}{\texttt{tmdb/get\_movie\_details}}
\KV{ARG}{\texttt{movie\_id="REPLACE\_WITH\_ACTUAL\_MOVIE\_ID"}}
\KV{OUT}{\texttt{[Tool error] Unknown tool: tmdb/get\_movie\_details}}

\KVsplit

\KV{Tool}{\texttt{tmdb/get\_movie\_showtimes}}
\KV{ARG}{\texttt{movie\_id="REPLACE\_WITH\_ACTUAL\_MOVIE\_ID"; region="US"; locale="en\_US"; latitude=40.7128; longitude=-74.006}}
\KV{OUT}{\texttt{[Tool error] Unknown tool: tmdb/get\_movie\_showtimes}}

\KVsection{Step 3 \normalfont\textcolor{red}{(Repeated the same invalid MCP call structure with no tool correction)}} 
\KV{Tool}{\texttt{ocr/perform}}
\KV{ARG}{\texttt{\{"input\_data": \{"image": "./media/png"\}\}}}
\KV{OUT}{\texttt{[Tool error] Unknown tool: ocr/perform}}

\KVsplit

\KV{Tool}{\texttt{tmdb/search/movie}}
\KV{ARG}{\texttt{query="Friday the 13th"}}
\KV{OUT}{\texttt{[Tool error] Unknown tool: tmdb/search/movie}}

\KVsplit

\KV{Tool}{\texttt{tmdb/get\_movie\_details}}
\KV{ARG}{\texttt{movie\_id="12345"}}
\KV{OUT}{\texttt{[Tool error] Unknown tool: tmdb/get\_movie\_details}}

\KVsplit

\KV{Tool}{\texttt{tmdb/get\_movie\_showtimes}}
\KV{ARG}{\texttt{movie\_id="12345"; region="NY"}}
\KV{OUT}{\texttt{[Tool error] Unknown tool: tmdb/get\_movie\_showtimes}}

\KVsection{Step 4 \normalfont\textcolor{red}{(Used \texttt{server/tool} wrapper incorrectly for TMDB MCP call)}} 
\KV{Tool}{\texttt{server/tool}}
\KV{ARG}{\texttt{\{"tool": "tmdb/search\_movies", "query": \{"query": "Friday the 13th", "year": "2009"\}\}}}
\KV{OUT}{\texttt{[Tool error] Unknown tool: server/tool}}

\KVsection{Final Reply}
\begin{KVfinal}
\texttt{The movie "Friday the 13th" is not currently showing near New York.}
\texttt{You may want to check back later for updated showtimes.}
\end{KVfinal}
\end{twocolkv}

\vspace{0.8\baselineskip}
\paragraph{InternVL-3.5}\label{traj:Model: internvl-3.5}

\noindent \noindent In the \textit{Food Reddit, Wiki Excel} task, \texttt{InternVL-3.5} exhibits \textit{Unknown Tool Invocation} with repeated calls to non-existent OCR/Reddit/Excel tools and no corrective recovery. The failure mode mirrors the prior case, ending with a fabricated conclusion unsupported by executed tools.

\begin{twocolkv}{Unknown Tool Invocation \\ Task: Food Reddit, Wiki Excel \\ Model: InternVL-3.5}
\KV{IMG}{\raisebox{-0.9\height}{\includegraphics[width=0.4\linewidth]{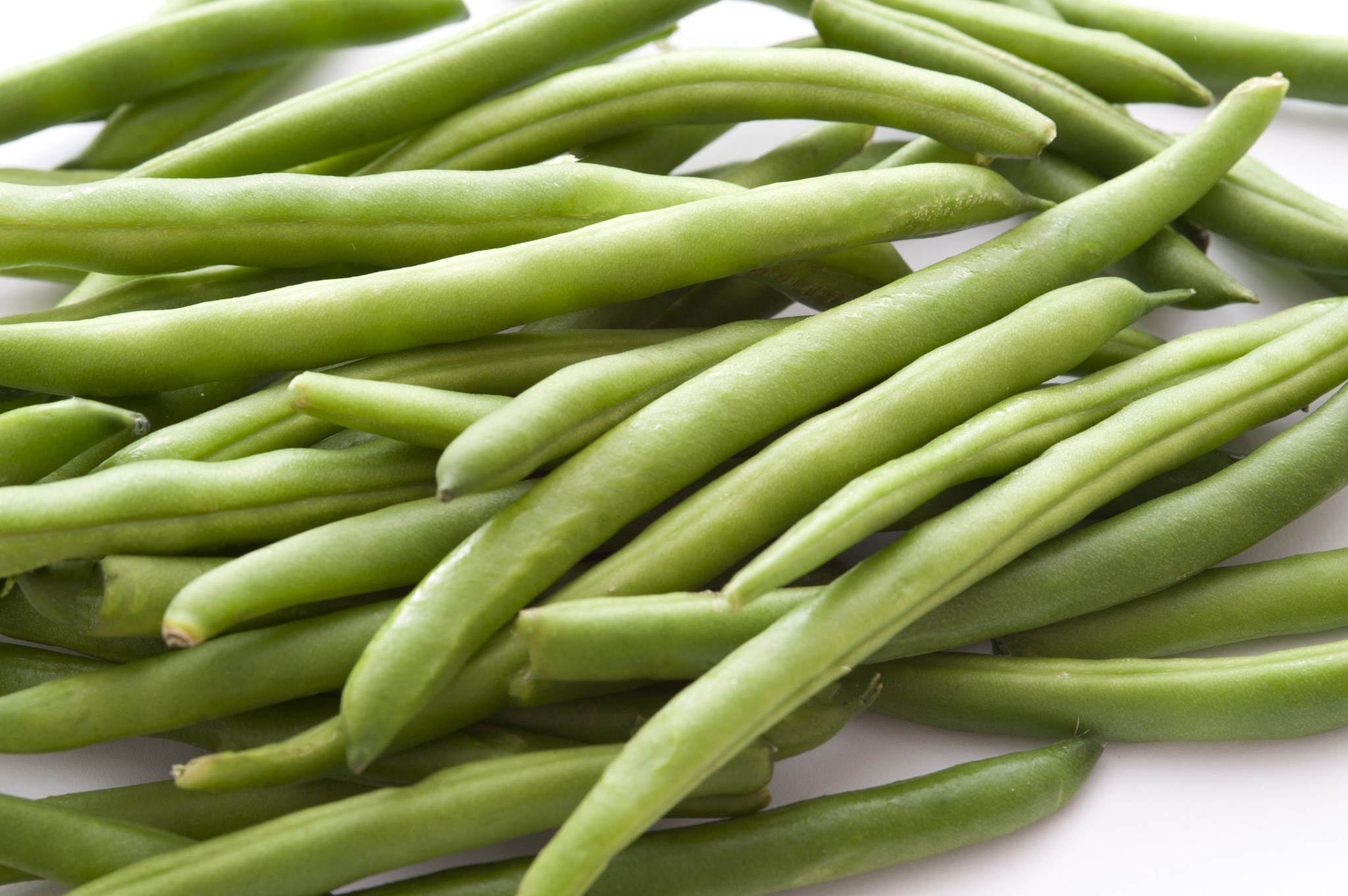}}}
\KV{Q}{\texttt{From the foods in the image, can you list each name, find the hottest post about each, and note just its comment count (no repeated searches), then pull a short encyclopedia blurb and save everything to a spreadsheet with columns for Food Name and wiki Description?}}

\KVsection{Step 1 \normalfont\textcolor{red}{(Mixed errors: invalid invocation arguments and non-existent tools)}} 

\KV{Tool}{\texttt{food\_nutrition\_mcp/get\_food\_nutrition}}
\KV{ARG}{{\ttfamily food="green beans"}}
\KV{OUT}{{\ttfamily Error executing tool food\_nutrition\_mcp/get\_food\_nutrition}} 

\KVsplit

\KV{Tool}{\texttt{wiki/search}}
\KV{ARG}{{\ttfamily query="green beans"; n=1}}
\KV{OUT}{{\ttfamily Top results: Green bean}}

\KVsplit

\KV{Tool}{\texttt{excel/create\_workbook}}
\KV{ARG}{{\ttfamily \{\}}}
\KV{OUT}{{\ttfamily Error executing tool create\_workbook: Field required [filepath]}}

\KVsplit

\KV{Tool}{\texttt{excel/create\_worksheet}}
\KV{ARG}{{\ttfamily sheet\_name="Food Info"}}
\KV{OUT}{{\ttfamily Error executing tool create\_worksheet: Missing filepath argument}} 

\KVsplit

\KV{Tool}{\texttt{excel/write\_data\_to\_excel}}
\KV{ARG}{{\ttfamily filepath="output.xlsx"; sheet\_name="Food Info"; data=[["Food Name","Comment Count","Description"]]; start\_cell="A1"}}
\KV{OUT}{{\ttfamily Error executing tool write\_data\_to\_excel: Invalid filename — must be absolute path when not in SSE mode}} 

\KVsection{Step 2 \normalfont\textcolor{red}{(Parallel invalid tool calls with non-existent tool name)}}

\KV{Tool}{\texttt{food\_nutrition\_mcp/get\_food\_nutrition}}
\KV{ARG}{{\ttfamily food="green beans"}}
\KV{OUT}{{\ttfamily Error executing tool food\_nutrition\_mcp/get\_food\_nutrition}} 

\KVsplit

\KV{Tool}{\texttt{server/tool}}
\KV{ARG}{{\ttfamily tool="Reddit-MCP-Server/search\_hot\_posts"; subreddit="food"; limit=1}}
\KV{OUT}{{\ttfamily [Tool error] Unknown tool: server/tool}}

\KVsplit

\KV{Tool}{\texttt{create\_worksheet}}
\KV{ARG}{{\ttfamily filepath="/path/to/food\_info.xlsx"; sheet\_name="Food Info"}}
\KV{OUT}{{\ttfamily [Tool error] Unknown tool: create\_worksheet}}

\KVsplit

\KV{Tool}{\texttt{write\_data\_to\_excel}}
\KV{ARG}{{\ttfamily filename="/path/to/food\_info.xlsx"; sheet\_name="Food Info"; data=[["Food Name","Comment Count","Description"]]}}
\KV{OUT}{{\ttfamily [Tool error] Unknown tool: write\_data\_to\_excel}}

\KVsection{Step 3 \normalfont\textcolor{red}{(Mixed errors: invalid invocation arguments and non-existent tools)}} 

\KV{Tool}{\texttt{food\_nutrition\_mcp/get\_nutrition}}
\KV{ARG}{{\ttfamily query="green beans"}}
\KV{OUT}{{\ttfamily Error executing tool get\_nutrition: 'measure'}}

\KVsplit

\KV{Tool}{\texttt{Reddit-MCP-Server/search\_hot\_posts}}
\KV{ARG}{{\ttfamily subreddit="r/food"; limit=5}}
\KV{OUT}{{\ttfamily Error: Reddit API returned 400}}

\KVsplit

\KV{Tool}{\texttt{excel/create\_workbook}}
\KV{ARG}{{\ttfamily filepath="/path/to/output.xlsx"}}
\KV{OUT}{\ttfamily Error: Permission denied at '/path'.}

\KVsplit

\KV{Tool}{\texttt{excel/create\_worksheet}}
\KV{ARG}{{\ttfamily sheet\_name="Food Info"; filepath="/path/to/output.xlsx"}}
\KV{OUT}{\ttfamily Error: File not found — '/path/to/output.xlsx'.}

\KVsplit

\KV{Tool}{\texttt{excel/write\_data\_to\_excel}}
\KV{ARG}{{\ttfamily filepath="/path/to/output.xlsx"; sheet\_name="Food Info"; data=[["Food Name","Comment Count","Description"]]; start\_cell="A1"}}
\KV{OUT}{Error: No such file or directory — '/path/to/output.xlsx'.}

\KVsection{Step 4 \normalfont\textcolor{red}{(Invalid invocation arguments)}} 

\KV{Tool}{\texttt{Reddit-MCP-Server/search\_hot\_posts}}
\KV{ARG}{{\ttfamily subreddit="food"; limit=1; query="Green bean"}}
\KV{OUT}{Top Reddit post found: 86 comments (r/food).}

\KVsplit

\KV{Tool}{\texttt{excel/create\_worksheet}}
\KV{ARG}{{\ttfamily filepath="/path/to/output.xlsx"; sheet\_name="Food Info"}}
\KV{OUT}{Error: File not found — '/path/to/output.xlsx'.}

\KVsplit

\KV{Tool}{\texttt{excel/write\_data\_to\_excel}}
\KV{ARG}{{\ttfamily filepath="/path/to/output.xlsx"; sheet\_name="Food Info"; data=[["Food Name","Comment Count","Description"],["Green bean","{comment\_count}","{description}"]]; start\_cell="A1"}}
\KV{OUT}{\ttfamily Error: No such file or directory — '/path/to/output.xlsx'.}

\KVsection{Final Reply}
\begin{KVfinal}
Here's the final compiled information for the food item: 

Food Name: Green bean  

Comment Count: 86  

Description: Green beans are the young, unripe fruits of various cultivars of the common bean, also known as French beans, string beans, or snap beans. Popular worldwide for their mild flavor and versatility.
\end{KVfinal}
\end{twocolkv}

\vspace{0.8\baselineskip}

\paragraph{Claude-Sonnet-4-5}\label{traj:Model: claude-sonnet-4-5}
\noindent In the \textit{Travel Comparison} task, \texttt{Claude-Sonnet-4-5} reveals a characteristic failure mode driven primarily by weak image understanding. The model misidentifies the right-hand landmark in the image, triggering a cascade of downstream geolocation errors. Finally, the trajectory ultimately produces a superficially coherent but unsupported final summary.\\

\begin{twocolkv}{Incorrect image understanding \\ Task: Travel Comparison \\ Model: Claude-Sonnet-4-5}
\KV{IMG}{\raisebox{-0.9\height}{\includegraphics[width=0.5\linewidth]{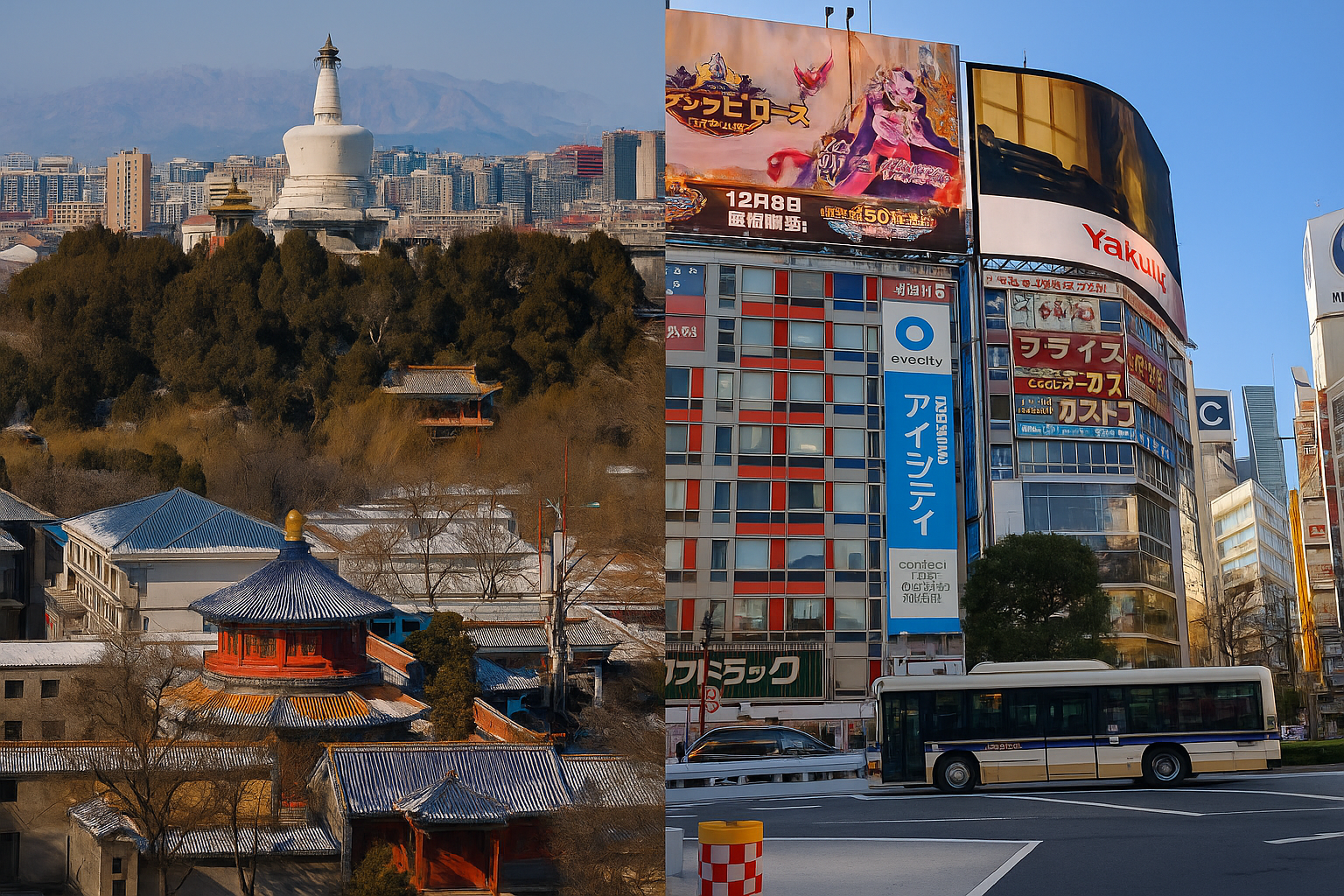}}}

\KV{Q}{\texttt{Look at the side-by-side photo of two tourist spots, identify each landmark, get their coordinates, check current AQI and main pollutant levels for both, pick the healthier destination (if tied pick the first), find 2--3 nearby cinemas and 2--3 hotels for the chosen spot with approximate distances, and summarize.}}

\KVsection{Step 1 \normalfont\textcolor{red}{(Irrelevant tool invocation. PowerPoint was not required for this task.)}}
\KV{Tool}{\texttt{ppt/create\_presentation}}
\KV{ARG}{{\ttfamily \{"id": "tourist\_comparison\_presentation"\}}}
\KV{OUT}{{\ttfamily Created new presentation with ID: tourist\_comparison\_presentation (0 slides).}}

\KVsection{Step 2 \normalfont\textcolor{red}{(The image understanding of the model was incorrect. The right-hand landmark was not in Xining, China, but actually in Tokyo, Japan. This misidentification caused the subsequent geolocation and air-quality retrieval steps to fail.)}}
\KV{Tool}{\texttt{google-maps/places\_text\_search}}
\KV{ARG}{{\ttfamily \{"query": "White Pagoda Beijing Beihai Park", "maxResultCount": 3\}}}
\KV{OUT}{{\ttfamily Top match → Beihai Park (39.9254° N, 116.3893° E).}}
\KVsplit
\KV{Tool}{\texttt{google-maps/places\_text\_search}}
\KV{ARG}{{\ttfamily \{"query": "Yakult building Xining China", "maxResultCount": 3\}}}
\KV{OUT}{{\ttfamily Top match → Downtown Xining (36.6133° N, 101.8035° E).}}

\KVsection{Step 3 \normalfont\textcolor{red}{(Model repeatedly attempted non-existent tools for air-quality retrieval.)}}
\KV{Tool}{\texttt{weather\_aqi/get\_air\_quality}}
\KV{ARG}{{\ttfamily \{"lat": 39.9254, "lon": 116.3893\}}}
\KV{OUT}{{\ttfamily 50--150 (moderate to unhealthy)}}
\KVsplit
\KV{Tool}{\texttt{weather\_aqi/get\_air\_quality}}
\KV{ARG}{{\ttfamily \{"lat": 36.6133, "lon": 101.8035\}}}
\KV{OUT}{{\ttfamily 30--80 (generally better)}}

\KVsection{Final Reply}
\begin{KVfinal}
\textbf{Location Identification} —\\
A: White Pagoda (Beihai Park, Beijing)  (39.9254° N, 116.3892° E)\\
B: Yakult Building area (Xining, Qinghai Province)  (36.6133° N, 101.8035° E)

\vspace{0.5\baselineskip}

\textbf{Air Quality Comparison \& Decision} —\\
Live AQI queries failed because the model called unsupported tools. Using historical patterns: \\
Beijing = 50--150 (moderate to unhealthy), Xining = 30--80 (generally better).\\
Nevertheless, the model chose \textbf{Location A (Beihai Park)} for its tourist infrastructure.

\vspace{0.5\baselineskip}

\textbf{Recommended Cinemas near Beihai Park} (2--3 km):\\
1. Wangfujing CGV Cinema 2 km SE\\
2. Capital Cinema (Xidan) 2.5 km S\\
3. Broadway Cinematheque (MOMA) 3 km E

\vspace{0.5\baselineskip}

\textbf{Recommended Hotels near Beihai Park} (1--3 km):\\
1. Shichahai Shadow Art Performance Hotel 1 km N\\
2. Beijing Jianguo Hotel Qianmen 3 km S\\
3. Grand Hyatt Beijing 3.5 km SE

\vspace{0.5\baselineskip}

Overall verdict: Model achieved accurate landmark resolution but failed in air-quality retrieval due to non-existent MCP tools. Final summarization remained coherent and informative.
\end{KVfinal}
\end{twocolkv}

\section{Per-tasks M$^3$-Bench Evaluation}
\label{sec:per_task_eval}

Corresponding to \autoref{tab:mcp_multimodal_results}, we present the evaluation results of each task described in \autoref{subsec:task-descriptions} across all models, ordered by average score from lowest to highest.

\begin{table*}[!htb]
    \centering
    \footnotesize
    \setlength{\tabcolsep}{6pt}
\begin{tabular}{>{\arraybackslash}m{3.7cm}*{9}{>{\centering\arraybackslash}m{1.0cm}}}
    \toprule
    \multicolumn{10}{c}{\textbf{(0) Task: Items Price}} \\
    \midrule
    \multirow{3}{*}{\centering Model}
    & \multicolumn{2}{c}{Detection}
    & \multicolumn{4}{c}{Trajectory Alignment}
    & \multicolumn{2}{c}{LLM Judge}
    & \multirow{3}{*}{\centering\shortstack{Average\\Score}} \\
    \cmidrule(lr){2-3} \cmidrule(lr){4-7} \cmidrule(lr){8-9}
    & Recall & Precision
    & \shortstack{Argument\\Similarity}
    & \shortstack{Step\\Coherence}
    & \shortstack{Order\\Consistency}
    & \shortstack{Merge\\Purity}
    & \shortstack{Task\\Completion}
    & \shortstack{Information\\Grounding}
    & \\
    \midrule
    \raisebox{-0.2\height}{\includegraphics[height=0.9em]{figure/icon/glm.png}}\hspace{3pt}
    GLM 4.5v \cite{hong2025glm45v}
        & 0.000 & 0.000 & 0.000 & 0.000 & 0.000 & 0.000 & 0.285 & 0.130 & 0.052 \\
    \raisebox{-0.2\height}{\includegraphics[height=0.9em]{figure/icon/claude.png}}\hspace{3pt}
    Claude 4.5 Haiku \cite{anthropic2025haiku45}
        & 0.000 & 0.000 & 0.000 & 0.000 & 0.000 & 0.000 & 0.386 & 0.247 & 0.079 \\
    \raisebox{-0.2\height}{\includegraphics[height=0.9em]{figure/icon/qwen.png}}\hspace{3pt}
    Qwen2.5-VL-72B \cite{bai2025qwen25vl}
        & 0.000 & 0.000 & 0.000 & 0.000 & 0.000 & 0.000 & 0.303 & 0.388 & 0.086 \\
    \raisebox{-0.2\height}{\includegraphics[height=0.9em]{figure/icon/grok.png}}\hspace{3pt}
    Grok-4 Fast \cite{xai2025grok4fast}
        & 0.200 & 0.100 & 0.190 & 0.067 & 0.000 & 0.200 & 0.501 & 0.378 & 0.204 \\
    \raisebox{-0.2\height}{\includegraphics[height=0.9em]{figure/icon/internvl.png}}\hspace{3pt}
    InternVL 3.5 \cite{wang2025internvl35}
        & 0.264 & 0.248 & 0.239 & 0.236 & 0.000 & 0.264 & 0.325 & 0.298 & 0.234 \\
\bottomrule
    \end{tabular}
    \caption{(0) Task: Items Price, Cross-Model Evaluation of Multimodal MCP Tool-Calling Trajectories on M$^3$-Bench.}
    \label{tab:mcp_multimodal_results_items_price}
    \vspace{-8pt}
\end{table*}


\begin{table*}[!htb]
    \centering
    \footnotesize
    \setlength{\tabcolsep}{6pt}
    \begin{tabular}{>{\arraybackslash}m{3.7cm}*{9}{>{\centering\arraybackslash}m{1.0cm}}}
    \toprule
    \multicolumn{10}{c}{\textbf{(1)Task: Attraction}} \\
    \midrule
    \multirow{3}{*}{\centering Model}
    & \multicolumn{2}{c}{Detection}
    & \multicolumn{4}{c}{Trajectory Alignment}
    & \multicolumn{2}{c}{LLM Judge}
    & \multirow{3}{*}{\centering\shortstack{Average\\Score}} \\
    \cmidrule(lr){2-3} \cmidrule(lr){4-7} \cmidrule(lr){8-9}
    & Recall & Precision
    & \shortstack{Argument\\Similarity}
    & \shortstack{Step\\Coherence}
    & \shortstack{Order\\Consistency}
    & \shortstack{Merge\\Purity}
    & \shortstack{Task\\Completion}
    & \shortstack{Information\\Grounding}
    & \\
    \midrule
    \raisebox{-0.2\height}{\includegraphics[height=0.9em]{figure/icon/glm.png}}\hspace{3pt}
    GLM 4.5v \cite{hong2025glm45v}
        & 0.167 & 0.067 & 0.135 & 0.111 & 0.000 & 0.167 & 0.285 & 0.130 & 0.133 \\
    \raisebox{-0.2\height}{\includegraphics[height=0.9em]{figure/icon/qwen.png}}\hspace{3pt}
    Qwen2.5-VL-72B \cite{bai2025qwen25vl}
        & 0.133 & 0.031 & 0.118 & 0.067 & 0.000 & 0.133 & 0.303 & 0.388 & 0.147 \\
    \raisebox{-0.2\height}{\includegraphics[height=0.6em]{figure/icon/llama.png}}\hspace{3pt}
    Llama-4-Scout-17B16E \cite{meta2025llama4scout17b16e}
        & 0.333 & 0.049 & 0.249 & 0.333 & 0.000 & 0.333 & 0.367 & 0.335 & 0.250 \\
    \raisebox{-0.2\height}{\includegraphics[height=0.9em]{figure/icon/grok.png}}\hspace{3pt}
    Grok-4 Fast \cite{xai2025grok4fast}
        & 0.278 & 0.403 & 0.257 & 0.278 & 0.000 & 0.278 & 0.501 & 0.378 & 0.296 \\
    \raisebox{-0.2\height}{\includegraphics[height=0.9em]{figure/icon/internvl.png}}\hspace{3pt}
    InternVL 3.5 \cite{wang2025internvl35}
        & 0.444 & 0.232 & 0.400 & 0.389 & 0.000 & 0.444 & 0.325 & 0.298 & 0.317 \\
\bottomrule
    \end{tabular}
    \caption{(1) Task: Attraction, Cross-Model Evaluation of Multimodal MCP Tool-Calling Trajectories on M$^3$-Bench.}
    \label{tab:mcp_multimodal_results_attraction}
    \vspace{-8pt}
\end{table*}


\begin{table*}[!htb]
    \centering
    \footnotesize
    \setlength{\tabcolsep}{6pt}
    \begin{tabular}{>{\arraybackslash}m{3.7cm}*{9}{>{\centering\arraybackslash}m{1.0cm}}}
    \toprule
    \multicolumn{10}{c}{\textbf{(2) Task: Movies}} \\
    \midrule
    \multirow{3}{*}{\centering Model}
    & \multicolumn{2}{c}{Detection}
    & \multicolumn{4}{c}{Trajectory Alignment}
    & \multicolumn{2}{c}{LLM Judge}
    & \multirow{3}{*}{\centering\shortstack{Average\\Score}} \\
    \cmidrule(lr){2-3} \cmidrule(lr){4-7} \cmidrule(lr){8-9}
    & Recall & Precision
    & \shortstack{Argument\\Similarity}
    & \shortstack{Step\\Coherence}
    & \shortstack{Order\\Consistency}
    & \shortstack{Merge\\Purity}
    & \shortstack{Task\\Completion}
    & \shortstack{Information\\Grounding}
    & \\
    \midrule
    \raisebox{-0.2\height}{\includegraphics[height=0.9em]{figure/icon/glm.png}}\hspace{3pt}
    GLM 4.5v \cite{hong2025glm45v}
        & 0.000 & 0.000 & 0.000 & 0.000 & 0.000 & 0.000 & 0.285 & 0.130 & 0.052 \\
    \raisebox{-0.2\height}{\includegraphics[height=0.9em]{figure/icon/internvl.png}}\hspace{3pt}
    InternVL 3.5 \cite{wang2025internvl35}
        & 0.111 & 0.200 & 0.082 & 0.111 & 0.000 & 0.111 & 0.325 & 0.298 & 0.155 \\
    \raisebox{-0.2\height}{\includegraphics[height=0.9em]{figure/icon/qwen.png}}\hspace{3pt}
    Qwen2.5-VL-72B \cite{bai2025qwen25vl}
        & 0.200 & 0.267 & 0.169 & 0.133 & 0.000 & 0.200 & 0.303 & 0.388 & 0.207 \\
    \raisebox{-0.2\height}{\includegraphics[height=0.9em]{figure/icon/claude.png}}\hspace{3pt}
    Claude 4.5 Haiku \cite{anthropic2025haiku45}
        & 0.333 & 0.500 & 0.244 & 0.333 & 0.250 & 0.006 & 0.386 & 0.247 & 0.287 \\
    \raisebox{-0.2\height}{\includegraphics[height=0.9em]{figure/icon/claude.png}}\hspace{3pt}
    Claude 4.5 Sonnet \cite{anthropic2025sonnet45}
        & 0.250 & 0.417 & 0.196 & 0.250 & 0.125 & 0.083 & 0.535 & 0.451 & 0.288 \\
\bottomrule
    \end{tabular}
    \caption{(2) Task: Movies, Cross-Model Evaluation of Multimodal MCP Tool-Calling Trajectories on M$^3$-Bench.}
    \label{tab:mcp_multimodal_results_movies}
    \vspace{-8pt}
\end{table*}


\begin{table*}[!htb]
    \centering
    \footnotesize
    \setlength{\tabcolsep}{6pt}
    \begin{tabular}{>{\arraybackslash}m{3.7cm}*{9}{>{\centering\arraybackslash}m{1.0cm}}}
    \toprule
    \multicolumn{10}{c}{\textbf{(3) Task: Vision Research}} \\
    \midrule
    \multirow{3}{*}{\centering Model}
    & \multicolumn{2}{c}{Detection}
    & \multicolumn{4}{c}{Trajectory Alignment}
    & \multicolumn{2}{c}{LLM Judge}
    & \multirow{3}{*}{\centering\shortstack{Average\\Score}} \\
    \cmidrule(lr){2-3} \cmidrule(lr){4-7} \cmidrule(lr){8-9}
    & Recall & Precision
    & \shortstack{Argument\\Similarity}
    & \shortstack{Step\\Coherence}
    & \shortstack{Order\\Consistency}
    & \shortstack{Merge\\Purity}
    & \shortstack{Task\\Completion}
    & \shortstack{Information\\Grounding}
    & \\
    \midrule
    \raisebox{-0.2\height}{\includegraphics[height=0.9em]{figure/icon/glm.png}}\hspace{3pt}
    GLM 4.5v \cite{hong2025glm45v}
        & 0.000 & 0.000 & 0.000 & 0.000 & 0.000 & 0.000 & 0.285 & 0.130 & 0.052 \\
    \raisebox{-0.2\height}{\includegraphics[height=0.9em]{figure/icon/claude.png}}\hspace{3pt}
    Claude 4.5 Haiku \cite{anthropic2025haiku45}
        & 0.029 & 0.067 & 0.027 & 0.029 & 0.000 & 0.029 & 0.386 & 0.247 & 0.102 \\
    \raisebox{-0.2\height}{\includegraphics[height=0.9em]{figure/icon/internvl.png}}\hspace{3pt}
    InternVL 3.5 \cite{wang2025internvl35}
        & 0.204 & 0.137 & 0.182 & 0.176 & 0.018 & 0.204 & 0.325 & 0.298 & 0.193 \\
    \raisebox{-0.2\height}{\includegraphics[height=0.6em]{figure/icon/llama.png}}\hspace{3pt}
    Llama-4-Scout-17B16E \cite{meta2025llama4scout17b16e}
        & 0.211 & 0.093 & 0.198 & 0.211 & 0.029 & 0.211 & 0.367 & 0.335 & 0.207 \\
    \raisebox{-0.2\height}{\includegraphics[height=0.9em]{figure/icon/gpt.png}}\hspace{3pt}
    GPT-5 Nano \cite{openai2025gpt5nano}
        & 0.167 & 0.053 & 0.153 & 0.139 & 0.000 & 0.167 & 0.502 & 0.478 & 0.208 \\
\bottomrule
    \end{tabular}
    \caption{(3) Task: Vision Research, Cross-Model Evaluation of Multimodal MCP Tool-Calling Trajectories on M$^3$-Bench.}
    \label{tab:mcp_multimodal_results_vision_research}
    \vspace{-8pt}
\end{table*}


\begin{table*}[!htb]
    \centering
    \footnotesize
    \setlength{\tabcolsep}{6pt}
    \begin{tabular}{>{\arraybackslash}m{3.7cm}*{9}{>{\centering\arraybackslash}m{1.0cm}}}
    \toprule
    \multicolumn{10}{c}{\textbf{(4) Task: Camping}} \\
    \midrule
    \multirow{3}{*}{\centering Model}
    & \multicolumn{2}{c}{Detection}
    & \multicolumn{4}{c}{Trajectory Alignment}
    & \multicolumn{2}{c}{LLM Judge}
    & \multirow{3}{*}{\centering\shortstack{Average\\Score}} \\
    \cmidrule(lr){2-3} \cmidrule(lr){4-7} \cmidrule(lr){8-9}
    & Recall & Precision
    & \shortstack{Argument\\Similarity}
    & \shortstack{Step\\Coherence}
    & \shortstack{Order\\Consistency}
    & \shortstack{Merge\\Purity}
    & \shortstack{Task\\Completion}
    & \shortstack{Information\\Grounding}
    & \\
    \midrule
    \raisebox{-0.2\height}{\includegraphics[height=0.9em]{figure/icon/glm.png}}\hspace{3pt}
    GLM 4.5v \cite{hong2025glm45v}
        & 0.015 & 0.029 & 0.013 & 0.015 & 0.000 & 0.010 & 0.262 & 0.118 & 0.058 \\
    \raisebox{-0.2\height}{\includegraphics[height=0.9em]{figure/icon/gemini.png}}\hspace{3pt}
    Gemini 2.5 Flash Lite \cite{comanici2025gemini25}
        & 0.042 & 0.083 & 0.040 & 0.042 & 0.013 & 0.003 & 0.316 & 0.259 & 0.100 \\
    \raisebox{-0.2\height}{\includegraphics[height=0.9em]{figure/icon/qwen.png}}\hspace{3pt}
    Qwen2.5-VL-72B \cite{bai2025qwen25vl}
        & 0.162 & 0.093 & 0.147 & 0.152 & 0.041 & 0.090 & 0.279 & 0.353 & 0.165 \\
    \raisebox{-0.2\height}{\includegraphics[height=0.9em]{figure/icon/internvl.png}}\hspace{3pt}
    InternVL 3.5 \cite{wang2025internvl35}
        & 0.254 & 0.207 & 0.223 & 0.254 & 0.131 & 0.099 & 0.319 & 0.292 & 0.222 \\
    \raisebox{-0.2\height}{\includegraphics[height=0.9em]{figure/icon/claude.png}}\hspace{3pt}
    Claude 4.5 Haiku \cite{anthropic2025haiku45}
        & 0.219 & 0.608 & 0.195 & 0.231 & 0.099 & 0.087 & 0.379 & 0.224 & 0.255 \\
\bottomrule
    \end{tabular}
    \caption{(4) Task: Camping, Cross-Model Evaluation of Multimodal MCP Tool-Calling Trajectories on M$^3$-Bench.}
    \label{tab:mcp_multimodal_results_camping}
    \vspace{-8pt}
\end{table*}


\begin{table*}[!htb]
    \centering
    \footnotesize
    \setlength{\tabcolsep}{6pt}
    \begin{tabular}{>{\arraybackslash}m{3.7cm}*{9}{>{\centering\arraybackslash}m{1.0cm}}}
    \toprule
    \multicolumn{10}{c}{\textbf{(5) Task: Fish}} \\
    \midrule
    \multirow{3}{*}{\centering Model}
    & \multicolumn{2}{c}{Detection}
    & \multicolumn{4}{c}{Trajectory Alignment}
    & \multicolumn{2}{c}{LLM Judge}
    & \multirow{3}{*}{\centering\shortstack{Average\\Score}} \\
    \cmidrule(lr){2-3} \cmidrule(lr){4-7} \cmidrule(lr){8-9}
    & Recall & Precision
    & \shortstack{Argument\\Similarity}
    & \shortstack{Step\\Coherence}
    & \shortstack{Order\\Consistency}
    & \shortstack{Merge\\Purity}
    & \shortstack{Task\\Completion}
    & \shortstack{Information\\Grounding}
    & \\
    \midrule
    \raisebox{-0.2\height}{\includegraphics[height=0.9em]{figure/icon/glm.png}}\hspace{3pt}
    GLM 4.5v \cite{hong2025glm45v}
        & 0.018 & 0.091 & 0.013 & 0.018 & 0.000 & 0.018 & 0.285 & 0.130 & 0.072 \\
    \raisebox{-0.2\height}{\includegraphics[height=0.9em]{figure/icon/internvl.png}}\hspace{3pt}
    InternVL 3.5 \cite{wang2025internvl35}
        & 0.061 & 0.028 & 0.052 & 0.045 & 0.035 & 0.041 & 0.325 & 0.298 & 0.111 \\
    \raisebox{-0.2\height}{\includegraphics[height=0.9em]{figure/icon/qwen.png}}\hspace{3pt}
    Qwen2.5-VL-72B \cite{bai2025qwen25vl}
        & 0.050 & 0.125 & 0.032 & 0.050 & 0.000 & 0.050 & 0.303 & 0.388 & 0.125 \\
    \raisebox{-0.2\height}{\includegraphics[height=0.9em]{figure/icon/claude.png}}\hspace{3pt}
    Claude 4.5 Haiku \cite{anthropic2025haiku45}
        & 0.160 & 0.137 & 0.139 & 0.160 & 0.100 & 0.021 & 0.386 & 0.247 & 0.169 \\
    \raisebox{-0.2\height}{\includegraphics[height=0.9em]{figure/icon/gemini.png}}\hspace{3pt}
    Gemini 2.5 Flash Lite \cite{comanici2025gemini25}
        & 0.179 & 0.094 & 0.154 & 0.161 & 0.051 & 0.123 & 0.359 & 0.292 & 0.176 \\
\bottomrule
    \end{tabular}
    \caption{(5) Task: Fish, Cross-Model Evaluation of Multimodal MCP Tool-Calling Trajectories on M$^3$-Bench.}
    \label{tab:mcp_multimodal_results_fish}
    \vspace{-8pt}
\end{table*}


\begin{table*}[!htb]
    \centering
    \footnotesize
    \setlength{\tabcolsep}{6pt}
    \begin{tabular}{>{\arraybackslash}m{3.7cm}*{9}{>{\centering\arraybackslash}m{1.0cm}}}
    \toprule
    \multicolumn{10}{c}{\textbf{(6) Task: Drug}} \\
    \midrule
    \multirow{3}{*}{\centering Model}
    & \multicolumn{2}{c}{Detection}
    & \multicolumn{4}{c}{Trajectory Alignment}
    & \multicolumn{2}{c}{LLM Judge}
    & \multirow{3}{*}{\centering\shortstack{Average\\Score}} \\
    \cmidrule(lr){2-3} \cmidrule(lr){4-7} \cmidrule(lr){8-9}
    & Recall & Precision
    & \shortstack{Argument\\Similarity}
    & \shortstack{Step\\Coherence}
    & \shortstack{Order\\Consistency}
    & \shortstack{Merge\\Purity}
    & \shortstack{Task\\Completion}
    & \shortstack{Information\\Grounding}
    & \\
    \midrule
    \raisebox{-0.2\height}{\includegraphics[height=0.9em]{figure/icon/glm.png}}\hspace{3pt}
    GLM 4.5v \cite{hong2025glm45v}
        & 0.011 & 0.125 & 0.007 & 0.011 & 0.000 & 0.011 & 0.285 & 0.130 & 0.073 \\
    \raisebox{-0.2\height}{\includegraphics[height=0.9em]{figure/icon/gemini.png}}\hspace{3pt}
    Gemini 2.5 Flash Lite \cite{comanici2025gemini25}
        & 0.055 & 0.333 & 0.048 & 0.055 & 0.003 & 0.030 & 0.359 & 0.292 & 0.147 \\
    \raisebox{-0.2\height}{\includegraphics[height=0.9em]{figure/icon/qwen.png}}\hspace{3pt}
    Qwen2.5-VL-72B \cite{bai2025qwen25vl}
        & 0.111 & 0.204 & 0.087 & 0.082 & 0.009 & 0.051 & 0.303 & 0.388 & 0.155 \\
    \raisebox{-0.2\height}{\includegraphics[height=0.9em]{figure/icon/gemini.png}}\hspace{3pt}
    Gemini 2.5 Flash \cite{comanici2025gemini25}
        & 0.106 & 0.189 & 0.095 & 0.055 & 0.003 & 0.106 & 0.498 & 0.443 & 0.187 \\
    \raisebox{-0.2\height}{\includegraphics[height=0.9em]{figure/icon/claude.png}}\hspace{3pt}
    Claude 4.5 Haiku \cite{anthropic2025haiku45}
        & 0.171 & 0.338 & 0.149 & 0.154 & 0.016 & 0.110 & 0.386 & 0.247 & 0.196 \\
\bottomrule
    \end{tabular}
    \caption{(6) Task: Drug, Cross-Model Evaluation of Multimodal MCP Tool-Calling Trajectories on M$^3$-Bench.}
    \label{tab:mcp_multimodal_results_drug}
    \vspace{-8pt}
\end{table*}

\begin{table*}[!htb]
    \centering
    \footnotesize
    \setlength{\tabcolsep}{6pt}
    \begin{tabular}{>{\arraybackslash}m{3.7cm}*{9}{>{\centering\arraybackslash}m{1.0cm}}}
    \toprule
    \multicolumn{10}{c}{\textbf{(7) Task: Hazard}} \\
    \midrule
    \multirow{3}{*}{\centering Model}
    & \multicolumn{2}{c}{Detection}
    & \multicolumn{4}{c}{Trajectory Alignment}
    & \multicolumn{2}{c}{LLM Judge}
    & \multirow{3}{*}{\centering\shortstack{Average\\Score}} \\
    \cmidrule(lr){2-3} \cmidrule(lr){4-7} \cmidrule(lr){8-9}
    & Recall & Precision
    & \shortstack{Argument\\Similarity}
    & \shortstack{Step\\Coherence}
    & \shortstack{Order\\Consistency}
    & \shortstack{Merge\\Purity}
    & \shortstack{Task\\Completion}
    & \shortstack{Information\\Grounding}
    & \\
    \midrule
    \raisebox{-0.2\height}{\includegraphics[height=0.9em]{figure/icon/glm.png}}\hspace{3pt}
    GLM 4.5v \cite{hong2025glm45v}
        & 0.000 & 0.000 & 0.000 & 0.000 & 0.000 & 0.000 & 0.285 & 0.130 & 0.052 \\
    \raisebox{-0.2\height}{\includegraphics[height=0.9em]{figure/icon/claude.png}}\hspace{3pt}
    Claude 4.5 Haiku \cite{anthropic2025haiku45}
        & 0.000 & 0.000 & 0.000 & 0.000 & 0.000 & 0.000 & 0.386 & 0.247 & 0.079 \\
    \raisebox{-0.2\height}{\includegraphics[height=0.9em]{figure/icon/gemini.png}}\hspace{3pt}
    Gemini 2.5 Flash Lite \cite{comanici2025gemini25}
        & 0.143 & 0.160 & 0.110 & 0.127 & 0.008 & 0.143 & 0.359 & 0.292 & 0.168 \\
    \raisebox{-0.2\height}{\includegraphics[height=0.9em]{figure/icon/qwen.png}}\hspace{3pt}
    Qwen2.5-VL-72B \cite{bai2025qwen25vl}
        & 0.163 & 0.045 & 0.138 & 0.139 & 0.020 & 0.163 & 0.303 & 0.388 & 0.170 \\
    \raisebox{-0.2\height}{\includegraphics[height=0.9em]{figure/icon/gpt.png}}\hspace{3pt}
    GPT-5 Nano \cite{openai2025gpt5nano}
        & 0.124 & 0.021 & 0.118 & 0.124 & 0.016 & 0.124 & 0.502 & 0.478 & 0.189 \\
\bottomrule
    \end{tabular}
    \caption{(7) Task: Hazard, Cross-Model Evaluation of Multimodal MCP Tool-Calling Trajectories on M$^3$-Bench.}
    \label{tab:mcp_multimodal_results_hazard}
    \vspace{-8pt}
\end{table*}

\begin{table*}[!htb]
    \centering
    \footnotesize
    \setlength{\tabcolsep}{6pt}
    \begin{tabular}{>{\arraybackslash}m{3.7cm}*{9}{>{\centering\arraybackslash}m{1.0cm}}}
    \toprule
    \multicolumn{10}{c}{\textbf{(8) Food Reddit Wiki Excel}} \\
    \midrule
    \multirow{3}{*}{\centering Model}
    & \multicolumn{2}{c}{Detection}
    & \multicolumn{4}{c}{Trajectory Alignment}
    & \multicolumn{2}{c}{LLM Judge}
    & \multirow{3}{*}{\centering\shortstack{Average\\Score}} \\
    \cmidrule(lr){2-3} \cmidrule(lr){4-7} \cmidrule(lr){8-9}
    & Recall & Precision
    & \shortstack{Argument\\Similarity}
    & \shortstack{Step\\Coherence}
    & \shortstack{Order\\Consistency}
    & \shortstack{Merge\\Purity}
    & \shortstack{Task\\Completion}
    & \shortstack{Information\\Grounding}
    & \\
    \midrule
    \raisebox{-0.2\height}{\includegraphics[height=0.9em]{figure/icon/glm.png}}\hspace{3pt}
    GLM 4.5v \cite{hong2025glm45v}
        & 0.000 & 0.000 & 0.000 & 0.000 & 0.000 & 0.000 & 0.285 & 0.130 & 0.052 \\
    \raisebox{-0.2\height}{\includegraphics[height=0.9em]{figure/icon/claude.png}}\hspace{3pt}
    Claude 4.5 Haiku \cite{anthropic2025haiku45}
        & 0.006 & 0.017 & 0.005 & 0.006 & 0.000 & 0.006 & 0.386 & 0.247 & 0.084 \\
    \raisebox{-0.2\height}{\includegraphics[height=0.9em]{figure/icon/qwen.png}}\hspace{3pt}
    Qwen2.5-VL-72B \cite{bai2025qwen25vl}
        & 0.171 & 0.032 & 0.156 & 0.135 & 0.057 & 0.103 & 0.303 & 0.388 & 0.168 \\
    \raisebox{-0.2\height}{\includegraphics[height=0.9em]{figure/icon/gemini.png}}\hspace{3pt}
    Gemini 2.5 Flash Lite \cite{comanici2025gemini25}
        & 0.163 & 0.168 & 0.143 & 0.128 & 0.029 & 0.163 & 0.359 & 0.292 & 0.181 \\
    \raisebox{-0.2\height}{\includegraphics[height=0.9em]{figure/icon/gpt.png}}\hspace{3pt}
    GPT-5 Nano \cite{openai2025gpt5nano}
        & 0.378 & 0.104 & 0.373 & 0.206 & 0.000 & 0.378 & 0.502 & 0.478 & 0.302 \\
\bottomrule
    \end{tabular}
    \caption{(8) Food Reddit Wiki Excel, Cross-Model Evaluation of Multimodal MCP Tool-Calling Trajectories on M$^3$-Bench.}
    \label{tab:mcp_multimodal_results_food_reddit_wiki_excel}
    \vspace{-8pt}
\end{table*}

\begin{table*}[!htb]
    \centering
    \footnotesize
    \setlength{\tabcolsep}{6pt}
    \begin{tabular}{>{\arraybackslash}m{3.7cm}*{9}{>{\centering\arraybackslash}m{1.0cm}}}
    \toprule
    \multicolumn{10}{c}{\textbf{(10) Task: Barcode Lookup}} \\
    \midrule
    \multirow{3}{*}{\centering Model}
    & \multicolumn{2}{c}{Detection}
    & \multicolumn{4}{c}{Trajectory Alignment}
    & \multicolumn{2}{c}{LLM Judge}
    & \multirow{3}{*}{\centering\shortstack{Average\\Score}} \\
    \cmidrule(lr){2-3} \cmidrule(lr){4-7} \cmidrule(lr){8-9}
    & Recall & Precision
    & \shortstack{Argument\\Similarity}
    & \shortstack{Step\\Coherence}
    & \shortstack{Order\\Consistency}
    & \shortstack{Merge\\Purity}
    & \shortstack{Task\\Completion}
    & \shortstack{Information\\Grounding}
    & \\
    \midrule
    \raisebox{-0.2\height}{\includegraphics[height=0.9em]{figure/icon/glm.png}}\hspace{3pt}
    GLM 4.5v \cite{hong2025glm45v}
        & 0.000 & 0.000 & 0.000 & 0.000 & 0.000 & 0.000 & 0.285 & 0.130 & 0.052 \\
    \raisebox{-0.2\height}{\includegraphics[height=0.9em]{figure/icon/gemini.png}}\hspace{3pt}
    Gemini 2.5 Flash Lite \cite{comanici2025gemini25}
        & 0.000 & 0.000 & 0.000 & 0.000 & 0.000 & 0.000 & 0.359 & 0.292 & 0.081 \\
    \raisebox{-0.2\height}{\includegraphics[height=0.9em]{figure/icon/grok.png}}\hspace{3pt}
    Grok-4 Fast \cite{xai2025grok4fast}
        & 0.083 & 0.033 & 0.083 & 0.083 & 0.000 & 0.083 & 0.501 & 0.378 & 0.156 \\
    \raisebox{-0.2\height}{\includegraphics[height=0.9em]{figure/icon/qwen.png}}\hspace{3pt}
    Qwen2.5-VL-72B \cite{bai2025qwen25vl}
        & 0.200 & 0.225 & 0.187 & 0.200 & 0.000 & 0.200 & 0.303 & 0.388 & 0.213 \\
    \raisebox{-0.2\height}{\includegraphics[height=0.9em]{figure/icon/claude.png}}\hspace{3pt}
    Claude 4.5 Haiku \cite{anthropic2025haiku45}
        & 0.250 & 0.417 & 0.230 & 0.250 & 0.000 & 0.250 & 0.386 & 0.247 & 0.254 \\
\bottomrule
    \end{tabular}
    \caption{(10) Task: Barcode Lookup, Cross-Model Evaluation of Multimodal MCP Tool-Calling Trajectories on M$^3$-Bench.}
    \label{tab:mcp_multimodal_results_barcode_lookup}
    \vspace{-8pt}
\end{table*}

\begin{table*}[!htb]
    \centering
    \footnotesize
    \setlength{\tabcolsep}{6pt}
    \begin{tabular}{>{\arraybackslash}m{3.7cm}*{9}{>{\centering\arraybackslash}m{1.0cm}}}
    \toprule
    \multicolumn{10}{c}{\textbf{(11) Biology Cut}} \\
    \midrule
    \multirow{3}{*}{\centering Model}
    & \multicolumn{2}{c}{Detection}
    & \multicolumn{4}{c}{Trajectory Alignment}
    & \multicolumn{2}{c}{LLM Judge}
    & \multirow{3}{*}{\centering\shortstack{Average\\Score}} \\
    \cmidrule(lr){2-3} \cmidrule(lr){4-7} \cmidrule(lr){8-9}
    & Recall & Precision
    & \shortstack{Argument\\Similarity}
    & \shortstack{Step\\Coherence}
    & \shortstack{Order\\Consistency}
    & \shortstack{Merge\\Purity}
    & \shortstack{Task\\Completion}
    & \shortstack{Information\\Grounding}
    & \\
    \midrule
    \raisebox{-0.2\height}{\includegraphics[height=0.9em]{figure/icon/gemini.png}}\hspace{3pt}
    Gemini 2.5 Flash Lite \cite{comanici2025gemini25}
        & 0.000 & 0.000 & 0.000 & 0.000 & 0.000 & 0.000 & 0.359 & 0.292 & 0.081 \\
    \raisebox{-0.2\height}{\includegraphics[height=0.9em]{figure/icon/internvl.png}}\hspace{3pt}
    InternVL 3.5 \cite{wang2025internvl35}
        & 0.024 & 0.056 & 0.024 & 0.024 & 0.000 & 0.024 & 0.325 & 0.298 & 0.097 \\
    \raisebox{-0.2\height}{\includegraphics[height=0.9em]{figure/icon/glm.png}}\hspace{3pt}
    GLM 4.5v \cite{hong2025glm45v}
        & 0.125 & 0.028 & 0.117 & 0.125 & 0.000 & 0.125 & 0.285 & 0.130 & 0.117 \\
    \raisebox{-0.2\height}{\includegraphics[height=0.9em]{figure/icon/gpt.png}}\hspace{3pt}
    GPT-5 Nano \cite{openai2025gpt5nano}
        & 0.000 & 0.000 & 0.000 & 0.000 & 0.000 & 0.000 & 0.502 & 0.478 & 0.123 \\
    \raisebox{-0.2\height}{\includegraphics[height=0.9em]{figure/icon/gpt.png}}\hspace{3pt}
    GPT-5 Mini \cite{openai2025gpt5mini}
        & 0.000 & 0.000 & 0.000 & 0.000 & 0.000 & 0.000 & 0.571 & 0.566 & 0.142 \\
\bottomrule
    \end{tabular}
    \caption{(11) Biology Cut, Cross-Model Evaluation of Multimodal MCP Tool-Calling Trajectories on M$^3$-Bench.}
    \label{tab:mcp_multimodal_results_biology_cut}
    \vspace{-8pt}
\end{table*}

\begin{table*}[!htb]
    \centering
    \footnotesize
    \setlength{\tabcolsep}{6pt}
    \begin{tabular}{>{\arraybackslash}m{3.7cm}*{9}{>{\centering\arraybackslash}m{1.0cm}}}
    \toprule
    \multicolumn{10}{c}{\textbf{(12) Fruit Cal}} \\
    \midrule
    \multirow{3}{*}{\centering Model}
    & \multicolumn{2}{c}{Detection}
    & \multicolumn{4}{c}{Trajectory Alignment}
    & \multicolumn{2}{c}{LLM Judge}
    & \multirow{3}{*}{\centering\shortstack{Average\\Score}} \\
    \cmidrule(lr){2-3} \cmidrule(lr){4-7} \cmidrule(lr){8-9}
    & Recall & Precision
    & \shortstack{Argument\\Similarity}
    & \shortstack{Step\\Coherence}
    & \shortstack{Order\\Consistency}
    & \shortstack{Merge\\Purity}
    & \shortstack{Task\\Completion}
    & \shortstack{Information\\Grounding}
    & \\
    \midrule
    \raisebox{-0.2\height}{\includegraphics[height=0.9em]{figure/icon/glm.png}}\hspace{3pt}
    GLM 4.5v \cite{hong2025glm45v}
        & 0.000 & 0.000 & 0.000 & 0.000 & 0.005 & 0.000 & 0.282 & 0.128 & 0.052 \\
    \raisebox{-0.2\height}{\includegraphics[height=0.9em]{figure/icon/internvl.png}}\hspace{3pt}
    InternVL 3.5 \cite{wang2025internvl35}
        & 0.071 & 0.034 & 0.045 & 0.069 & 0.007 & 0.069 & 0.324 & 0.298 & 0.115 \\
    \raisebox{-0.2\height}{\includegraphics[height=0.9em]{figure/icon/qwen.png}}\hspace{3pt}
    Qwen2.5-VL-72B \cite{bai2025qwen25vl}
        & 0.113 & 0.059 & 0.089 & 0.110 & 0.011 & 0.110 & 0.301 & 0.388 & 0.147 \\
    \raisebox{-0.2\height}{\includegraphics[height=0.9em]{figure/icon/grok.png}}\hspace{3pt}
    Grok 4 (0709) \cite{xai2025grok4}
        & 0.095 & 0.038 & 0.084 & 0.093 & 0.009 & 0.093 & 0.604 & 0.530 & 0.193 \\
    \raisebox{-0.2\height}{\includegraphics[height=0.9em]{figure/icon/claude.png}}\hspace{3pt}
    Claude 4.5 Haiku \cite{anthropic2025haiku45}
        & 0.207 & 0.268 & 0.167 & 0.202 & 0.020 & 0.202 & 0.381 & 0.247 & 0.212 \\
\bottomrule
    \end{tabular}
    \caption{(12) Fruit Cal, Cross-Model Evaluation of Multimodal MCP Tool-Calling Trajectories on M$^3$-Bench.}
    \label{tab:mcp_multimodal_results_fruit_cal}
    \vspace{-8pt}
\end{table*}

\begin{table*}[!htb]
    \centering
    \footnotesize
    \setlength{\tabcolsep}{6pt}
    \begin{tabular}{>{\arraybackslash}m{3.7cm}*{9}{>{\centering\arraybackslash}m{1.0cm}}}
    \toprule
    \multicolumn{10}{c}{\textbf{(13) Medical}} \\
    \midrule
    \multirow{3}{*}{\centering Model}
    & \multicolumn{2}{c}{Detection}
    & \multicolumn{4}{c}{Trajectory Alignment}
    & \multicolumn{2}{c}{LLM Judge}
    & \multirow{3}{*}{\centering\shortstack{Average\\Score}} \\
    \cmidrule(lr){2-3} \cmidrule(lr){4-7} \cmidrule(lr){8-9}
    & Recall & Precision
    & \shortstack{Argument\\Similarity}
    & \shortstack{Step\\Coherence}
    & \shortstack{Order\\Consistency}
    & \shortstack{Merge\\Purity}
    & \shortstack{Task\\Completion}
    & \shortstack{Information\\Grounding}
    & \\
    \midrule
    \raisebox{-0.2\height}{\includegraphics[height=0.9em]{figure/icon/glm.png}}\hspace{3pt}
    GLM 4.5v \cite{hong2025glm45v}
        & 0.000 & 0.000 & 0.000 & 0.000 & 0.000 & 0.000 & 0.285 & 0.130 & 0.052 \\
    \raisebox{-0.2\height}{\includegraphics[height=0.9em]{figure/icon/qwen.png}}\hspace{3pt}
    Qwen2.5-VL-72B \cite{bai2025qwen25vl}
        & 0.036 & 0.031 & 0.035 & 0.036 & 0.000 & 0.036 & 0.303 & 0.388 & 0.108 \\
    \raisebox{-0.2\height}{\includegraphics[height=0.9em]{figure/icon/claude.png}}\hspace{3pt}
    Claude 4.5 Haiku \cite{anthropic2025haiku45}
        & 0.152 & 0.472 & 0.119 & 0.152 & 0.026 & 0.091 & 0.386 & 0.247 & 0.206 \\
    \raisebox{-0.2\height}{\includegraphics[height=0.9em]{figure/icon/internvl.png}}\hspace{3pt}
    InternVL 3.5 \cite{wang2025internvl35}
        & 0.199 & 0.361 & 0.147 & 0.158 & 0.023 & 0.161 & 0.325 & 0.298 & 0.209 \\
    \raisebox{-0.2\height}{\includegraphics[height=0.9em]{figure/icon/gemini.png}}\hspace{3pt}
    Gemini 2.5 Flash Lite \cite{comanici2025gemini25}
        & 0.265 & 0.242 & 0.227 & 0.223 & 0.053 & 0.189 & 0.359 & 0.292 & 0.231 \\
\bottomrule
    \end{tabular}
    \caption{(13) Medical, Cross-Model Evaluation of Multimodal MCP Tool-Calling Trajectories on M$^3$-Bench.}
    \label{tab:mcp_multimodal_results_medical}
    \vspace{-8pt}
\end{table*}

\begin{table*}[!htb]
    \centering
    \footnotesize
    \setlength{\tabcolsep}{6pt}
    \begin{tabular}{>{\arraybackslash}m{3.7cm}*{9}{>{\centering\arraybackslash}m{1.0cm}}}
    \toprule
    \multicolumn{10}{c}{\textbf{(14) QR Attraction}} \\
    \midrule
    \multirow{3}{*}{\centering Model}
    & \multicolumn{2}{c}{Detection}
    & \multicolumn{4}{c}{Trajectory Alignment}
    & \multicolumn{2}{c}{LLM Judge}
    & \multirow{3}{*}{\centering\shortstack{Average\\Score}} \\
    \cmidrule(lr){2-3} \cmidrule(lr){4-7} \cmidrule(lr){8-9}
    & Recall & Precision
    & \shortstack{Argument\\Similarity}
    & \shortstack{Step\\Coherence}
    & \shortstack{Order\\Consistency}
    & \shortstack{Merge\\Purity}
    & \shortstack{Task\\Completion}
    & \shortstack{Information\\Grounding}
    & \\
    \midrule
    \raisebox{-0.2\height}{\includegraphics[height=0.9em]{figure/icon/glm.png}}\hspace{3pt}
    GLM 4.5v \cite{hong2025glm45v}
        & 0.019 & 0.034 & 0.016 & 0.020 & 0.000 & 0.023 & 0.270 & 0.122 & 0.063 \\
    \raisebox{-0.2\height}{\includegraphics[height=0.9em]{figure/icon/gpt.png}}\hspace{3pt}
    GPT-5 Nano \cite{openai2025gpt5nano}
        & 0.136 & 0.050 & 0.136 & 0.136 & 0.000 & 0.136 & 0.502 & 0.478 & 0.197 \\
    \raisebox{-0.2\height}{\includegraphics[height=0.9em]{figure/icon/internvl.png}}\hspace{3pt}
    InternVL 3.5 \cite{wang2025internvl35}
        & 0.214 & 0.236 & 0.168 & 0.214 & 0.000 & 0.214 & 0.319 & 0.292 & 0.207 \\
    \raisebox{-0.2\height}{\includegraphics[height=0.9em]{figure/icon/qwen.png}}\hspace{3pt}
    Qwen2.5-VL-72B \cite{bai2025qwen25vl}
        & 0.280 & 0.302 & 0.220 & 0.280 & 0.022 & 0.280 & 0.303 & 0.388 & 0.259 \\
    \raisebox{-0.2\height}{\includegraphics[height=0.9em]{figure/icon/claude.png}}\hspace{3pt}
    Claude 4.5 Haiku \cite{anthropic2025haiku45}
        & 0.285 & 0.717 & 0.237 & 0.322 & 0.118 & 0.190 & 0.390 & 0.231 & 0.311 \\
\bottomrule
    \end{tabular}
    \caption{(14) QR Attraction, Cross-Model Evaluation of Multimodal MCP Tool-Calling Trajectories on M$^3$-Bench.}
    \label{tab:mcp_multimodal_results_qr_attraction}
    \vspace{-8pt}
\end{table*}


\begin{table*}[!htb]
    \centering
    \footnotesize
    \setlength{\tabcolsep}{6pt}
    \begin{tabular}{>{\arraybackslash}m{3.7cm}*{9}{>{\centering\arraybackslash}m{1.0cm}}}
    \toprule
    \multicolumn{10}{c}{\textbf{(9) Task: MedExcel}} \\
    \midrule
    \multirow{3}{*}{\centering Model}
    & \multicolumn{2}{c}{Detection}
    & \multicolumn{4}{c}{Trajectory Alignment}
    & \multicolumn{2}{c}{LLM Judge}
    & \multirow{3}{*}{\centering\shortstack{Average\\Score}} \\
    \cmidrule(lr){2-3} \cmidrule(lr){4-7} \cmidrule(lr){8-9}
    & Recall & Precision
    & \shortstack{Argument\\Similarity}
    & \shortstack{Step\\Coherence}
    & \shortstack{Order\\Consistency}
    & \shortstack{Merge\\Purity}
    & \shortstack{Task\\Completion}
    & \shortstack{Information\\Grounding}
    & \\
    \midrule
    \raisebox{-0.2\height}{\includegraphics[height=0.9em]{figure/icon/glm.png}}\hspace{3pt}
    GLM 4.5v \cite{hong2025glm45v}
        & 0.004 & 0.009 & 0.003 & 0.005 & 0.000 & 0.005 & 0.266 & 0.121 & 0.052 \\
    \raisebox{-0.2\height}{\includegraphics[height=0.9em]{figure/icon/gemini.png}}\hspace{3pt}
    Gemini 2.5 Flash Lite \cite{comanici2025gemini25}
        & 0.036 & 0.013 & 0.027 & 0.036 & 0.000 & 0.036 & 0.316 & 0.259 & 0.090 \\
    \raisebox{-0.2\height}{\includegraphics[height=0.9em]{figure/icon/internvl.png}}\hspace{3pt}
    InternVL 3.5 \cite{wang2025internvl35}
        & 0.071 & 0.029 & 0.049 & 0.071 & 0.000 & 0.071 & 0.319 & 0.292 & 0.113 \\
    \raisebox{-0.2\height}{\includegraphics[height=0.9em]{figure/icon/gpt.png}}\hspace{3pt}
    GPT-5 Nano \cite{openai2025gpt5nano}
        & 0.050 & 0.021 & 0.043 & 0.050 & 0.000 & 0.020 & 0.402 & 0.378 & 0.121 \\
    \raisebox{-0.2\height}{\includegraphics[height=0.9em]{figure/icon/claude.png}}\hspace{3pt}
    Claude 4.5 Haiku \cite{anthropic2025haiku45}
        & 0.067 & 0.189 & 0.052 & 0.079 & 0.006 & 0.044 & 0.385 & 0.229 & 0.131 \\
\bottomrule
    \end{tabular}
    \caption{(9) Task: MedExcel, Cross-Model Evaluation of Multimodal MCP Tool-Calling Trajectories on M$^3$-Bench.}
    \label{tab:mcp_multimodal_results_medexcel}
    \vspace{-8pt}
\end{table*}


\begin{table*}[!htb]
    \centering
    \footnotesize
    \setlength{\tabcolsep}{6pt}
    \begin{tabular}{>{\arraybackslash}m{3.7cm}*{9}{>{\centering\arraybackslash}m{1.0cm}}}
    \toprule
    \multicolumn{10}{c}{\textbf{(15) Task: Routine}} \\
    \midrule
    \multirow{3}{*}{\centering Model}
    & \multicolumn{2}{c}{Detection}
    & \multicolumn{4}{c}{Trajectory Alignment}
    & \multicolumn{2}{c}{LLM Judge}
    & \multirow{3}{*}{\centering\shortstack{Average\\Score}} \\
    \cmidrule(lr){2-3} \cmidrule(lr){4-7} \cmidrule(lr){8-9}
    & Recall & Precision
    & \shortstack{Argument\\Similarity}
    & \shortstack{Step\\Coherence}
    & \shortstack{Order\\Consistency}
    & \shortstack{Merge\\Purity}
    & \shortstack{Task\\Completion}
    & \shortstack{Information\\Grounding}
    & \\
    \midrule
    \raisebox{-0.2\height}{\includegraphics[height=0.9em]{figure/icon/glm.png}}\hspace{3pt}
    GLM 4.5v \cite{hong2025glm45v}
        & 0.000 & 0.000 & 0.000 & 0.000 & 0.000 & 0.000 & 0.285 & 0.130 & 0.052 \\
    \raisebox{-0.2\height}{\includegraphics[height=0.9em]{figure/icon/gemini.png}}\hspace{3pt}
    Gemini 2.5 Flash Lite \cite{comanici2025gemini25}
        & 0.081 & 0.327 & 0.066 & 0.081 & 0.000 & 0.081 & 0.316 & 0.259 & 0.152 \\
    \raisebox{-0.2\height}{\includegraphics[height=0.6em]{figure/icon/llama.png}}\hspace{3pt}
    Llama-4-Scout-17B16E \cite{meta2025llama4scout17b16e}
        & 0.115 & 0.238 & 0.098 & 0.115 & 0.000 & 0.115 & 0.385 & 0.354 & 0.177 \\
    \raisebox{-0.2\height}{\includegraphics[height=0.9em]{figure/icon/qwen.png}}\hspace{3pt}
    Qwen2.5-VL-72B \cite{bai2025qwen25vl}
        & 0.162 & 0.162 & 0.132 & 0.162 & 0.031 & 0.106 & 0.303 & 0.388 & 0.181 \\
    \raisebox{-0.2\height}{\includegraphics[height=0.9em]{figure/icon/internvl.png}}\hspace{3pt}
    InternVL 3.5 \cite{wang2025internvl35}
        & 0.239 & 0.417 & 0.195 & 0.194 & 0.029 & 0.191 & 0.319 & 0.292 & 0.234 \\
\bottomrule
    \end{tabular}
    \caption{(15) Task: Routine, Cross-Model Evaluation of Multimodal MCP Tool-Calling Trajectories on M$^3$-Bench.}
    \label{tab:mcp_multimodal_results_routine}
    \vspace{-8pt}
\end{table*}


\begin{table*}[!htb]
    \centering
    \footnotesize
    \setlength{\tabcolsep}{6pt}
    \begin{tabular}{>{\arraybackslash}m{3.7cm}*{9}{>{\centering\arraybackslash}m{1.0cm}}}
    \toprule
    \multicolumn{10}{c}{\textbf{(16) Task: Biology Chain}} \\
    \midrule
    \multirow{3}{*}{\centering Model}
    & \multicolumn{2}{c}{Detection}
    & \multicolumn{4}{c}{Trajectory Alignment}
    & \multicolumn{2}{c}{LLM Judge}
    & \multirow{3}{*}{\centering\shortstack{Average\\Score}} \\
    \cmidrule(lr){2-3} \cmidrule(lr){4-7} \cmidrule(lr){8-9}
    & Recall & Precision
    & \shortstack{Argument\\Similarity}
    & \shortstack{Step\\Coherence}
    & \shortstack{Order\\Consistency}
    & \shortstack{Merge\\Purity}
    & \shortstack{Task\\Completion}
    & \shortstack{Information\\Grounding}
    & \\
    \midrule
    \raisebox{-0.2\height}{\includegraphics[height=0.9em]{figure/icon/glm.png}}\hspace{3pt}
    GLM 4.5v \cite{hong2025glm45v}
        & 0.011 & 0.050 & 0.011 & 0.011 & 0.000 & 0.011 & 0.285 & 0.130 & 0.064 \\
    \raisebox{-0.2\height}{\includegraphics[height=0.9em]{figure/icon/qwen.png}}\hspace{3pt}
    Qwen2.5-VL-72B \cite{bai2025qwen25vl}
        & 0.117 & 0.051 & 0.091 & 0.117 & 0.015 & 0.051 & 0.303 & 0.388 & 0.142 \\
    \raisebox{-0.2\height}{\includegraphics[height=0.9em]{figure/icon/gpt.png}}\hspace{3pt}
    GPT-5 Nano \cite{openai2025gpt5nano}
        & 0.049 & 0.063 & 0.048 & 0.049 & 0.000 & 0.049 & 0.502 & 0.478 & 0.155 \\
    \raisebox{-0.2\height}{\includegraphics[height=0.9em]{figure/icon/gemini.png}}\hspace{3pt}
    Gemini 2.5 Flash Lite \cite{comanici2025gemini25}
        & 0.150 & 0.202 & 0.130 & 0.150 & 0.035 & 0.026 & 0.316 & 0.259 & 0.158 \\
    \raisebox{-0.2\height}{\includegraphics[height=0.9em]{figure/icon/internvl.png}}\hspace{3pt}
    InternVL 3.5 \cite{wang2025internvl35}
        & 0.090 & 0.375 & 0.068 & 0.090 & 0.009 & 0.058 & 0.319 & 0.292 & 0.163 \\
\bottomrule
    \end{tabular}
    \caption{(16) Task: Biology Chain, Cross-Model Evaluation of Multimodal MCP Tool-Calling Trajectories on M$^3$-Bench.}
    \label{tab:mcp_multimodal_results_biology_chain}
    \vspace{-8pt}
\end{table*}


\begin{table*}[!htb]
    \centering
    \footnotesize
    \setlength{\tabcolsep}{6pt}
    \begin{tabular}{>{\arraybackslash}m{3.7cm}*{9}{>{\centering\arraybackslash}m{1.0cm}}}
    \toprule
    \multicolumn{10}{c}{\textbf{(17) Task: Generate PowerPoint}} \\
    \midrule
    \multirow{3}{*}{\centering Model}
    & \multicolumn{2}{c}{Detection}
    & \multicolumn{4}{c}{Trajectory Alignment}
    & \multicolumn{2}{c}{LLM Judge}
    & \multirow{3}{*}{\centering\shortstack{Average\\Score}} \\
    \cmidrule(lr){2-3} \cmidrule(lr){4-7} \cmidrule(lr){8-9}
    & Recall & Precision
    & \shortstack{Argument\\Similarity}
    & \shortstack{Step\\Coherence}
    & \shortstack{Order\\Consistency}
    & \shortstack{Merge\\Purity}
    & \shortstack{Task\\Completion}
    & \shortstack{Information\\Grounding}
    & \\
    \midrule
    \raisebox{-0.2\height}{\includegraphics[height=0.9em]{figure/icon/glm.png}}\hspace{3pt}
    GLM 4.5v \cite{hong2025glm45v}
        & 0.024 & 0.026 & 0.020 & 0.024 & 0.000 & 0.023 & 0.270 & 0.122 & 0.064 \\
    \raisebox{-0.2\height}{\includegraphics[height=0.9em]{figure/icon/qwen.png}}\hspace{3pt}
    Qwen2.5-VL-72B \cite{bai2025qwen25vl}
        & 0.200 & 0.057 & 0.164 & 0.200 & 0.083 & 0.103 & 0.303 & 0.388 & 0.187 \\
    \raisebox{-0.2\height}{\includegraphics[height=0.9em]{figure/icon/gemini.png}}\hspace{3pt}
    Gemini 2.5 Flash Lite \cite{comanici2025gemini25}
        & 0.210 & 0.083 & 0.176 & 0.210 & 0.048 & 0.210 & 0.316 & 0.259 & 0.189 \\
    \raisebox{-0.2\height}{\includegraphics[height=0.9em]{figure/icon/gpt.png}}\hspace{3pt}
    GPT-5 Nano \cite{openai2025gpt5nano}
        & 0.136 & 0.054 & 0.134 & 0.136 & 0.009 & 0.086 & 0.502 & 0.478 & 0.192 \\
    \raisebox{-0.2\height}{\includegraphics[height=0.9em]{figure/icon/gpt.png}}\hspace{3pt}
    GPT-5 Mini \cite{openai2025gpt5mini}
        & 0.200 & 0.094 & 0.196 & 0.200 & 0.026 & 0.159 & 0.571 & 0.566 & 0.252 \\
\bottomrule
    \end{tabular}
    \caption{(17) Task: Generate PowerPoint, Cross-Model Evaluation of Multimodal MCP Tool-Calling Trajectories on M$^3$-Bench.}
    \label{tab:mcp_multimodal_results_generate_powerpoint}
    \vspace{-8pt}
\end{table*}


\begin{table*}[!htb]
    \centering
    \footnotesize
    \setlength{\tabcolsep}{6pt}
    \begin{tabular}{>{\arraybackslash}m{3.7cm}*{9}{>{\centering\arraybackslash}m{1.0cm}}}
    \toprule
    \multicolumn{10}{c}{\textbf{(18) Task: Travel Comparison}} \\
    \midrule
    \multirow{3}{*}{\centering Model}
    & \multicolumn{2}{c}{Detection}
    & \multicolumn{4}{c}{Trajectory Alignment}
    & \multicolumn{2}{c}{LLM Judge}
    & \multirow{3}{*}{\centering\shortstack{Average\\Score}} \\
    \cmidrule(lr){2-3} \cmidrule(lr){4-7} \cmidrule(lr){8-9}
    & Recall & Precision
    & \shortstack{Argument\\Similarity}
    & \shortstack{Step\\Coherence}
    & \shortstack{Order\\Consistency}
    & \shortstack{Merge\\Purity}
    & \shortstack{Task\\Completion}
    & \shortstack{Information\\Grounding}
    & \\
    \midrule
    \raisebox{-0.2\height}{\includegraphics[height=0.9em]{figure/icon/glm.png}}\hspace{3pt}
    GLM 4.5v \cite{hong2025glm45v}
        & 0.023 & 0.017 & 0.020 & 0.020 & 0.000 & 0.024 & 0.270 & 0.122 & 0.062 \\
    \raisebox{-0.2\height}{\includegraphics[height=0.9em]{figure/icon/qwen.png}}\hspace{3pt}
    Qwen2.5-VL-72B \cite{bai2025qwen25vl}
        & 0.111 & 0.026 & 0.108 & 0.111 & 0.000 & 0.111 & 0.303 & 0.388 & 0.145 \\
    \raisebox{-0.2\height}{\includegraphics[height=0.9em]{figure/icon/claude.png}}\hspace{3pt}
    Claude 4.5 Haiku \cite{anthropic2025haiku45}
        & 0.348 & 0.359 & 0.297 & 0.317 & 0.199 & 0.200 & 0.390 & 0.231 & 0.293 \\
    \raisebox{-0.2\height}{\includegraphics[height=0.9em]{figure/icon/internvl.png}}\hspace{3pt}
    InternVL 3.5 \cite{wang2025internvl35}
        & 0.476 & 0.224 & 0.400 & 0.381 & 0.429 & 0.476 & 0.319 & 0.292 & 0.375 \\
    \raisebox{-0.2\height}{\includegraphics[height=0.9em]{figure/icon/gemini.png}}\hspace{3pt}
    Gemini 2.5 Pro \cite{comanici2025gemini25}
        & 0.571 & 0.088 & 0.509 & 0.429 & 0.571 & 0.356 & 0.487 & 0.427 & 0.430 \\
\bottomrule
    \end{tabular}
    \caption{(18) Task: Travel Comparison, Cross-Model Evaluation of Multimodal MCP Tool-Calling Trajectories on M$^3$-Bench.}
    \label{tab:mcp_multimodal_results_travel_comparison}
    \vspace{-8pt}
\end{table*}


\begin{table*}[!htb]
    \centering
    \footnotesize
    \setlength{\tabcolsep}{6pt}
    \begin{tabular}{>{\arraybackslash}m{3.7cm}*{9}{>{\centering\arraybackslash}m{1.0cm}}}
    \toprule
    \multicolumn{10}{c}{\textbf{(19) Task: Heatmap}} \\
    \midrule
    \multirow{3}{*}{\centering Model}
    & \multicolumn{2}{c}{Detection}
    & \multicolumn{4}{c}{Trajectory Alignment}
    & \multicolumn{2}{c}{LLM Judge}
    & \multirow{3}{*}{\centering\shortstack{Average\\Score}} \\
    \cmidrule(lr){2-3} \cmidrule(lr){4-7} \cmidrule(lr){8-9}
    & Recall & Precision
    & \shortstack{Argument\\Similarity}
    & \shortstack{Step\\Coherence}
    & \shortstack{Order\\Consistency}
    & \shortstack{Merge\\Purity}
    & \shortstack{Task\\Completion}
    & \shortstack{Information\\Grounding}
    & \\
    \midrule
    \raisebox{-0.2\height}{\includegraphics[height=0.9em]{figure/icon/glm.png}}\hspace{3pt}
    GLM 4.5v \cite{hong2025glm45v}
        & 0.016 & 0.028 & 0.014 & 0.017 & 0.000 & 0.017 & 0.269 & 0.122 & 0.060 \\
    \raisebox{-0.2\height}{\includegraphics[height=0.9em]{figure/icon/claude.png}}\hspace{3pt}
    Claude 4.5 Haiku \cite{anthropic2025haiku45}
        & 0.040 & 0.200 & 0.037 & 0.040 & 0.000 & 0.040 & 0.386 & 0.247 & 0.124 \\
    \raisebox{-0.2\height}{\includegraphics[height=0.9em]{figure/icon/qwen.png}}\hspace{3pt}
    Qwen2.5-VL-72B \cite{bai2025qwen25vl}
        & 0.172 & 0.090 & 0.158 & 0.177 & 0.038 & 0.146 & 0.286 & 0.364 & 0.179 \\
    \raisebox{-0.2\height}{\includegraphics[height=0.9em]{figure/icon/grok.png}}\hspace{3pt}
    Grok-4 Fast \cite{xai2025grok4fast}
        & 0.100 & 0.219 & 0.088 & 0.100 & 0.038 & 0.072 & 0.501 & 0.378 & 0.187 \\
    \raisebox{-0.2\height}{\includegraphics[height=0.9em]{figure/icon/internvl.png}}\hspace{3pt}
    InternVL 3.5 \cite{wang2025internvl35}
        & 0.225 & 0.280 & 0.214 & 0.225 & 0.038 & 0.194 & 0.319 & 0.292 & 0.223 \\
\bottomrule
    \end{tabular}
    \caption{(19) Task: Heatmap, Cross-Model Evaluation of Multimodal MCP Tool-Calling Trajectories on M$^3$-Bench.}
    \label{tab:mcp_multimodal_results_heatmap}
    \vspace{-8pt}
\end{table*}


\begin{table*}[!htb]
    \centering
    \footnotesize
    \setlength{\tabcolsep}{6pt}
    \begin{tabular}{>{\arraybackslash}m{3.7cm}*{9}{>{\centering\arraybackslash}m{1.0cm}}}
    \toprule
    \multicolumn{10}{c}{\textbf{(20) Task: Museum}} \\
    \midrule
    \multirow{3}{*}{\centering Model}
    & \multicolumn{2}{c}{Detection}
    & \multicolumn{4}{c}{Trajectory Alignment}
    & \multicolumn{2}{c}{LLM Judge}
    & \multirow{3}{*}{\centering\shortstack{Average\\Score}} \\
    \cmidrule(lr){2-3} \cmidrule(lr){4-7} \cmidrule(lr){8-9}
    & Recall & Precision
    & \shortstack{Argument\\Similarity}
    & \shortstack{Step\\Coherence}
    & \shortstack{Order\\Consistency}
    & \shortstack{Merge\\Purity}
    & \shortstack{Task\\Completion}
    & \shortstack{Information\\Grounding}
    & \\
    \midrule
    \raisebox{-0.2\height}{\includegraphics[height=0.9em]{figure/icon/glm.png}}\hspace{3pt}
    GLM 4.5v \cite{hong2025glm45v}
        & 0.014 & 0.034 & 0.012 & 0.010 & 0.000 & 0.014 & 0.262 & 0.118 & 0.058 \\
    \raisebox{-0.2\height}{\includegraphics[height=0.9em]{figure/icon/qwen.png}}\hspace{3pt}
    Qwen2.5-VL-72B \cite{bai2025qwen25vl}
        & 0.150 & 0.109 & 0.138 & 0.101 & 0.012 & 0.124 & 0.279 & 0.353 & 0.158 \\
    \raisebox{-0.2\height}{\includegraphics[height=0.9em]{figure/icon/internvl.png}}\hspace{3pt}
    InternVL 3.5 \cite{wang2025internvl35}
        & 0.200 & 0.217 & 0.198 & 0.163 & 0.000 & 0.200 & 0.319 & 0.292 & 0.199 \\
    \raisebox{-0.2\height}{\includegraphics[height=0.9em]{figure/icon/claude.png}}\hspace{3pt}
    Claude 4.5 Haiku \cite{anthropic2025haiku45}
        & 0.203 & 0.714 & 0.183 & 0.153 & 0.028 & 0.120 & 0.379 & 0.224 & 0.250 \\
    \raisebox{-0.2\height}{\includegraphics[height=0.6em]{figure/icon/llama.png}}\hspace{3pt}
    Llama-4-Scout-17B16E \cite{meta2025llama4scout17b16e}
        & 0.385 & 0.262 & 0.359 & 0.231 & 0.036 & 0.321 & 0.337 & 0.305 & 0.279 \\
\bottomrule
    \end{tabular}
    \caption{(20) Task: Museum, Cross-Model Evaluation of Multimodal MCP Tool-Calling Trajectories on M$^3$-Bench.}
    \label{tab:mcp_multimodal_results_museum}
    \vspace{-8pt}
\end{table*}


\begin{table*}[!htb]
    \centering
    \footnotesize
    \setlength{\tabcolsep}{6pt}
    \begin{tabular}{>{\arraybackslash}m{3.7cm}*{9}{>{\centering\arraybackslash}m{1.0cm}}}
    \toprule
    \multicolumn{10}{c}{\textbf{(21) Task: Research Station}} \\
    \midrule
    \multirow{3}{*}{\centering Model}
    & \multicolumn{2}{c}{Detection}
    & \multicolumn{4}{c}{Trajectory Alignment}
    & \multicolumn{2}{c}{LLM Judge}
    & \multirow{3}{*}{\centering\shortstack{Average\\Score}} \\
    \cmidrule(lr){2-3} \cmidrule(lr){4-7} \cmidrule(lr){8-9}
    & Recall & Precision
    & \shortstack{Argument\\Similarity}
    & \shortstack{Step\\Coherence}
    & \shortstack{Order\\Consistency}
    & \shortstack{Merge\\Purity}
    & \shortstack{Task\\Completion}
    & \shortstack{Information\\Grounding}
    & \\
    \midrule
    \raisebox{-0.2\height}{\includegraphics[height=0.9em]{figure/icon/glm.png}}\hspace{3pt}
    GLM 4.5v \cite{hong2025glm45v}
        & 0.014 & 0.034 & 0.012 & 0.010 & 0.000 & 0.014 & 0.262 & 0.118 & 0.058 \\
    \raisebox{-0.2\height}{\includegraphics[height=0.9em]{figure/icon/qwen.png}}\hspace{3pt}
    Qwen2.5-VL-72B \cite{bai2025qwen25vl}
        & 0.150 & 0.109 & 0.138 & 0.101 & 0.012 & 0.124 & 0.279 & 0.353 & 0.158 \\
    \raisebox{-0.2\height}{\includegraphics[height=0.9em]{figure/icon/internvl.png}}\hspace{3pt}
    InternVL 3.5 \cite{wang2025internvl35}
        & 0.200 & 0.217 & 0.198 & 0.163 & 0.000 & 0.200 & 0.319 & 0.292 & 0.199 \\
    \raisebox{-0.2\height}{\includegraphics[height=0.9em]{figure/icon/claude.png}}\hspace{3pt}
    Claude 4.5 Haiku \cite{anthropic2025haiku45}
        & 0.203 & 0.714 & 0.183 & 0.153 & 0.028 & 0.120 & 0.379 & 0.224 & 0.250 \\
    \raisebox{-0.2\height}{\includegraphics[height=0.6em]{figure/icon/llama.png}}\hspace{3pt}
    Llama-4-Scout-17B16E \cite{meta2025llama4scout17b16e}
        & 0.385 & 0.262 & 0.359 & 0.231 & 0.036 & 0.321 & 0.337 & 0.305 & 0.279 \\
\bottomrule
    \end{tabular}
    \caption{(21) Task: Research Station, Cross-Model Evaluation of Multimodal MCP Tool-Calling Trajectories on M$^3$-Bench.}
    \label{tab:mcp_multimodal_results_research_station}
    \vspace{-8pt}
\end{table*}


\begin{table*}[!htb]
    \centering
    \footnotesize
    \setlength{\tabcolsep}{6pt}
    \begin{tabular}{>{\arraybackslash}m{3.7cm}*{9}{>{\centering\arraybackslash}m{1.0cm}}}
    \toprule
    \multicolumn{10}{c}{\textbf{(22) Task: Hugeicons}} \\
    \midrule
    \multirow{3}{*}{\centering Model}
    & \multicolumn{2}{c}{Detection}
    & \multicolumn{4}{c}{Trajectory Alignment}
    & \multicolumn{2}{c}{LLM Judge}
    & \multirow{3}{*}{\centering\shortstack{Average\\Score}} \\
    \cmidrule(lr){2-3} \cmidrule(lr){4-7} \cmidrule(lr){8-9}
    & Recall & Precision
    & \shortstack{Argument\\Similarity}
    & \shortstack{Step\\Coherence}
    & \shortstack{Order\\Consistency}
    & \shortstack{Merge\\Purity}
    & \shortstack{Task\\Completion}
    & \shortstack{Information\\Grounding}
    & \\
    \midrule
    \raisebox{-0.2\height}{\includegraphics[height=0.9em]{figure/icon/glm.png}}\hspace{3pt}
    GLM 4.5v \cite{hong2025glm45v}
        & 0.014 & 0.036 & 0.010 & 0.013 & 0.000 & 0.016 & 0.262 & 0.118 & 0.059 \\
    \raisebox{-0.2\height}{\includegraphics[height=0.9em]{figure/icon/gemini.png}}\hspace{3pt}
    Gemini 2.5 Flash Lite \cite{comanici2025gemini25}
        & 0.113 & 0.256 & 0.096 & 0.113 & 0.000 & 0.113 & 0.316 & 0.259 & 0.158 \\
    \raisebox{-0.2\height}{\includegraphics[height=0.9em]{figure/icon/qwen.png}}\hspace{3pt}
    Qwen2.5-VL-72B \cite{bai2025qwen25vl}
        & 0.149 & 0.115 & 0.114 & 0.137 & 0.040 & 0.135 & 0.279 & 0.353 & 0.165 \\
    \raisebox{-0.2\height}{\includegraphics[height=0.9em]{figure/icon/internvl.png}}\hspace{3pt}
    InternVL 3.5 \cite{wang2025internvl35}
        & 0.208 & 0.190 & 0.164 & 0.208 & 0.000 & 0.208 & 0.319 & 0.292 & 0.199 \\
    \raisebox{-0.2\height}{\includegraphics[height=0.9em]{figure/icon/gpt.png}}\hspace{3pt}
    GPT-5 Nano \cite{openai2025gpt5nano}
        & 0.143 & 0.140 & 0.137 & 0.143 & 0.000 & 0.143 & 0.502 & 0.478 & 0.211 \\
\bottomrule
    \end{tabular}
    \caption{(22) Task: Hugeicons, Cross-Model Evaluation of Multimodal MCP Tool-Calling Trajectories on M$^3$-Bench.}
    \label{tab:mcp_multimodal_results_hugeicons}
    \vspace{-8pt}
\end{table*}


\begin{table*}[!htb]
    \centering
    \footnotesize
    \setlength{\tabcolsep}{6pt}
    \begin{tabular}{>{\arraybackslash}m{3.7cm}*{9}{>{\centering\arraybackslash}m{1.0cm}}}
    \toprule
    \multicolumn{10}{c}{\textbf{(23) Task: Crypto Purchase Estimation}} \\
    \midrule
    \multirow{3}{*}{\centering Model}
    & \multicolumn{2}{c}{Detection}
    & \multicolumn{4}{c}{Trajectory Alignment}
    & \multicolumn{2}{c}{LLM Judge}
    & \multirow{3}{*}{\centering\shortstack{Average\\Score}} \\
    \cmidrule(lr){2-3} \cmidrule(lr){4-7} \cmidrule(lr){8-9}
    & Recall & Precision
    & \shortstack{Argument\\Similarity}
    & \shortstack{Step\\Coherence}
    & \shortstack{Order\\Consistency}
    & \shortstack{Merge\\Purity}
    & \shortstack{Task\\Completion}
    & \shortstack{Information\\Grounding}
    & \\
    \midrule
    \raisebox{-0.2\height}{\includegraphics[height=0.9em]{figure/icon/glm.png}}\hspace{3pt}
    GLM 4.5v \cite{hong2025glm45v}
        & 0.021 & 0.051 & 0.018 & 0.019 & 0.000 & 0.016 & 0.262 & 0.118 & 0.063 \\
    \raisebox{-0.2\height}{\includegraphics[height=0.9em]{figure/icon/qwen.png}}\hspace{3pt}
    Qwen2.5-VL-72B \cite{bai2025qwen25vl}
        & 0.232 & 0.163 & 0.209 & 0.203 & 0.046 & 0.137 & 0.279 & 0.353 & 0.203 \\
    \raisebox{-0.2\height}{\includegraphics[height=0.9em]{figure/icon/gemini.png}}\hspace{3pt}
    Gemini 2.5 Flash Lite \cite{comanici2025gemini25}
        & 0.360 & 0.415 & 0.348 & 0.360 & 0.000 & 0.360 & 0.316 & 0.259 & 0.302 \\
    \raisebox{-0.2\height}{\includegraphics[height=0.9em]{figure/icon/claude.png}}\hspace{3pt}
    Claude 4.5 Haiku \cite{anthropic2025haiku45}
        & 0.314 & 1.000 & 0.278 & 0.309 & 0.110 & 0.132 & 0.379 & 0.224 & 0.352 \\
    \raisebox{-0.2\height}{\includegraphics[height=0.6em]{figure/icon/llama.png}}\hspace{3pt}
    Llama-4-Scout-17B16E \cite{meta2025llama4scout17b16e}
        & 0.596 & 0.394 & 0.544 & 0.465 & 0.142 & 0.355 & 0.337 & 0.305 & 0.392 \\
\bottomrule
    \end{tabular}
    \caption{(23) Task: Crypto Purchase Estimation, Cross-Model Evaluation of Multimodal MCP Tool-Calling Trajectories on M$^3$-Bench.}
    \label{tab:mcp_multimodal_results_crypto_purchase_estimation}
    \vspace{-8pt}
\end{table*}

\begin{table*}[!htb]
    \centering
    \footnotesize
    \setlength{\tabcolsep}{6pt}
    \begin{tabular}{>{\arraybackslash}m{3.7cm}*{9}{>{\centering\arraybackslash}m{1.0cm}}}
    \toprule
    \multicolumn{10}{c}{\textbf{(24) Task: Stock Growth Estimation}} \\
    \midrule
    \multirow{3}{*}{\centering Model}
    & \multicolumn{2}{c}{Detection}
    & \multicolumn{4}{c}{Trajectory Alignment}
    & \multicolumn{2}{c}{LLM Judge}
    & \multirow{3}{*}{\centering\shortstack{Average\\Score}} \\
    \cmidrule(lr){2-3} \cmidrule(lr){4-7} \cmidrule(lr){8-9}
    & Recall & Precision
    & \shortstack{Argument\\Similarity}
    & \shortstack{Step\\Coherence}
    & \shortstack{Order\\Consistency}
    & \shortstack{Merge\\Purity}
    & \shortstack{Task\\Completion}
    & \shortstack{Information\\Grounding}
    & \\
    \midrule
    \raisebox{-0.2\height}{\includegraphics[height=0.9em]{figure/icon/glm.png}}\hspace{3pt}
    GLM 4.5v \cite{hong2025glm45v}
        & 0.050 & 0.041 & 0.048 & 0.050 & 0.000 & 0.050 & 0.128 & 0.022 & 0.049 \\
    \raisebox{-0.2\height}{\includegraphics[height=0.9em]{figure/icon/gemini.png}}\hspace{3pt}
    Gemini 2.5 Flash Lite \cite{comanici2025gemini25}
        & 0.388 & 0.275 & 0.384 & 0.388 & 0.071 & 0.263 & 0.184 & 0.165 & 0.265 \\
    \raisebox{-0.2\height}{\includegraphics[height=0.6em]{figure/icon/llama.png}}\hspace{3pt}
    Llama-4-Scout-17B16E \cite{meta2025llama4scout17b16e}
        & 0.575 & 0.156 & 0.562 & 0.525 & 0.198 & 0.187 & 0.287 & 0.412 & 0.363 \\
    \raisebox{-0.2\height}{\includegraphics[height=0.9em]{figure/icon/qwen.png}}\hspace{3pt}
    Qwen2.5-VL-72B \cite{bai2025qwen25vl}
        & 0.531 & 0.444 & 0.514 & 0.531 & 0.306 & 0.206 & 0.234 & 0.312 & 0.385 \\
    \raisebox{-0.2\height}{\includegraphics[height=0.9em]{figure/icon/claude.png}}\hspace{3pt}
    Claude 4.5 Haiku \cite{anthropic2025haiku45}
        & 0.563 & 0.361 & 0.552 & 0.563 & 0.243 & 0.365 & 0.355 & 0.331 & 0.417 \\
\bottomrule
    \end{tabular}
    \caption{(24) Task: Stock Growth Estimation, Cross-Model Evaluation of Multimodal MCP Tool-Calling Trajectories on M$^3$-Bench.}
    \vspace{-8pt}
\end{table*}

\begin{table*}[!htb]
    \centering
    \footnotesize
    \setlength{\tabcolsep}{6pt}
    \begin{tabular}{>{\arraybackslash}m{3.7cm}*{9}{>{\centering\arraybackslash}m{1.0cm}}}
    \toprule
    \multicolumn{10}{c}{\textbf{(25) Task: FIPE Car Variant Lookup}} \\
    \midrule
    \multirow{3}{*}{\centering Model}
    & \multicolumn{2}{c}{Detection}
    & \multicolumn{4}{c}{Trajectory Alignment}
    & \multicolumn{2}{c}{LLM Judge}
    & \multirow{3}{*}{\centering\shortstack{Average\\Score}} \\
    \cmidrule(lr){2-3} \cmidrule(lr){4-7} \cmidrule(lr){8-9}
    & Recall & Precision
    & \shortstack{Argument\\Similarity}
    & \shortstack{Step\\Coherence}
    & \shortstack{Order\\Consistency}
    & \shortstack{Merge\\Purity}
    & \shortstack{Task\\Completion}
    & \shortstack{Information\\Grounding}
    & \\
    \midrule
    \raisebox{-0.2\height}{\includegraphics[height=0.9em]{figure/icon/glm.png}}\hspace{3pt}
    GLM 4.5v \cite{hong2025glm45v}
        & 0.033 & 0.028 & 0.030 & 0.033 & 0.000 & 0.033 & 0.204 & 0.126 & 0.061 \\
    \raisebox{-0.2\height}{\includegraphics[height=0.9em]{figure/icon/gemini.png}}\hspace{3pt}
    Gemini 2.5 Flash Lite \cite{comanici2025gemini25}
        & 0.033 & 0.167 & 0.033 & 0.033 & 0.000 & 0.033 & 0.258 & 0.089 & 0.081 \\
    \raisebox{-0.2\height}{\includegraphics[height=0.9em]{figure/icon/claude.png}}\hspace{3pt}
    Claude 4.5 Haiku \cite{anthropic2025haiku45}
        & 0.200 & 0.184 & 0.191 & 0.200 & 0.024 & 0.133 & 0.355 & 0.244 & 0.191 \\
    \raisebox{-0.2\height}{\includegraphics[height=0.9em]{figure/icon/internvl.png}}\hspace{3pt}
    InternVL 3.5 \cite{wang2025internvl35}
        & 0.200 & 0.299 & 0.188 & 0.200 & 0.048 & 0.134 & 0.367 & 0.255 & 0.211 \\
    \raisebox{-0.2\height}{\includegraphics[height=0.6em]{figure/icon/llama.png}}\hspace{3pt}
    Llama-4-Scout-17B16E \cite{meta2025llama4scout17b16e}
        & 0.333 & 0.086 & 0.316 & 0.333 & 0.143 & 0.062 & 0.267 & 0.320 & 0.233 \\
\bottomrule
    \end{tabular}
    \caption{(25) Task: FIPE Car Variant Lookup, Cross-Model Evaluation of Multimodal MCP Tool-Calling Trajectories on M$^3$-Bench.}
    \vspace{-8pt}
\end{table*}

\begin{table*}[!htb]
    \centering
    \footnotesize
    \setlength{\tabcolsep}{6pt}
    %
    \caption{(26) Task: Class Demand Estimation, Cross-Model Evaluation of Multimodal MCP Tool-Calling Trajectories on M$^3$-Bench.}
    \vspace{-8pt}
\end{table*}

\begin{table*}[!htb]
    \centering
    \footnotesize
    \setlength{\tabcolsep}{6pt}
    %
    \caption{(27) Task: NixOS Diagnostic, Cross-Model Evaluation of Multimodal MCP Tool-Calling Trajectories on M$^3$-Bench.}
    \vspace{-8pt}
\end{table*}

\end{document}